\documentclass[10pt,journal,compsoc]{IEEEtran}
\usepackage{amsmath,amssymb}
\usepackage{array}

\usepackage{theorem}
\usepackage{graphicx,psfrag}
\usepackage{enumerate,placeins,wrapfig,chngcntr}

\usepackage{microtype}
\usepackage{booktabs} %
\usepackage{multirow}
\usepackage{threeparttable}
\usepackage{hyperref}
\usepackage{epstopdf}
\usepackage{bbding}

\RequirePackage{fancyhdr}
\RequirePackage{xcolor}
\RequirePackage{algorithm}
\RequirePackage{algorithmic}
\RequirePackage{eso-pic} %
\RequirePackage{forloop}

\usepackage{caption}
\usepackage{subcaption}
\graphicspath{{figures/}}

\def \A {\mathcal{A}}
\def \T {\mathcal{T}}
\def \D {\mathcal{D}}

\def \Ir {\mathbf{I}}
\def \u {\mathbf{u}}
\def \v {\mathbf{v}}
\def \yy {\mathbf{y}}

\def \ome {\boldsymbol{\omega}}
\def \lamm {\boldsymbol{\lambda}}

\def \num {\mathtt{num}}
\def \net {\mathtt{net}}

\usepackage{bm}

\def\H{\mathbf{G}}
\def\S{\mathcal{S}}
\def\R{\mathbb{R}}
\def\Q{\mathbf{Q}}
\def\n{\mathbf{n}}
\def\b{\mathbf{b}}
\def\W{\mathbf{W}}

\def\z{\mathbf{z}}

\newtheorem{proposition}{Proposition}[section]
\newtheorem{assumption}{Assumption}[section]
\newtheorem{thm}{Theorem}[section]

\ifCLASSINFOpdf
\else
\fi

\begin{document}
	\title{Hierarchical Optimization-Derived Learning }

\author
{
	Risheng~Liu,~\IEEEmembership{Member,~IEEE,}
        Xuan~Liu,
        Shangzhi~Zeng,
        Jin~Zhang,
        and~Yixuan~Zhang %
    \IEEEcompsocitemizethanks{
	\IEEEcompsocthanksitem Risheng Liu and Xuan Liu are with the DUT-RU International School of Information Science and Engineering, Dalian University of Technology, Dalian 116024, China,
	and also with the Key Laboratory for Ubiquitous Network and Service Software of Liaoning Province, Dalian 116024, China (e-mail: rsliu@dlut.edu.cn, liuxuan\_16@126.com). 

    \IEEEcompsocthanksitem Shangzhi Zeng is with 
    the Department of Mathematics and Statistics, University of Victoria, Victoria, BC V8P5C2, Canada (e-mail: \\ zengshangzhi@uvic.ca).

    \IEEEcompsocthanksitem 
    Jin Zhang is with the Department of Mathematics, SUSTech International Center for Mathematics, Southern University of Science and Technology, National Center for Applied Mathematics Shenzhen, and Peng Cheng Laboratory, Shenzhen 518055, China (Corresponding author, e-mail: zhangj9@sustech.edu.cn).
    
    \IEEEcompsocthanksitem Yixuan Zhang is with the Department of Applied Mathematics, the Hong Kong Polytechnic University,
    Hong Kong SAR, China (e-mail: yi-xuan.zhang@connect.polyu.hk). 

}%

\thanks{Manuscript received April 19, 2005; revised August 26, 2015.}
}

	\markboth{Journal of \LaTeX\ Class Files,~Vol.~14, No.~8, August~2015}%
	{Shell \MakeLowercase{\textit{et al.}}: Bare Advanced Demo of IEEEtran.cls for IEEE Computer Society Journals}

	\IEEEtitleabstractindextext{%
		\begin{abstract}  
			
In recent years, by utilizing optimization techniques to formulate the propagation of deep model, a variety of so-called Optimization-Derived Learning (ODL) approaches have been proposed to address diverse learning and vision tasks. Although having achieved relatively satisfying practical performance, there still exist fundamental issues in existing ODL methods. In particular, current ODL methods tend to consider model constructing and learning as two separate phases, and thus fail to formulate their underlying coupling and depending relationship. In this work, we first establish a new framework, named Hierarchical ODL (HODL), to simultaneously investigate the intrinsic behaviors of optimization-derived model construction and its corresponding learning process. Then we rigorously prove the joint convergence of these two sub-tasks, from the perspectives of both approximation quality and stationary analysis. To our best knowledge, this is the first theoretical guarantee for these two coupled ODL components: optimization and learning. We further demonstrate the flexibility of our framework by applying HODL to challenging learning tasks, which have not been properly addressed by existing ODL methods. Finally, we conduct extensive experiments on both synthetic data and real applications in vision and other learning tasks to verify the theoretical properties and practical performance of HODL in various application scenarios.

		\end{abstract}
		
		}

	\maketitle

	\IEEEdisplaynontitleabstractindextext

	\IEEEpeerreviewmaketitle

	\ifCLASSOPTIONcompsoc
	\IEEEraisesectionheading{\section{Introduction}\label{sec:introduction}}
	\else
	\section{Introduction}
	\label{sec:introduction}
	\fi

\IEEEPARstart{O}ptimization-Derived Learning (ODL) is a class of methods for constructing deep models based on optimization techniques~\cite{monga2021algorithm,chen2021learning}
and has been widely used in different vision tasks in the past years~\cite{liu2022optimization,feurer2019hyperparameter,thornton2013auto,schuler2015learning,chen2016trainable}. 
Specifically, each of the optimization iteration is regarded as a layer of the network. All of these layers are concatenated to form a deep model. Passing through the network is equivalent to performing a finite number of optimization iterations. In addition, the optimization algorithm parameters (e.g., model parameters and regularization coefficients) are transferred to the learning variables in the network.  In this way, the training network can be naturally interpreted as a parameterized optimization model, effectively overcoming the lack of interpretability in most of the traditional neural networks and leading to excellent performance as well.

	Hence, a core problem of ODL is how to design the network structure based on the optimization model. In other words, ODL focuses on how to embed learnable modules into the optimization model.
	Depending on the way in which the learnable module is handled, existing approaches can be broadly classified into two main categories, respectively called ODL based on Unrolling with Numerical Hyper-parameters (UNH), which aims to embed learnable modules under reliable theories and focuses on the convergence guarantee of the algorithm~\cite{ablin2019learning, liu2019deep,liu2019convergence},
	and ODL Embedded with Network Architectures (ENA), which heuristically embeds learnable modules into the optimization algorithm, focusing more on the performance of practical tasks~\cite{gregor2010learning,chen2018theoretical,xie2019differentiable,chan2016plug,zhang2019deep,zhang2020plug}. Unfortunately, UNH usually treats optimization and networks as two separated modules, and ENA ignores the optimization process after designing the network structure. As a new framework, our Hierarchical ODL (HODL) also treats optimization and networks as two modules. However, unlike existing approaches, HODL establishes the nested relationship between optimization and networks via a hierarchical structure, and specifies the influence of learnable networks on the optimization process.

	\begin{table*}[htbp]
		\centering
		\caption{Various ODL methods whose base models include Proximal Gradient (PG), Augmented Lagrangian Method (ALM), and Half-Quadratic Splitting (HQS).
			Existing ODL methods are widely used in many fields, but the analysis of convergence, especially the convergence of learning variables $\ome$, is still insufficient. 
		}
		\begin{tabular}{c|c|c|c|c|c|c}
			\hline
			\multirow{1}{*}{Method} &\multirow{1}{*}{Category}&\multirow{1}{*}{Base Model} & $\u^K\rightarrow\u^*$ &$\inf\varphi_K(\ome)\rightarrow\inf\varphi(\ome)$&$\nabla\varphi(\ome)\rightarrow 0 $&	\multirow{1}{*}{Application}  \\
			\hline
			
			ISTA-Net~\cite{2017ISTA}  &\multirow{6}{*}{ENA}& PG      &\XSolidBrush&\XSolidBrush &\XSolidBrush & CS Reconstruction\\
			ADMM-Net~\cite{yang2017admm}& & ALM    & \XSolidBrush &\XSolidBrush &\XSolidBrush& CS-MRI\\
			DUBLID~\cite{li2020efficient}& & HQS     &\XSolidBrush &\XSolidBrush&\XSolidBrush & Image Deconvolution \\
			LISTA~\cite{chen2018theoretical}& & PG      & \CheckmarkBold &\XSolidBrush &\XSolidBrush&Sparse Coding \\
			DLADMM~\cite{xie2019differentiable} & & ALM  & \CheckmarkBold &\XSolidBrush&\XSolidBrush & Image Deblurring\\
			PnP~\cite{ryu2019plug} && ALM, PG    & \CheckmarkBold &\XSolidBrush&\XSolidBrush & Image Super Resolution\\
			
			\hline
			PADNet~\cite{liu2019deep}  &\multirow{3}{*}{UNH}& ALM  &\CheckmarkBold &\XSolidBrush &\XSolidBrush& Image Haze Removal\\
			FIMA~\cite{liu2019convergence} & & PG  &\CheckmarkBold &\XSolidBrush &\XSolidBrush& Image Restoration \\
			OISTA~\cite{ablin2019learning} && PG  &\CheckmarkBold &\XSolidBrush &\XSolidBrush& Sparse coding \\
			\hline
			\hline
			\multirow{4}{*}{HODL}    &\multirow{5}{*}{Both}&\multirow{5}{*}{\begin{tabular}{c}
					Flexible					
			\end{tabular}}   &\multirow{5}{*}{\CheckmarkBold} & \multirow{5}{*}{\CheckmarkBold} &\multirow{5}{*}{\CheckmarkBold} & \multirow{5}{*}{\begin{tabular}{c}
					Sparse Coding\\
					Image Restoration\\
					Hyper-parameter Optimization\\
					Few-shot Learning\\
					Generative Adversarial Learning\\
			\end{tabular} }\\
			&&&&&\\
			\multirow{2}{*}{(Ours)} &&&&&\\
			&&&&&\\
			&&&&&\\
			\hline
			
		\end{tabular}%
		\label{tab:table 1}%
	\end{table*}%

	\subsection{Related Works} 
	\label{sec:related works}

	As introduced in the last paragraph, existing ODL approaches are classified into UNH and ENA.
	Earlier UNH methods usually set learnable modules as some hyper-parameters in the optimization algorithm which do not affect the convergence, such as the step size~\cite{ablin2019learning}. These methods avoid damaging the convergence results, but the number of learnable parameters is limited, making it hard to be applied to various practical tasks flexibly.
	In recent years, some UNH methods have embedded learnable modules to replace the descent direction in optimization as a novel perspective~\cite{liu2019deep,liu2019convergence}. 
	In particular, they use the learnable module to provide the actual descent directions and set a convergence criterion to adjust it. 
	For these methods, if the learnable modules are decoupled from the optimization model, the system flexibility is greatly enhanced. 
	As for ENA, it often considers the optimization objective as a prototype to motivate its network model design for specific tasks. 
	Specifically, ENA greatly improves the flexibility of traditional optimization by replacing some structures in the optimization model directly with learnable modules having similar effects. %
	Some ENA methods only regard the linear layers of networks as learnable matrices.
	In LISTA~\cite{gregor2010learning} and CPSS~\cite{chen2018theoretical}, a non-linear feed-forward predictor is trained to produce the best approximation of sparse coding;
	in DLADMM~\cite{xie2019differentiable}, some learnable network modules are embedded into LADMM, and some learnable parameters are embedded into the proximal operator. 
	In addition, some other ENA methods utilize more general networks.
	For instance, 
	ISTA-Net~\cite{2017ISTA} is based on ISTA as the fundamental iteration scheme, and adds a range of filters to learn parameters for image compressive sensing (CS);
	Plug-and-Play ADMM~\cite{chan2016plug} replaces the projection gradient operator in ADMM with an implicit denoising module;
	pre-trained-CNN-based modules such as DPSR and DPIR~\cite{zhang2019deep,zhang2020plug} are introduced to handle image restoration problems such as deconvolution, denoising, and super-resolution.
	\textcolor{black}{Furthermore, a class of methods called learning to optimize~\cite{chen2021learning,li2017learning} can be regarded as ODL methods and unified under the HODL framework.
	Indeed, any method that utilizes an optimization model as the assistance in network construction falls within the scope of HODL. 	
	Specifically, approaches that focus on using the network to assist in optimizing the original objective function are classified into UNH, while those methods that utilize optimization as an assistance in network construction and aim to enhance the task performance rather than to optimize a specific objective function are referred to as ENA.}

	However, these existing ODL methods ignore the relationship between learnable modules and optimization models, leading to some drawbacks in methodology and theory.  
	From the methodological perspective,
	since the convergence of UNH depends entirely on the original optimization algorithm, its performance is limited to manually designed target features, 
	and it is impossible to further narrow the gap between target features and real-world tasks. 
	ENA relies on the fact that the pre-trained network modules need to indeed have similar performance to the replaced part, which usually can only be promised by proper pre-training.
	Furthermore, existing ODL approaches have another common shortcoming in methodology: they deal with the optimization model and the learnable module separately, meaning that existing learnable modules are often trained independently of the optimization model. 
	While it is still possible to obtain the modules needed to optimize the model on the macroscopic level (e.g., replacing soft threshold operation with noise reduction modules), this has led to a gap between the modules needed to optimize the model and those that are actually learning.
	Although new methods exist to better isolate the learnable modules, this gap cannot be fundamentally addressed.

	In terms of theoretical perspective,
	some works have analyzed the convergence of optimization process with the help of classic optimization techniques.
	To be specific, in~\cite{chan2016plug,teodoro2018convergent,sun2019online} authors consider the non-expansive property of optimization iterative process under the condition that the embedded networks are bounded; in~\cite{ryu2019plug} the convergence is achieved when the Lipschitz constant of network residuals is strictly smaller than one.
	However, these works only focus on the convergence towards the fixed points of the approximated optimization model,
	but not the solution to the intrinsic task considering both optimization models and learnable modules.
	An intuitive treatment to handle this problem is to learn fewer learning variables.
	For example, in~\cite{ablin2019learning}, only the step size of ISTA is learned,
	which nevertheless restricts the model.
	In addition, for ODL, additional artificially designed corrections are needed when learning the network.
	For example, in~\cite{liu2019convergence,moeller2019controlling,heaton2020safeguarded} authors manually design various rules to decide updates from the temporary updates generated by networks and optimization algorithms.
	However, the lack of learning variables and the manual design of rules severely limit these methods.
	Furthermore, ignoring the convergence of learning process also leads to some theoretical defects.
	First, as aforementioned, learning variables are fixed in the optimization process, and thus it is only able to consider the convergence of optimization variables, instead of the convergence of learning variables in learnable modules.
	Second, the learnable modules for ODL are too complex to determine the relationship between the true solution to the task and the obtained fixed points.
	Moreover, the convergence analysis of most existing ODL methods is developed from a specific optimization framework,
	so it is difficult to be extended to other optimization models.

	\subsection{Our Contributions}

	To address the aforementioned problems, we explicitly model ODL as a hierarchical relationship paradigm between the learnable module and the optimization algorithm, called HODL.
	Subsequently, in order to jointly train the optimization variables and the learning variables, we propose the corresponding solution strategy for solving HODL. 
	We further put forward its simplified version to speed up the algorithm, 
	and the simplified solution strategy can contain existing gradient-based unrolling algorithms as special cases.
	After that we provide the convergence analysis for this algorithmic framework.
	To be specific, we strictly prove the detailed theoretical properties to guarantee the joint convergence of optimization variables and learning variables, 
	containing the convergence on approximation quality analysis and stationary analysis.
	We also conduct plenty of experiments on various learning and vision tasks to verify the effectiveness and wide applications of HODL. 
	Our contributions can be summarized as follows,
	and the overall comparison of our HODL and existing ODL methods is displayed in Table~\ref{tab:table 1}.

	\begin{itemize}
		
		\item Unlike existing works that only pay attention to either learning or optimization process in ODL, we take both learning and optimization into consideration as two nested solution processes and formulate the general ODL paradigm, allowing us to further analyze the hierarchical relationship between the optimization and learning variables.
		
		\item From the hierarchical perspective, we build up the HODL framework and provide the novel and general ODL solution strategy. Our framework considers the nested relationship between optimization and learning, making it possible to jointly train optimization variables and learning variables.
		
		\item This work provides the strict joint convergence analysis of optimization variables and learning variables under the HODL framework,
		both on the approximation quality and on the stationary convergence.
		We additionally put forward a fast algorithm for HODL and its convergence analysis, which significantly extend the results in~\cite{liu2022optimization}.
		
		\item We apply our HODL framework and the solution strategies to various learning tasks, containing sparse coding as the toy example, and image processing tasks (e.g., rain streak removal, image deconvolution, and low-light enhancement). In addition, our HODL can also handle bilevel optimization tasks that cannot be handled by existing ODL methods, such as adversarial learning, hyper-parameter optimization, and few-shot learning.

	\end{itemize}

	\section{The Proposed Algorithmic Framework} 
	\label{sec:algorithm}
	
	In this section, we first put forward the general ODL paradigm, and introduce our Hierarchical Optimization-Derived Learning (HODL) framework to unify the optimization algorithms and learnable modules. 
	Then the solution strategies for this HODL framework are provided.

	\subsection{The General ODL Paradigm}\label{sec:general ODL}
	ODL usually translates the application problem into two parts of the optimization problem, the task term and the learnable term, with respect to the optimization variable~$\u~\in~{U}$. The task term is usually an objective function $f(\u)$ that represents the dependence of the solution of $\u$ on the task itself. The learnable term, on the other hand, can be classified into two common forms, the regularization term $g(\u)$ and the linear constraint term $\A(\u)=\yy$, which are used to represent the task prior that aids in solving the problem. Hence, ODL usually transforms the specific task into the following form

	\begin{equation}\label{eq:min f}
		\min_{\u \in U} \overbrace{f(\u)}^{\text{Task\ Term}}+ \overbrace{\underbrace{g(\u,\ome)}_{\text{Regularization}}  \text{and/or \ }\underbrace{ \text{s.t. } \A(\u,\ome)=\yy(\ome)}_{\text{Constraint}}}^{\text{Learnable\ Term}},
	\end{equation}
	where $\ome$, the parameters of learnable term, is called the learning variable.
	Denote the solution set with respect to $\u$ for a given $\ome$ to be $\S(\ome)$,
	and denote the corresponding algorithmic operator 
	for solving Eq.~\eqref{eq:min f} to be~$\D$.
	In classical optimization methods $\D$ is usually constructed manually by optimization experts based on theory and experience.
	As a paradigm for designing network structures, ODL designs the network from an optimization perspective. To be specific, by building the model based on classical optimization process as the structural basis and embedding learnable modules, ODL generates a complete network structure with both interpretability of optimization models and learnability of neural networks. This paradigm is flexible enough that the learnable module can be not only the hyper-parameters in the numerical optimization process, but also the entire networks used to replace certain process steps. 
	The corresponding networks are respectively denoted as $\D_{\mathtt{num}}$ and $\D_{\mathtt{net}}$.

	Unfortunately, existing ODL methods only consider optimization when building the initial network structure, and follow the ordinary deep neural network strategy during training, instead of combining optimization and learning. 
	This splits ODL into two parts:	during the training procedure, they only care about the convergence of learning variables $\ome$ and ignore the iterations of optimization variables $\u$;
	while in testing, they fix $\ome$ and hope $\u$ to converge in the optimization process under the fixed network structure.

	\subsection{Our Meta Optimization Framework \protect\footnote{In numerical optimization, meta-optimization is the use of one optimization method to tune another optimization method~\cite{krus2013performance}.}
	}
	\label{sec:GKM}

	To address the fragmentation of optimization and learning processes, we use the idea of optimization not only when building the network structure, but also during the training procedure. 
	Despite the embedded learnable module, the network structure of ODL can still be considered as an optimization process for solving a specific problem. 
	Hence, by nesting the results of the optimization process into the inputs of the learning process, 
	for the problem in Eq.~\eqref{eq:min f},
	we can transfer it to the following
	\begin{equation} %
	    \begin{aligned}
	        & \min\limits_{\u\in U,\ome\in\Omega}  \ell(\u,\ome),  
	         \text{ s.t. } \u \in \S(\ome), 
	    \end{aligned}
	\end{equation}
	where $\ell$ is the objective function.

	Next, we put forward a unified form in dealing with all kinds of problems in Eq.~\eqref{eq:min f}, which also facilitates our subsequent analysis.
	Specifically, each iteration of the ODL method constitutes an operator origin from the optimization algorithm but embedded with a learnable module, and the result of a stable iteration is taken as the output of ODL. 
	Therefore, a reasonable assumption is to consider the operator as non-expansive and the output of ODL as the fixed point of the corresponding iterative operator for solving Eq.~\eqref{eq:min f}. 
	Therefore, we model the optimal solution of ODL uniformly by $\u =\D(\u,\ome)$ to find the fixed point,
	where $\ome$ is the learning variable, and $\D$ is the non-expansive operator.
	Here $\D(\cdot, \ome) \in \left\{  \D_\num(\cdot, \ome) \circ \D_\net(\cdot, \ome) \right\}$,
	where $\circ$ represents compositions of operators.
	Same as introduced in Section~\ref{sec:general ODL},~$\D_\num$ regards the hyper-parameters in the numerical optimization process as learnable modules, while~$\D_\net$ replaces certain process steps to be networks directly.
	Hence, this form not only includes optimization algorithms,
	but also contains other implicitly defined models,
	which originate from optimization but are added with learnable modules additionally.
	The process to find the fixed point can be implemented via the classical Krasnoselskii-Mann updating scheme~\cite{reich2000convergence} generalized with learning variables $\ome$, in the form of 
	$
	\T(\u^k,\ome) = \u^k + \alpha (\D(\u^k,\ome) - \u^k),
	$ 
	as the $k$-th iteration step,	
	where $\alpha\in(0,1)$.
	Note that if $\D$ is non-expansive, then $\T$ is an $\alpha$-averaged non-expansive operator. 
	Furthermore, the fixed point of $\D$ is also a fixed point of $\T$.
	In experiments, for guaranteeing that $\D$ is non-expansive,
	some normalization techniques such as spectral normalization~\cite{miyato2018spectral} are implemented on parameters.
	By choosing  the solution of the fixed point problem $\u =\T(\u,\ome)$ as the input for learning $\ome$, the hierarchical formulation of a general ODL problem can be expressed in the following form
	\begin{equation}\label{eq:bilevel_fix}
		\min\limits_{\u\in U,\ome\in\Omega} \ell(\u,\ome),
		\text{  s.t. } \u =\T(\u,\ome),
	\end{equation}
	where $\ell$ is the loss function corresponding to the learning process, and $\T$ denotes the optimization process.
	We call problems of this formulation as Hierarchical Optimization-Derived Learning (HODL), which also serve as our meta optimization framework.	
	\textcolor{black}{In learning to optimize, a class of methods are designed to learn an optimizer to optimize an objective function for a specific task, and this optimizer corresponds to the operator $\T$ under our HODL framework.}

	Actually, HODL can overcome several shortcomings in existing ODL methods mentioned in Section~\ref{sec:introduction} thanks to its hierarchical modeling.
	From the viewpoint of theory, HODL makes it possible to study the joint convergence of $\ome$ and $\u$ under their nested relationship,
	in place of only considering one of them independently.
	Hence, instead of only obtaining the fixed points of the optimization process for a fixed $\ome$, we can approach the true optimal solution of the whole problem.
	We will provide the detailed convergence analysis in Section~\ref{sec:theoritical}.
	From the viewpoint of applications,
	in the practical training procedure, the learning variables $\ome$ are also adjusted along with the iterations of optimization variables~$\u$,
	rather than just embedding a network that ignores the optimization structure.

	\subsection{Efficient Solution Strategy} \label{sec: solution strategy}

	Next we establish the algorithm to simultaneously solve the optimization variables $\u$ and learning variables~$\ome$. 
	Existing ODL methods usually update the optimization variables with fixed pre-trained learning variables, ignoring the nested relationships in ODL when training the optimization variables and learning variables and failing to solve them together.

	\textbf{The Nested Learning Iteration.} 
	To begin with, the learning variables $\ome$ are nested into the optimization variables $\u$.
	Note that existing ODL approaches ignore the hierarchical structure of $\ome$ and $\u$ in modeling, so their algorithms also do not contain their hierarchy and are unavailable under our HODL framework.
	We design the training of~$\ome$ %
	in order that the nested relationship between $\u$ and $\ome$ can be effectively exploited. 
	Specifically, each iterative step 
	of~$\u$ is parameterized by $\ome$, so the iteration result of~$\u$ is a function of $\ome$, i.e., $\u^k(\ome)$.
	This reveals the dependence of optimization variables $\u$ on the learning variables $\ome$, 
	and thus the complete optimization iteration of $\u$ (inner loop) is embedded within the learning iteration of $\ome$ (outer loop).
	Hence, the objective function of learning $\ome$ contains the entire iterative trajectory of $\u$, which effectively exploits their nested relationship.

	\textbf{The Nested Optimization Iteration.} 
	For the iteration of optimization variables $\u$, we also add an additional nested structure related to the learning process. 
	To begin with, we compute the iterative direction $\v_{l}$ from the optimization process corresponding to $\T$ in Eq.~\eqref{eq:bilevel_fix} (lower level). 
	At the $k$-th step, to 
	approach the fixed point of $\T(\cdot,\ome)$ 
	for a given $\ome$, $\v_{l}^{k} =\T \left( \u^{k-1},\ome \right)$ is defined as an update direction of $\u$.
	Note that here the operator $\T$ is adjusted to be non-expansive under the induced norm $\| \cdot \|_{\H_{\ome}}$ where $\H_{\ome}$ is a positive-definite correction matrix parameterized by~$\ome$ and will be discussed in detail in Section~\ref{sec:theoritical}.
	Next, we compute another iterative direction $\v_{u}$ from the learning process in Eq.~\eqref{eq:bilevel_fix} (upper level). 
	It makes our updating direction of $\u$ able to utilize the information of $\ome$ by using the gradient of loss function $\ell$ with respect to $\u$.
	Nevertheless, directly applying its gradient may destroy the non-expansive property with respect to $\| \cdot \|_{\H_{\ome}}$.
	Consequently, for the consistent non-expansive property with direction $\v_{l}$, we further add an additional correction $\H_{\ome}^{-1}$ to the gradient of $\ell$,
	and request the corresponding step sizes $s_k$ to be a decreasing sequence
	for assuring the correctness of this iterative direction $\v_{u}$,
	i.e., 
	$
	\v_{u}^{k} =\u^{k-1}-s_k \H_{\ome}^{-1} \frac{\partial }{\partial \u}\ell(\u^{k-1},\ome),
	$
	where $s_k \rightarrow 0$ as $k$ increases.
	Lastly, inspired by~\cite{liu2022general}, 
	we generate the final updating direction of $\u$ by aggregating the two iterative directions $\v_{l}$ and $\v_{u}$ via a linear combination under the projection,
	i.e., $\u^k =\mathtt{Proj}_{U, \H_{\ome}}\left(\mu \v_{u}^{k}+(1-\mu) \v_{l}^{k}\right)$,
	where $\mu \in (0,1)$.
	Here the projection operator $\mathrm{Proj}_{U,\H_{\ome}}(\cdot)$ is associated to $\H_{\ome}$ with the definition $\mathtt{Proj}_{U,\H_{\ome}}(\u) = \mathrm{argmin}_{\bar{\u} \in U} \|\bar{\u} - \u\|_{\H_{\ome}}$.
	Note that in the theoretical analysis part, the projection is only used to guarantee the boundedness of $\u^k$;
	while in practical experiments and applications, generally $U$ is set to be such a large bounded set or even unbounded $\R^n$ that the projection operator can be ignored.
	To conclude, the iterations of optimization variables $\u$ in our solution strategy for HODL reads as
	\begin{equation}\label{eq:simple_bilevel_alg}
		\left\{
		\begin{aligned}
			\v^k_l(\ome) & =  \T(\u^{k-1}(\ome),\ome), \\
			\v^k_u(\ome) & = \u^{k-1}(\ome) - s_k \H_{\ome}^{-1}\frac{\partial }{\partial \u}\ell(\u^{k-1}(\ome),\ome), \\
			\u^k(\ome) & = \mathtt{Proj}_{U,\H_{\ome}} \big( \mu \v^k_u(\ome) + (1-\mu) \v^k_l(\ome) \big),
		\end{aligned}\right	.
	\end{equation}
	where $k = 1,\ldots, K$.

	Here our solution strategy to solve the HODL problem is with aggregation of $\v_l$ and $\v_u$, so it is shortened as HODL with aggregation (aHODL for short).
	On the other hand, from the viewpoint of computational efficiency in practical applications,
	the algorithm can be further improved.
	To be specific, a computational drawback comes from the need for gradual decay of $s_k$ in Eq.~\eqref{eq:simple_bilevel_alg}, which leads to an increase in the number of training iteration.
	In addition, $\H^{-1}_{\ome}$ may be challenging to compute according to different forms of $\D$, and even $\H_{\ome}$ itself may be hard to estimate. 
	Therefore, we adjust aHODL and put forward a simplified HODL (sHODL for short) without the aggregation step as in Eq.~\eqref{eq:simple_bilevel_alg}.
	That is, we let $\mu$ in Eq.~\eqref{eq:simple_bilevel_alg} to be $0$,
	and then $\u^K(\ome)$ is iterated as 
	\begin{equation}\label{eq:sHODL}
	    \u^k(\ome) = \mathtt{Proj}_{U,\H_{\ome}} \left(  \v^k_l(\ome)\right),
	\end{equation}
	where $\v^k_l(\ome) =  \T(\u^{k-1}(\ome),\ome)$, and $k = 1,\ldots, K$.
	Compared with Eq.~\eqref{eq:simple_bilevel_alg}, sHODL, the strategy without aggregation, is simpler to implement with higher efficiency as a fast algorithm than aHODL.
	Hence, our HODL framework can then be extended to more application tasks.
	Convergence of aHODL and sHODL will be discussed in the next section,
	which also indicates the superiority of aHODL over sHODL in theory.
	The algorithmic flow for aHODL and sHODL is summarized in Algorithm~\ref{alg:bmo}.

	\begin{algorithm} %
		\caption{HODL}\label{alg:bmo}
		\begin{algorithmic}[1]
			\REQUIRE Step sizes $\{s_k\}$, $\gamma$ and parameter $\mu$.
			\STATE Initialize $\ome^0$ .
			\FOR {$t=1\rightarrow T$}
			\STATE Initialize  $\u^0$.
			\FOR {$k=1\rightarrow K$}
			\STATE Compute $\u^k$ by Eq.~\eqref{eq:simple_bilevel_alg} (aHODL) or the simplified version Eq.~\eqref{eq:sHODL} (sHODL).
			\ENDFOR
			\STATE $\ome^{t}= \mathtt{Proj}_{\Omega,\H_{\ome}} \big( \ome^{t-1}-\gamma\frac{\partial}{\partial \ome}\ell(\u^K(\ome^{t-1}),\ome^{t-1}) \big)$.
			\ENDFOR
		\end{algorithmic}
	\end{algorithm}

	\section{Theoretical Analysis}
	\label{sec:theoritical}

	In this section,
	we propose the convergence analysis of the solution strategies for HODL problems in Eq.~\eqref{eq:bilevel_fix} 
	with respect to both optimization variables $\u$ and learning variables~$\ome$. 
	Our analysis for the solution strategy of HODL is separated into two parts,
	the approximation quality analysis on the convergence of optimal value in Section~\ref{sec: theo: approximation quality}, and the stationary analysis on the convergence of stationary points in Section~\ref{sec: theo: stationary}.
	For the simplified solution strategy without aggregation sHODL as mentioned in Section~\ref{sec: solution strategy}, 
	we also provide further analysis in Section~\ref{sec: theo: sHODL}.
	Note that since HODL in Eq.~\eqref{eq:bilevel_fix} is a general form of ODL problems, our analysis also serves as a unified route of theoretical analysis for other methods and more problems with hierarchical structures.

	To begin with, we denote the fixed point set of operator~$\T$ to be $\mathtt{Fix}(\T(\cdot,\ome))$ for a given $\ome$,
	and then the HODL problem in Eq.~\eqref{eq:bilevel_fix} can be rewritten as 
	\begin{equation}\label{eq:appe phi_def}
		\min_{\ome \in \Omega} \ \varphi(\ome),\quad  \text{where} \quad \varphi(\ome) := \inf_{\u \in \mathtt{Fix}(\T(\cdot,\ome)) \cap U } \ \ell(\u,\ome).
	\end{equation}
	In Algorithm~\ref{alg:bmo}, $\u^K(\ome)$ is obtained by iterating as Eq.~\eqref{eq:simple_bilevel_alg} (aHODL) or its simplification in Eq.~\eqref{eq:sHODL} (sHODL), to solve the simple bilevel problem $\inf_{\u \in \mathtt{Fix}(\T(\cdot,\ome)) \cap U } \ \ell(\u,\ome)$.
	Substituting $\u^K(\ome)$ for $\u$ in~$\ell(\u,\ome)$ of Eq.~\ref{eq:appe phi_def},
	we have its approximation problem as the following 
	\begin{equation}\label{eq:phiK_def}
		\min_{\ome \in \Omega} \ \varphi_K(\ome) := \ell(\u^K(\ome),\ome),
	\end{equation}
	which is only about the variable $\ome$, and is solved by the sequence $\{\ome^t\}$ generated by Algorithm~\ref{alg:bmo}.

	\subsection{Approximation Quality Analysis} \label{sec: theo: approximation quality}

	In this part, we show that Eq.~\eqref{eq:phiK_def} obtained by aHODL is actually an appropriate approximation to Eq.~\eqref{eq:bilevel_fix},
	meaning that any limit point $(\bar{\u},\bar{\ome})$ of the sequence $\left\{\left( \u^K(\ome^K),\ome^K\right)\right\}$ is a solution to the HODL problem in Eq.~\eqref{eq:bilevel_fix},
	where $\ome^K \in \mathrm{argmin}_{\ome \in\Omega}\varphi_{K}(\ome)$ as a solution to Eq.~\eqref{eq:phiK_def} is generated by Algorithm~\ref{alg:bmo} and $\u^K(\ome)$ is computed from Eq.~\eqref{eq:simple_bilevel_alg}.
	Hence, we can approach the optimal solution of HODL in Eq.~\eqref{eq:bilevel_fix} by solving Eq.~\eqref{eq:phiK_def}.

	We make the following standing assumptions throughout this part, and then show that Algorithm~\ref{alg:bmo} can achieve convergence in the sense of approximation quality under mild conditions.

	\begin{assumption}\label{assum_F}
		$\Omega$ is a compact set and $U$ is a convex compact set. $\mathtt{Fix}(\T(\cdot,\ome))$ is nonempty for any $\ome \in \Omega$. $\ell(\u,\ome)$ is continuous on $\R^n \times \Omega$. For any $\ome \in \Omega$, $\ell(\cdot,\ome) : \R^n \rightarrow \R$ is $L_\ell$-smooth, convex and bounded below by $M_0$.
	\end{assumption}

	Please notice that function $\ell$ is usually defined to be the MSE loss, so Assumption~\ref{assum_F} is quite standard for ODL problems~\cite{ryu2019plug,zhang2020plug}. 
	Next we present some necessary  preliminaries.
	For any two matrices $\H_1, \H_2 \in \R^{n \times n}$, we consider the following partial ordering relation:
	\[
	\H_1 \succeq \H_2 \quad \Leftrightarrow \quad  \langle \u,\H_1\u \rangle \ge \langle \u,\H_2\u \rangle, \quad \forall \u \in \R^n.
	\]
	If $\H \succ 0$, then $\langle \u_1, \H \u_2 \rangle$ for $\u_1,\u_2 \in \R^n$ defines an inner product on $\R^n$. 
	Denote the induced norm with $\| \cdot \|_\H$, %
	i.e., $\| \u \|_\H := \sqrt{\langle \u,\H \u \rangle}$ for any $\u \in \R^n$.
	We assume that $\D(\cdot,\ome)$ satisfies the following assumptions throughout this part.
	\begin{assumption} \label{assum_T}
		There exist $\H_{ub} \succeq \H_{lb} \succ 0$, %
		such that for each $\ome \in \Omega$, there exists $\H_{ub} \succeq \H_{\ome} \succeq \H_{lb}$ such that
		\begin{itemize}
			\item[(1)] $\D(\cdot,\ome)$ is non-expansive with respect to $\| \cdot \|_{\H_{\ome}}$, i.e., for all $(\u_1,\u_2) \in \R^n \times \R^n$,
			\begin{equation*}
				\|\D(\u_1,\ome) - \D(\u_2,\ome) \|_{\H_{\ome}} \le \| \u_1 - \u_2\|_{\H_{\ome}}.
			\end{equation*}
			\item[(2)] $\D(\cdot,\ome)$ is closed, i.e., $\mathrm{gph} \,\D(\cdot,\ome)$ is closed, where
			\[
			\mathrm{gph} \,\D(\cdot,\ome) := \{(\u,\v) \in \R^n \times \R^n ~|~ \v = \D(\u,\ome)\}.
			\]
		\end{itemize}
	\end{assumption}

	The non-expansive property of $\T(\cdot,\ome)$ in Eq.~\eqref{eq:bilevel_fix} can be obtained immediately from that of $\D(\cdot,\ome)$ in Assumption~\ref{assum_T}~\cite{Heinz-MonotoneOperator-2011}[Proposition~4.25].
	Then
	we can prove that the sequence $\{\u^k(\ome)\}$ generated by Eq.~\eqref{eq:simple_bilevel_alg} not only converges to the solution set of $\inf_{\u \in \mathtt{Fix}(\T(\cdot,\ome)) \cap U } \ \ell(\u,\ome)$, %
	but also admits a uniform convergence towards the fixed point set $\mathtt{Fix}(\T(\cdot,\ome))$ 
	with respect to $\| \u^k(\ome) - \T(\u^k(\ome),\ome) \|_{\H_{lb}}^2$ for $\ome \in \Omega$. %
	Thanks to the uniform convergence property of the sequence $\{\u^k(\ome)\}$, 
	inspired by the arguments used in~\cite{liu2022general},
	we can establish the convergence on both $\u$ and $\ome$ of Algorithm~\ref{alg:bmo} towards the solution of HODL problem in Eq.~\eqref{eq:bilevel_fix}.
	The convergence results of approximation quality are summarized in the following theorem.
	Please refer to our conference version in~\cite{liu2022optimization} for detailed proofs.

\begin{thm}
	Suppose Assumptions~\ref{assum_F} and~\ref{assum_T} are satisfied.
	Let $\{\u^k(\ome)\}$ be the sequence generated by Eq.~\eqref{eq:simple_bilevel_alg} with $\mu \in (0,1)$ and $s_k = \frac{s}{k+1}$, 
	where $s \in (0, \frac{\lambda_{\min}(\H_{lb})}{L_{\ell}} )$, 
	and $\lambda_{\min}(\H_{lb})$ denotes the smallest eigenvalue of matrix $\H_{lb}$. 
	\begin{itemize}
		\item[(1)] For any $\ome \in \Omega$, we have
		\begin{equation*}
			\begin{array}{c}
				\lim\limits_{k \rightarrow \infty}\mathrm{dist}(\u^k(\ome),\mathtt{Fix}(\T(\cdot,\ome)) = 0,
			\end{array}
		\end{equation*}
		and
		\begin{equation*}
			\begin{array}{c}
				\lim\limits_{k \rightarrow \infty}\ell(\u^k(\ome),\ome) =  \varphi(\ome).
			\end{array}
		\end{equation*}
		Furthermore, there exits $C > 0$ such that for any $\ome \in \Omega$,
		\[
		\| \u^k(\ome) - \T(\u^k(\ome),\ome) \|_{\H_{lb}}^2 \le C\sqrt{\frac{1+\ln(1+k)}{k^{\frac{1}{4}}}}.
		\]
		
		\item[(2)] Let $\ome^K \in \mathrm{argmin}_{\ome \in\Omega}\varphi_{K}(\ome)$, and we have
		any limit point $(\bar{\u},\bar{\ome})$ of the sequence $\{(\u^K(\ome^K),\ome^K) \}$ is a solution to the problem in Eq.~\eqref{eq:bilevel_fix}, i.e., $\bar{\ome}\in\mathrm{argmin}_{\ome\in\Omega}\varphi(\ome)$ and $\bar{\u} = \T(\bar{\u},\bar{\ome}) $.
		Furthermore, $\inf_{\ome \in \Omega}\varphi_K(\ome) \rightarrow \inf_{\ome \in \Omega} \varphi(\ome)$ as $K \rightarrow \infty$.
	\end{itemize}
	
\end{thm}

	\subsection{Stationary Analysis} \label{sec: theo: stationary}
	
	Next, we put forward the convergence analysis of our solution strategy with aggregation aHODL (using Eq.~\eqref{eq:simple_bilevel_alg} to compute $\u^K$ in Algorithm~\ref{alg:bmo}) %
	on stationary points.	
	That is, for any limit point $\bar{\ome}$ of the sequence $\{\ome^K\}$, we have $\nabla \varphi(\bar{\ome}) = 0$, where $\varphi(\ome)$ is defined in Eq.~\eqref{eq:appe phi_def}.

	Here we make $U = \R^n$ and suppose the operator $\T$
	has a unique fixed point, 
	which means
	the fixed point set $\mathtt{Fix}(\T(\cdot,\ome))$ is a singleton. %
	We denote the unique solution by $\u^*(\ome)$. 
	Our analysis is partly inspired by~\cite{liu2022general} and~\cite{grazzi2020iteration}.

	\begin{assumption}\label{assum_stationary ell}
		$\Omega$ is a compact set and $U = \R^n$. 
		$\mathtt{Fix}(\T(\cdot,\ome))$ is nonempty
		for any $\ome \in \Omega$. 
		$\ell(\u,\ome)$ is twice continuously differentiable on $\R^n \times \Omega$. 
		For any $\ome \in \Omega$, $\ell(\cdot,\ome) : \R^n \rightarrow \R$ is $L_\ell$-smooth, convex and bounded below by $M_0$.
	\end{assumption}

	For $\D(\cdot,\ome)$ we request a stronger assumption than Assumption~\ref{assum_T}
	that $\D(\cdot,\ome)$ is contractive with respect to $\| \cdot \|_{\H_{\ome}}$ throughout this part, to guarantee the uniqueness of the fixed point. 
	\begin{assumption} \label{assum_stationary D}
		There exist $\H_{ub} \succeq \H_{lb} \succ 0$, %
		such that for each $\ome \in \Omega$, there exists $\H_{ub} \succeq \H_{\ome} \succeq \H_{lb}$ such that
		\begin{itemize}
			\item[(1)] $\D(\cdot,\ome)$ is contractive with respect to $\| \cdot \|_{\H_{\ome}}$, i.e., there exists $\bar{\rho} \in (0,1)$, such that for all $(\u_1,\u_2) \in \R^n \times \R^n$,
			\begin{equation*}
				\|\D(\u_1,\ome) - \D(\u_2,\ome) \|_{\H_{\ome}} \le \bar{\rho} \| \u_1 - \u_2\|_{\H_{\ome}}.
			\end{equation*}
			
			\item[(2)] $\D(\cdot,\ome)$ is closed. %
		\end{itemize}
	\end{assumption}

	Denote $\hat{\S}(\ome):= \mathrm{argmin}_{\u \in \mathtt{Fix}(\T(\cdot, \ome)) \cap U } \ell (\u,\ome)$,
	and we have the following stationary analysis results.

	\begin{thm}\label{theorem stationary}
		Suppose Assumptions~\ref{assum_stationary ell} and~\ref{assum_stationary D} are satisfied, 
		$\frac{\partial}{\partial \u} \T(\u,\ome)$ and $\frac{\partial}{\partial \ome} \T(\u,\ome)$ are Lipschitz continuous with respect to $\u$,
		and $\hat{\S}(\ome)$ is nonempty for all $\ome \in \Omega$. 
		Let $\{\u^k(\ome)\}$ be the sequence generated by Eq.~\eqref{eq:simple_bilevel_alg} with $\mu \in (0,1)$ and $s_k = \frac{s}{k+1}$, 
		where $s \in (0, \frac{\lambda_{\min}(\H_{lb})}{L_{\ell}} )$.
		\begin{itemize}
			\item[(1)]
			We have 
			\begin{equation*}
				\sup_{\ome \in \Omega} \left\| \nabla \varphi_k(\ome) - \nabla \varphi(\ome) \right\|_\H \rightarrow 0,\ \text{as}\ k \rightarrow \infty.
			\end{equation*}
			
			\item[(2)]
			Let $\ome^K$ be an $\varepsilon_K$-stationary point of $\varphi_{K}(\ome)$, i.e., 
			\begin{equation*}
				\varepsilon_K = \nabla \varphi_K(\ome^K).
			\end{equation*}
			Then if $\varepsilon_K \rightarrow 0$, we have that any limit point $\bar{\ome}$ of the sequence $\{\ome^K\}$ is a stationary point of $\varphi$, i.e., 
			\begin{equation*}
				0 = \nabla \varphi(\bar{\ome}).
			\end{equation*}
			
		\end{itemize}

	\end{thm}

	For detailed proofs of the above results,	
	please refer to our conference version in~\cite{liu2022optimization}. %

	\subsection{Convergence of HODL without Aggregation (sHODL)} \label{sec: theo: sHODL}
	
	In Section~\ref{sec: theo: approximation quality} and~\ref{sec: theo: stationary}, we discuss the convergence properties (approximation quality and stationary analysis) of solution strategy aHODL (using Eq.~\eqref{eq:simple_bilevel_alg} to compute $\u^K$). Now we further extend these convergence properties to the solution strategy without aggregation sHODL introduced in Section~\ref{sec: solution strategy} (using Eq.~\eqref{eq:sHODL} to compute $\u^K$).

	On the approximation quality, 
	based on Assumptions~\ref{assum_F} and~\ref{assum_T},
	under the further assumptions that $\ell(\cdot,\ome)$ is uniformly Lipschitz continuous and $\T(\cdot,\ome)$ has a unique fixed point, 
	the approximation quality result for the solution strategy without aggregation can be obtained.
	For detailed discussions 
	please refer to~\cite{franceschi2018bilevel,liu2020generic}.	
	Note that for the convergence guarantee, compared with aHODL, the simplified solution strategy sHODL reduces the computational burden but requires a stronger assumption that the operator $\T$ is contractive, i.e., the set $\mathtt{Fix}(\T(\cdot,\ome))$ is a singleton.
	Also note that in this situation the convexity of $\ell$ is not required.
	Corresponding to those classic gradient-based unrolling algorithms without linear constraints, they require that the objective function in Eq.~\eqref{eq:min f} is strongly convex~\cite{pedregosa2016hyperparameter,franceschi2018bilevel}.
	If the solutions to the optimization process are not unique (such as $f$ is only convex, i.e.,  the corresponding operator is only non-expansive), and substituted to the learning process directly, %
	then the obtained solution may be far away from the true solution of the original bilevel problem. Please refer to the counter-example in~\cite{liu2020generic}.
	However, using the solution strategy with aggregation aHODL (using Eq.~\eqref{eq:simple_bilevel_alg} to compute $\u^K$) which aggregates the upper and lower iterative directions~$\v_{l}$ and $\v_{u}$,
	then even if the fixed points are not unique (the lower iterative operator is merely non-expansive), we can still approach the true solution with joint convergence.

	On the stationary analysis, 
	please note that our stationary analysis in Section~\ref{sec: theo: stationary} is also a unified convergence analysis of our solution strategies with and without aggregation (aHODL and sHODL),
	so it is applicable to all kinds of hierarchical problems.
	Specifically, $\mu$ in aHODL (using Eq.~\eqref{eq:simple_bilevel_alg} to compute $\u^K$) is taken to be between 0 and 1, while in the solution strategy without aggregation sHODL (using the simplified form Eq.~\eqref{eq:sHODL} to compute $\u^K$), it is taken to be 0.
	Taking $\mu = 0$, Theorem~\ref{theorem stationary} also holds, and the proofs parallel.
    Please also refer to~\cite{pedregosa2016hyperparameter} for the stationary analysis of the classic gradient-based unrolling algorithms as the special case of our solution strategy without aggregation.
    The discussions above for the convergence properties of HODL without aggregation  (sHODL) can be concluded in the following proposition.

	\begin{proposition} %
	    Suppose $\{\u^k(\ome)\}$ to be the sequence generated by sHODL in Section~\ref{sec: solution strategy}.
	    \begin{itemize}
		\item[(1)]
	    Suppose Assumptions~\ref{assum_F} and~\ref{assum_T} are satisfied, $\ell(\cdot,\ome)$ is uniformly Lipschitz continuous and $\T(\cdot,\ome)$ has a unique fixed point.
		Then, let $\ome^K \in \mathrm{argmin}_{\ome \in\Omega}\varphi_{K}(\ome)$, and we have any limit point $(\bar{\u},\bar{\ome})$ of the sequence $\{(\u^K(\ome^K),\ome^K) \}$ is a solution to the problem in Eq.~\eqref{eq:bilevel_fix}, i.e., $\bar{\ome}\in\mathrm{argmin}_{\ome\in\Omega}\varphi(\ome)$ and $\bar{\u} = \T(\bar{\u},\bar{\ome}) $.
		Further, $\inf_{\ome \in \Omega}\varphi_K(\ome) \rightarrow \inf_{\ome \in \Omega} \varphi(\ome)$ as $K \rightarrow \infty$.

		\item[(2)]
		Suppose Assumptions~\ref{assum_stationary ell} and~\ref{assum_stationary D} are satisfied, 
		$\frac{\partial}{\partial \u} \T(\u,\ome)$ and $\frac{\partial}{\partial \ome} \T(\u,\ome)$ are Lipschitz continuous with respect to $\u$,
		and $\hat{\S}(\ome)$ is nonempty for all $\ome \in \Omega$. 
		Let $\ome^K$ be an $\varepsilon_K$-stationary point of $\varphi_{K}(\ome)$, i.e., 
		$
			\varepsilon_K = \nabla \varphi_K(\ome^K).
		$
		Then if $\varepsilon_K \rightarrow 0$, we have that any limit point $\bar{\ome}$ of the sequence $\{\ome^K\}$ is a stationary point of $\varphi$, i.e., 
		$
			0 = \nabla \varphi(\bar{\ome}).
		$
		\end{itemize}
	\end{proposition}

\section{Applications} \label{sec:application}

In this section, we first compare HODL with other established ODL methods in detail, and then demonstrate the applications of HODL in solving practical problems of various forms and the specific settings under these forms.
Summary of operators $\D_\num$ and $\D_\net$ for problems of various forms and corresponding applications
is shown in Table~\ref{tab:appendix summary of D},
where the applications for other learning tasks regarded as hierarchical models will be discussed in Section~\ref{sec:extension}.

\begin{table*}[!htbp]\small 
	\centering
	\caption{Summary of operator $\D_\num$, $\D_\net$, and applications for various models. Here NE-net denotes Non-Expansive networks ($1$-Lipschitz continuous, with respect to $\| \cdot \|_{\H_{\ome}}$), and GD is short of gradient descent. 
	\textcolor{black}{Note that here N/A means that $\D_\mathtt{num}$ or $\D_\mathtt{net}$ is not employed for the corresponding tasks following the common settings. }}
	\label{tab:appendix summary of D}
	\renewcommand\arraystretch{1.2}
	\setlength{\tabcolsep}{1mm}{
		\renewcommand\arraystretch{1.1}
		\resizebox{\textwidth}{!}{
			\begin{tabular}{c|c|c|c}
				\hline
				Model& $\D_{\mathtt{num}}$&$\D_{\mathtt{net}}$&Applications\\
				\hline
				{\begin{tabular}{c}
						{Constrained}\\
						{Problems}\\
				\end{tabular}}&${\mathtt{ALM}}:\left\{\	
				\begin{aligned}
					\u^{k+1} &= \underset{\u}{\mathrm{argmin}}~\left\{ f(\u) + \langle \lamm^k, \A(\ome)\u - \yy(\ome) \rangle + \frac{\beta}{2}\|\A(\ome)\u - \yy (\ome)\|^2 + \frac{1}{2}\|\u- \u^k\|_{\H_{\ome}}^2\right\} \\
					\lamm^{k+1} & = \lamm^k + \beta( \A(\ome)\u^{k+1} - \yy (\ome))
				\end{aligned} \right.
				$&\multirow{4}{*}{NE-net}&\begin{tabular}{c}
					{Sparse Coding}\\
					{Rain Streak Removal}\\
				\end{tabular}
				\\
				\cline{1-2}
				\cline{4-4}

				\multirow{2}{*}{Regularized}&	\multirow{3}{*}{${\mathtt{PG}}:\u^{k+1} = \underset{\u}{\mathrm{argmin}}~\left\{ f(\u^k) + \langle \nabla_\u f(\u^k) ,\u - \u^k \rangle + g(\u,\ome) +  \frac{1}{2\gamma}\|\u- \u^k\|^2_{\H_{\ome}} \right\}$} &&Sparse Coding\\
				\multirow{2}{*}{Problems}&&&Image Deconvolution\\
				&&&{Low-light Enhancement}\\

				\hline
				\multirow{2}{*}{Hierarchical}&\multirow{2}{*}{ ${\mathtt{GD}}:\u^{k+1}=\u^{k}-\nabla_u f(\u^k)$ }&\multirow{2}{*}{N/A}&Hyper-parameter Optimization\\
				\multirow{2}{*}{Models}&&&Few-shot Learning\\
				\cline{2-4}
				&N/A&NE-net&Adversarial Learning\\
				\hline
		\end{tabular}		}
	}
\end{table*}

\subsection{Comparison with Existing ODL Methods}

Compared with existing ODL methods, HODL additionally considers the optimal update of learning variables $\ome$, thus providing better theoretical guarantees and higher application value. 
Existing ODL methods only focus on the output of optimization model, i.e., the final iterative results of the optimization variables $\u$. 
Usually, their selection of learning variables $\ome$ is just a direct extraction of network modules from similar learning tasks~\cite{2017ISTA,yang2017admm}. %
Hence, this selection method ignores the convergence of learning variables $\ome$ and can be considered as the optimization strategy of random search for similar learning tasks in the search space of learning variables~$\ome$. 
On the contrary, HODL focuses on the iterative results of both optimization variables $\u$ and learning variables $\ome$, and performs gradient descent on learning variables $\ome$, thus providing sufficient theoretical guarantees and clear application framework. 
In a word, compared with existing ODL methods, HODL makes up for the weakness of ODL in theory and upgrades from random search to gradient descent for application, providing the theoretical guarantee and usability of ODL models that existing methods cannot achieve.
Under the HODL framework, the difference among algorithms for various applications lies in the operator $\D$ introduced in Section~\ref{sec:GKM}.
Next we introduce the specific forms of $\D$ in these applications.

\subsection{Application for Sparse Coding} \label{sec:constrained HODL} 
Taking sparse coding as an example, we first describe how HODL can be applied to constrained and regularized problems and show how the coupling between the optimization model and optimization variables can be handled. 
Specifically, the sparse coding task is dedicated to representing given data $\b$ as a sparse coefficient representation $\u$ of a set of basis vectors $\Q$, i.e., $\Q\u=\b$. 
As the basis vectors in the transform matrix $\Q$ are usually overcomplete, we introduce additional sparsity criterion to address the degeneracy problem caused by overcompleteness.
Depending on how to force the algorithm to provide a satisfactory representation of $\b$, sparse coding can be considered as a constrained or regularized problem.
Note that in both cases, usually we set the objective function $\ell$ in Eq.~\eqref{eq:bilevel_fix} to be the MSE loss.

\textbf{Constrained Sparse Coding.} 
The constrained sparse coding form is based on linear equality constraints $\Q\u=\b$, corresponding to the constraint term in Eq.~\eqref{eq:min f} as a guarantee of reconfigurability. 
As the reconstruction is usually imperfect, 
Since the transform matrix $\Q$ is usually generated from clear data,
noise in the given data $\b$ cannot be perfectly restored, 
so the noise estimation term $\u_n$ is added as a complement to adhere to the task information, i.e., $\Q\u+\u_n=\b$. 
Note that here we need to additionally estimate the noise term $\u_n$, and in other cases if the noise is a constant vector, we just denote it to be $\n$. 
As an overcomplete task, %
the $\ell_1$ paradigm is usually used as a sparsity penalty which forces our representation of $\u$ and~$\u_n$ to be sparse. 
We model the constrained sparse coding problem as the following
\begin{equation}\label{eq:sparse coding_C}    \min_{\u,\u_n}\kappa\Vert\u\Vert_1+\Vert\u_n\Vert_1\quad \text{s.t. }\quad\Q\u+\u_n=\b
\end{equation}%
where $\kappa$ is a scaling constant to determine the relative importance of the two norms. 
In order to solve the constrained optimization problem while satisfying the assumptions of HODL, we use the ALM method to determine $\D_{\mathtt{ALM}}$ as $\D_\num$ as shown in Table~\ref{tab:appendix summary of D}. 
It can be proved that corresponding $\D_{\mathtt{ALM}}$ for Eq.~\eqref{eq:sparse coding_C} satisfies Assumption~\ref{assum_T} under mild conditions. 
Please refer to~\cite[Appendix B]{liu2022optimization} for details.

\textbf{Regularized Sparse Coding.}
Another common type of task prior is to add regularization terms as the learnable module in Eq.~\eqref{eq:min f} to the objective function. The regularized sparse coding form is based on  reconstruction term $\Vert\Q\u-\b\Vert_2$ as a guarantee of reconfigurability.  As an overcomplete task regularization, %
it also uses~$\ell_1$ paradigm as a sparsity penalty to force the representation of $\u$ to be sparse. 
We define the objective function for regularized sparse coding as
\begin{equation}\label{eq:sparse coding_R}
    \min_{\u}\Vert\Q\u-\b\Vert_2+\kappa\Vert\u\Vert_1
\end{equation} 
where $\kappa$ is a scaling constant to determine the relative importance between reconstruction term and regularization term. In order to solve the regularized optimization problem while satisfying the assumptions of HODL, we use the PG method to determine $\D_{\mathtt{PG}}$ as $\D_\num$ as shown in Table~\ref{tab:appendix summary of D}.
In~\cite[Appendix B]{liu2022optimization}, it is proved that corresponding $\D_{\mathtt{PG}}$ satisfies Assumption~\ref{assum_T} under mild conditions.

\textbf{Composition of $\D_{\num}$ and $\D_{\net}$.}
In the above discussion we use a fully connected layer network with ReLU activation and spectral normalization as $\D_\net$.
When compositing $\D_{\num}$ and $\D_{\net}$ for better performance,
we use a non-expansive $\D_{\net}$ as shown in Table~\ref{tab:appendix summary of D} and composite them to satisfy the assumptions of HODL solution strategy.

The convergence guarantee will hold when compositing $\D_{\num}$ and $\D_{\net}$,
because if $\D_{\num}$ and $\D_{\net}$ satisfy Assumption~\ref{assum_T}(or~\ref{assum_stationary D}) with the same $\H_{\ome}$,
then $\D_{\num} \circ \D_{\net}$ also satisfies these assumptions.
To be specific, the non-expansive (or contractive) property of $\D_{\num} \circ \D_{\net}$ with $\H_{\ome}$ can be easily verified from the definition.
As for the closeness of $\D_{\num}( \cdot, \ome) \circ \D_{\net}( \cdot, \ome)$ for a fixed $\ome \in \Omega$, we consider the sequence $\{(\u^k,\v^k)\} \in \mathrm{gph} (\D_{\num}( \cdot, \ome) \circ \D_{\net}( \cdot, \ome))$ satisfying $(\u^k,\v^k) \rightarrow (\bar{\u}, \bar{\v})$.
From the boundedness of $\{\u^k\}$ and the non-expansive (or contractive) property of $\D_{\num} \circ \D_{\net}$ with $\H_{\ome} \succ 0$ ,
it can be obtained that $\D_{\net}( \u^k, \ome)$ is bounded, 
so there exists a subsequence 
$\{(\u^i,\v^i)\} \subseteq \{(\u^k,\v^k)\}$ such that $\D_{\net}( \u^i, \ome) \rightarrow \bar{\ome}$. 
Then it follows from the closeness of $\D_{\net}( \cdot, \ome)$ and $\D_{\num}( \cdot, \ome)$
that $(\bar{\u}, \bar{\ome}) \in \mathrm{gph} \D_{\net}( \cdot, \ome)$ and $(\bar{\ome}, \bar{\v}) \in \mathrm{gph} \D_{\num}( \cdot, \ome) $.
Hence, $(\bar{\u}, \bar{\v}) \in \mathrm{gph} (\D_{\num}( \cdot, \ome) \circ \D_{\net}( \cdot, \ome))$. 
Note that given any non-expansive $\D_{\net}$  (which can be achieved by spectral normalization) and positive-definite matrix $\H_{\ome}$, by setting  $\D_{\mathtt{net^*}}=\H_{\ome}^{-1/2}\D_{\net}\H_{\ome}^{1/2}$, then $\D_{\mathtt{net^*}}$  satisfies Assumption~\ref{assum_T}(or~\ref{assum_stationary D}) with $\H_{\ome}$.

\subsection{Applications for Vision Tasks}
In this subsection, we illustrate the applications of ODL in vision tasks, describe the shortcomings of existing ODL methods, and demonstrate how to apply HODL in vision tasks. 
In these applications, we use $\D_{\mathtt{ALM}}$ and $\D_{\mathtt{PG}}$ as $\D_{\mathtt{num}}$ for constrained and regularized problems, respectively, consistent with the discussion for sparse coding.
\textcolor{black}{In addition, we set the objective function $\ell$ in Eq.~\eqref{eq:bilevel_fix} to be the MSE loss.}

\textbf{Rain Streak Removal}. 
An application scenario of constrained HODL requires using variable separation to aid in problem solving. 
As an example, in the rain streak removal task, the sparse solutions of rain line and background are solved separately by adding auxiliary variables~\cite{liu2021investigating}. 
This scenario requires the auxiliary variables and the original variables to be kept equal, and it is suitable to use HODL framework with equality constraints. 
Specifically, given the input rainy image $\Ir_r$, the goal is to decompose it into a rain-free background $\u_b$ and a rain streak layer $\u_r$, i.e., $\Ir_r=\u_b + \u_r$, 
to enhance the visibility. 
The problem can be reformulated as 
$
	\min\limits_{\u_b,\u_r}\frac{1}{2}\Vert\u_b+\u_r-\Ir_r\Vert_2^2+\psi_b(\u_b)+\psi_r(\u_r),
$
where $\psi_b(\u_b)$ and $\psi_r(\u_r)$ are set to be $\psi_b(\u_b)=\kappa_b\Vert\u_b\Vert_1$ and $\psi_r(\u_r)=\kappa_r\Vert\nabla\u_r\Vert_1$,
representing the priors on the background layer and rain streak layer respectively.
Then we introduce auxiliary variables $\v_b$ and $\v_r$, and transfer the problem to be $ \min\limits_{\u_b,\u_r,\v_b,\v_r} \frac{1}{2}\Vert\u_b+\u_r-\b\Vert_2^2+\kappa_b\Vert\v_b\Vert_1+\kappa_r\Vert\v_r\Vert_1$, s.t.,  $\v_b=\u_b,\v_r=\nabla\u_r$,
where $\nabla=\left[ \nabla_h;\nabla_v\right] $ denotes the gradient in horizontal and vertical directions.
Existing ODL methods usually solve $\u_b, \u_r$ using $\D_{\mathtt{num}}$ and solve $\v_b, \v_r$ by a pre-trained $\D_{\mathtt{net}}$, usually leading to a gap between the pre-trained task and current task. 
HODL, in contrast, ensures that $\D_{\mathtt{net}}$ learns valid rain streak information by using a regularized $\D_{\mathtt{net}}$ trained on current task jointly with~$\D_{\mathtt{num}}$.

\textbf{Image Deconvolution}. 
As an application of regularized HODL, 
image deconvolution does not strive for perfect image restoration, but pursues a balance between restoration and deconvolution effects whenever possible~\cite{chen2018theoretical}. 
Specifically, the input image can be expressed as $\b = \Q \ast\u + \n$, where $\Q ,\u$, and $\n$ respectively denote the blur kernel, latent clean image, and additional noise, and~$\ast$ denotes the two-dimensional convolution operator. %
Here the regularization is implemented based on Maximum A Posteriori (MAP) estimation. %
Then the problem is transferred to $\min_{\u\in U}\Vert\Q \ast \u-\b\Vert^2_2+g(\u)$, 
where $g(\u)$ is the prior function of the image. %
We set $g(\u)$ to be $\kappa\Vert \W\u\Vert_1$, where $\W$ is the wavelet transform matrix,
considering that there is usually a sparse image after the wavelet transform.
In this task, existing ODL approaches typically have two ideas.
One uses $\D_{\mathtt{num}}$ for task fidelity term $\Vert\Q \ast \u-\b\Vert^2_2$ and pre-trained $\D_{\mathtt{net}}$  for regularization term $g(\u)$ to guarantee clarity. 
Similar to the previous task, this makes the pre-trained~$\D_{\mathtt{net}}$ not well adapted to the current task details such as convolution kernels and object edges. 
The other is to train $\D_{\mathtt{net}}$ in the current task, but ignore $\D_{\mathtt{num}}$ during training after having built $\D_{\mathtt{net}}$ from $\D_{\mathtt{num}}$. HODL, on the other hand, ensures $\D_{\mathtt{num}}$ to control over the iteration and enables $\D_{\mathtt{net}}$ to adapt to the current task through joint training. %

\textbf{Low-light Enhancement}. 
As another application of regularized HODL, 
low-light enhancement usually employs a complex network to estimate illumination in order for a higher image quality. 
Hence, compared with linear equality constraint terms, it is more appropriate to use regularization terms as a priori.
Specifically, we follow the simple Retinex rule $\mathbf{y}=\mathbf{x}\otimes\mathbf{u}$, where $\mathbf{y}$ is the captured underexposed observation which is a given low-light image, $\mathbf{x}$ is the desired recovery,
$\u$ is the illumination to be determined for enhancement,
and the operator $\otimes$ denotes element-wise multiplication.
To accurately estimate $\u$, inspired by the work in~\cite{liu2021retinex}, we estimate $\mathbf{u}$ by 
$
\min\limits_{\u}\Vert\u-\phi(\mathbf{y})\Vert^2_2+\psi(\u),
$
where $\phi$ is a given estimated illumination mapping,
and $\psi$ is a regularization function estimated implicitly from a CNN.
In this task, existing ODL methods usually construct the network %
for solving task term $\Vert\u-\phi(\mathbf{y})\Vert^2_2$ and regularization term $\psi(\u)$ 
from an optimization problem,
but along with the training procedure, the network structure will be away from the original optimization structure.
However, HODL is able to retain the optimization structure in training, thus effectively improving the image fidelity.

\section{Extensions to Other Learning Tasks}\label{sec:extension}

In this section we illustrate how to apply the hierarchical modeling of HODL to a wide range of learning tasks beyond ODL.
Specifically, as a methodology, HODL framework with hierarchical structures is not limited to specific methods and can be used to uncover the hierarchical relationships in multi-task coupled learning tasks as well. 
Since learning tasks can be considered as optimization problems based on loss functions and specific optimizers, HODL, which is dedicated to modeling hierarchical relationships between optimization and learning, can also accommodate hierarchical coupling in multiple learning tasks.
For example, by setting the optimization operator $\T$ in Eq.~\eqref{eq:bilevel_fix} as the gradient descent operator for optimizing the sub-task loss function, and $\ell$ in Eq.~\eqref{eq:bilevel_fix} as the loss function for another sub-task, HODL can be easily migrated to any learning application with multiple sub-tasks. 
\textcolor{black}{
Actually, bilevel optimization can be regarded as a special case of HODL framework, if we restrict~$\T$ to be the operators for solving optimization problems.
More specifically,
for adversarial learning, $\ell$ is used to characterize the antagonistic relationship between the generator and discriminator;
for hyper-parameter optimization and few-shot learning, $\ell$ is the cross entropy loss function on the validation set.
Please also refer to~\cite{liu2021value} for more detailed expression of $\ell$ and $\T$.
}
Therefore, HODL can be widely applied in adversarial learning~\cite{metz2016unrolled,pfau2016connecting}, hyper-parameter optimization~\cite{liu2021value,okuno2018hyperparameter}, few-shot learning~\cite{franceschi2018bilevel}, and so on, as shown in Table~\ref{tab:appendix summary of D}.

\textbf{Adversarial Learning.}
As the best-known application of adversarial learning, Generative Adversarial Networks (GAN) has received much attention in recent years, which adversarially trains generators to solve real-world tasks by means of additional discriminators.
In GAN, the generator depends on the discrimination from the discriminator to learn the features, while the discriminator depends on the output of generator to learn the classification. 
Therefore, by taking the update of discriminator as the operator $\T$ in Eq.~\eqref{eq:bilevel_fix} and the learning process of generator as $\ell$ in Eq.~\eqref{eq:bilevel_fix}, HODL can effectively model the coupling relationship between the two sub-tasks of GAN.

\textbf{Hyper-parameter Optimization}. 
The increasing complexity of machine learning algorithms has driven plenty of research in the field of hyper-parameter optimization. 
In machine learning, hyper-parameter optimization aims at choosing a set of optimal hyper-parameters for learning algorithms. 
Hyper-parameters are a class of parameters whose values are used to control the learning process. 
Therefore, by taking the learning process as the operator $\T$ in Eq.~\eqref{eq:bilevel_fix} and the objective function to choose optimal hyper-parameters as $\ell$ in Eq.~\eqref{eq:bilevel_fix}, our HODL approach is equally effective when dealing with hyper-parameter optimization.

\textbf{Few-shot Learning}. 
Few-shot learning ($N$-way $M$-shot) is a multi-task $N$-way classification which aims to learn the feature extraction structure with generalization ability, so that each new task can be solved only through $M$-training samples. 
This task has nested hierarchies, which respectively classify $M$ samples and learn a feature structure that can be used for new tasks. 
Therefore, by taking the classification optimization process as the operator $\T$ in Eq.~\eqref{eq:bilevel_fix} and the learning process of feature structure as function $\ell$ in Eq.~\eqref{eq:bilevel_fix}, our HODL approach can also be applied.

Besides, in some applications, the operator $\T$ corresponds to optimizing an implicit energy function that is solved indirectly through a neural network. 
In this case, by applying spectral normalization to the network, we can still obtain a non-expansive mapping. 
We verify the necessity of the non-expansive property of neural network in Section~\ref{sec:experiment toy}.

\begin{figure}
	\centering
	\includegraphics[height=3cm,width=4cm]{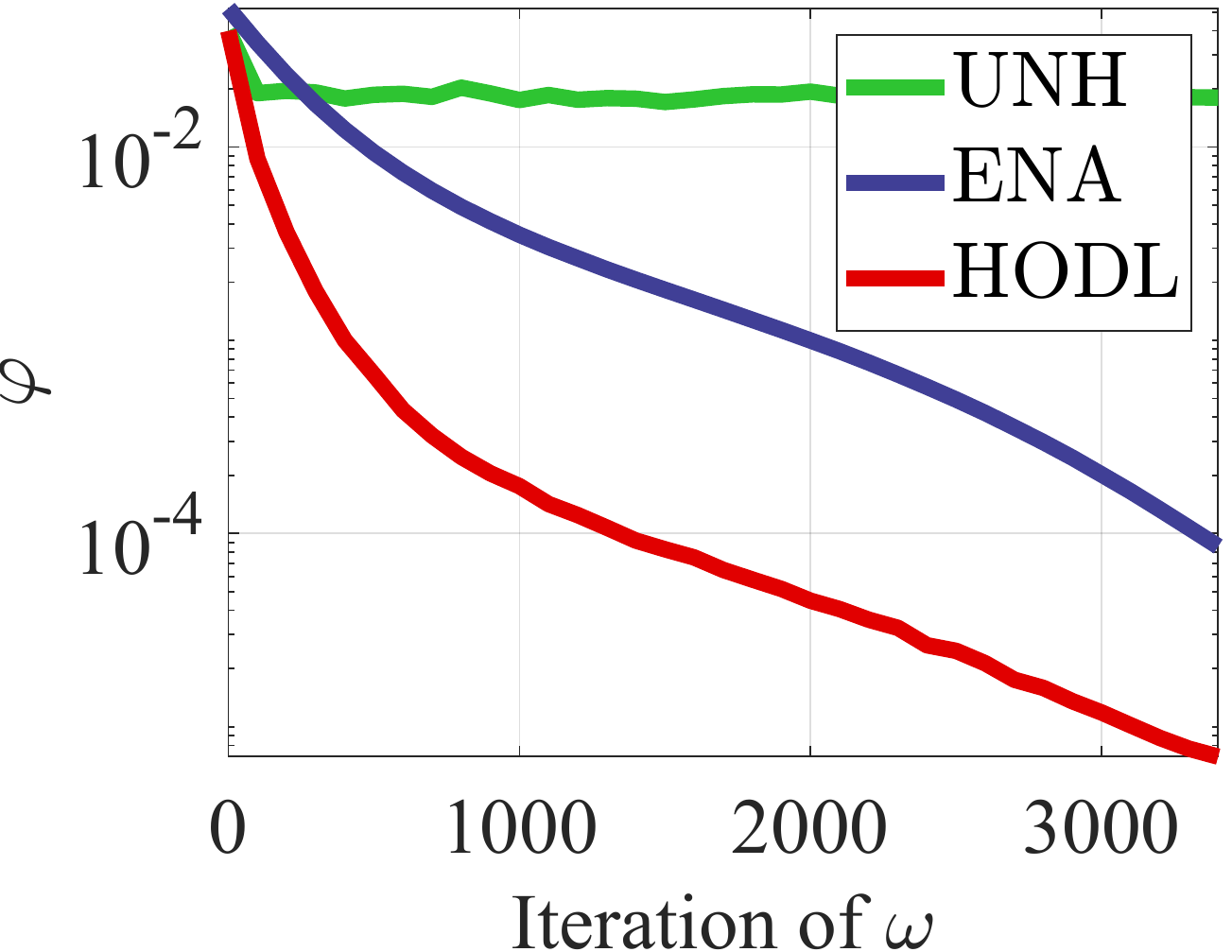}
	\includegraphics[height=3cm,width=4cm]{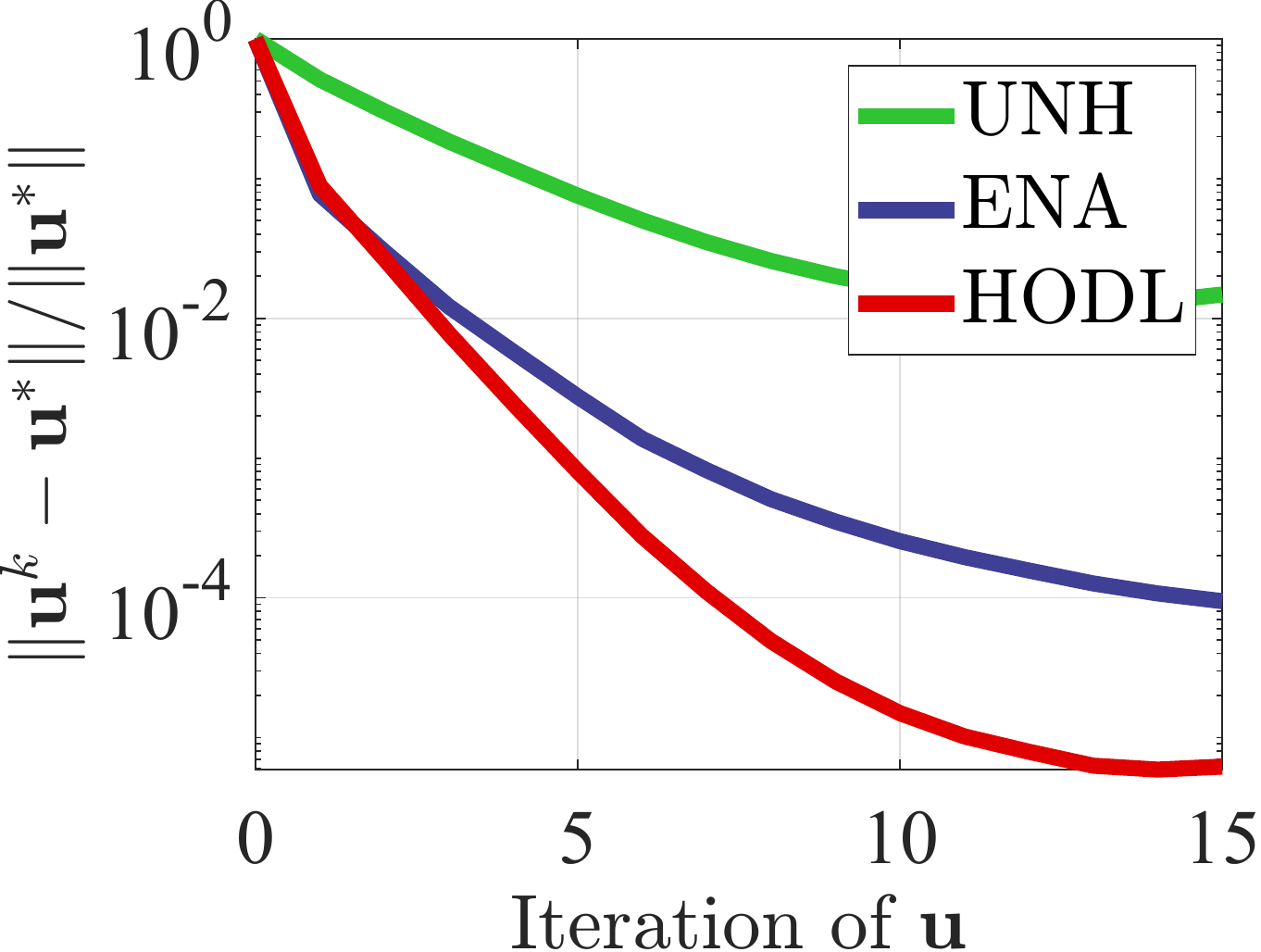}
	\caption{The convergence behavior of $\ome$ and $\u$ by UNH, ENA, and HODL for regularized sparse coding.
		It can be seen that for regularized problems using PG, our HODL has better convergence results.}
	\label{fig:convergence behaviorPG}
\end{figure}

\section{Experimental Results} \label{sec:experiment}

In this section, we first verify the theoretical properties of HODL on synthetic experiments in the sparse coding task. 
We subsequently apply HODL to visual experiments containing rain streak removal, image deconvolution, and low-light enhancement. 
Finally, we extend HODL to other applications with hierarchies, including adversarial learning, hyper-parameter optimization, and few-shot learning.
We conduct our experiments mainly on a PC with Intel Core i9-10900KF CPU (3.70GHz), 128GB RAM and two NVIDIA GeForce RTX 3090 24GB GPUs. 
\textcolor{black}{
All experiments are implemented on synthetic datasets,
and the Adam optimizer is adopted to update variable $\ome$.
Note that other acceleration techniques are also applicable under the HODL framework.
}

\subsection{Model Evaluation} \label{sec:experiment toy}

This part first verifies that HODL improves the overall performance compared with existing ODL methods. %
More specifically,
we analyze the performance on convergence by HODL in terms of learning variables and optimization variables for the learning process and optimization process, respectively. 
After that, we investigate some factors that may affect the performance of HODL.
To illustrate the generality of HODL, we verify the performance on constrained and regularized sparse coding problems.

For regularized problems, we use the regularized sparse coding model introduced in Section~\ref{sec:constrained HODL}.
We set $m = 500, n = 250$ ($\Q$ in Eq.~\eqref{eq:sparse coding_R} is a $ m\times n$ matrix), and the training and testing samples are 10000 and 1000, respectively. The elements of matrix $\Q$ are sampled from the standard Gaussian distribution, and the column vector of matrix $\Q$ is standardized to have the unit $\ell_2$ norm. 
The sparse vector~$\u$ is sampled from the standard Gaussian distribution, and the distribution of non-zero elements follows the Bernoulli distribution with probability~0.1. The intensity of noise $\n$ is 0.01 times the standard Gaussian distribution, and all data are generated by the model $\b=\Q\u+\n$. 
To be fair for the comparisons, $\Q$ and $\b$ are fixed in the experiment. 
We use the MSE loss as the supervised loss.
\textcolor{black}{Note that here $\D_\num$ is $\D_\mathtt{PG}$, and $\D_\net$ is a fully connect network with ReLU activation. 
	For comparison, UNH stands for the method that only learns the step size, while ENA stands for the method that learns~$\D_\net$.}
	To show the performance of HODL for regularized problems, we compare the convergence of different methods in the optimization process for $\u$ and learning process for~$\ome$ in Figure~\ref{fig:convergence behaviorPG}. 
	It can be seen that, HODL performs better in the convergence of optimization and learning than other~methods.

	\begin{table} 
		\centering
		\small
		\caption{PSNR and SSIM results for constrained sparse coding on Set14. 
			Best and second best results are marked in red and blue respectively.}
		\label{tab:sparse code table}
		\begin{tabular}{c|c|c|c} %
			\hline
			\multicolumn{1}{c}{Methods } & \multicolumn{1}{c}{Layers } & PSNR  & SSIM  \\
			\hline
			\multirow{2}{*}{UNH } & 5     & 10.47$\pm$2.36  & 0.41$\pm$0.14  \\
			\cline{2-4}          
			& 25    & 11.31$\pm$2.29  & 0.41$\pm$0.15  \\
			\hline
			\multirow{2}{*}{ENA} & 5     & 15.59$\pm$0.81  & 0.52$\pm$0.13  \\
			\cline{2-4}          & 25    & 15.64$\pm$0.87  & 0.52$\pm$0.13  \\
			\hline
			\multirow{2}{*}{HODL} & 5     & \textcolor{blue}{\textbf{18.82$\pm$1.59}}  & \textcolor{blue}{\textbf{0.63$\pm$0.16}}  \\
			\cline{2-4}          & 25    & \textcolor{red}{\textbf{18.98$\pm$2.53}}  & \textcolor{red}{\textbf{{0.65$\pm$0.15}}}  \\
			\hline
		\end{tabular}%
	\end{table}

	For constrained problems, we follow the setting in~\cite{chen2018theoretical}
	to use the classic Set14 dataset as experimental data, in which the salt-and-pepper noise is added to $10 \%$ pixels of each image. 
	The rectangle of each image is divided into non-overlapping patches of size $16 \times 16$. We use the patch dictionary method to learn a $256 \times 512$ dictionary $\Q$. %
	We set batch size $= 128$, training set size $= 10000$, and random seed $= 1126$. 
	The testing set size depends on the size of each image. 
	Because we conduct unsupervised single image training, we do not use the MSE loss between the clear picture and the generated picture, but instead use the same unsupervised loss as in~\cite{xie2019differentiable}.

	To show the performance of HODL for constrained sparse coding, we present the PSNR and SSIM results in Table~\ref{tab:sparse code table}.
    It can be seen that the performance of our HODL on both PSNR and SSIM is superior than UNH and ENA.
    This is because UNH can only train few learning variables (such as the step size) to maintain convergence, %
    and the neglect of the original optimization structure during training by ENA brings about a distance from the real fixed point model. %
    In contrary, thanks to the hybrid strategy to incorporate optimization and learning processes, %
    HODL allows more learning variables to improve the performance.
    Considering the consistent performance of constrained HODL and regularized HODL, for simplicity, we base our subsequent analysis on the constrained HODL. %

	\begin{figure}
		\centering
		\includegraphics[height=3cm,width=4cm]{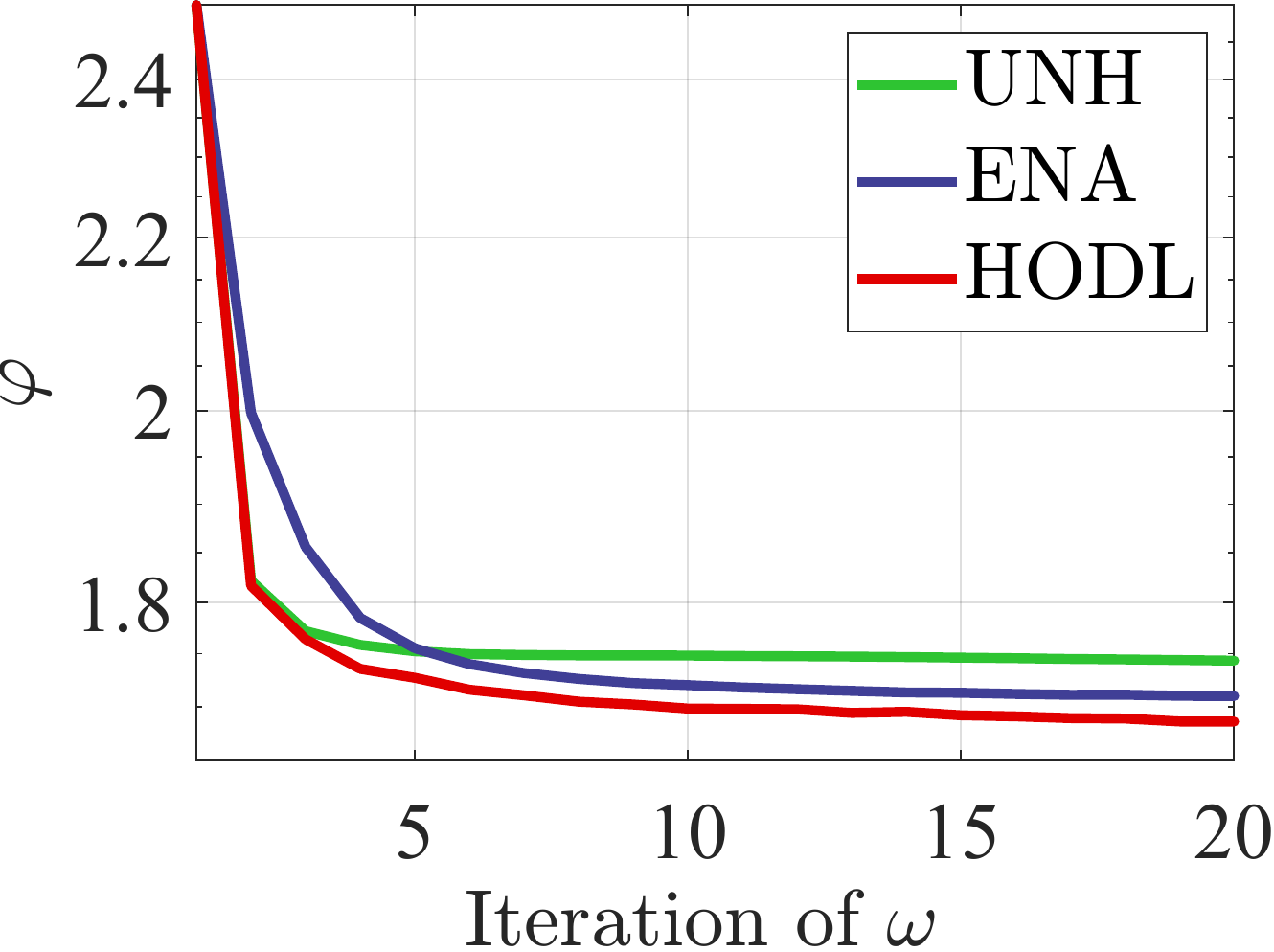}
		\includegraphics[height=3cm,width=4cm]{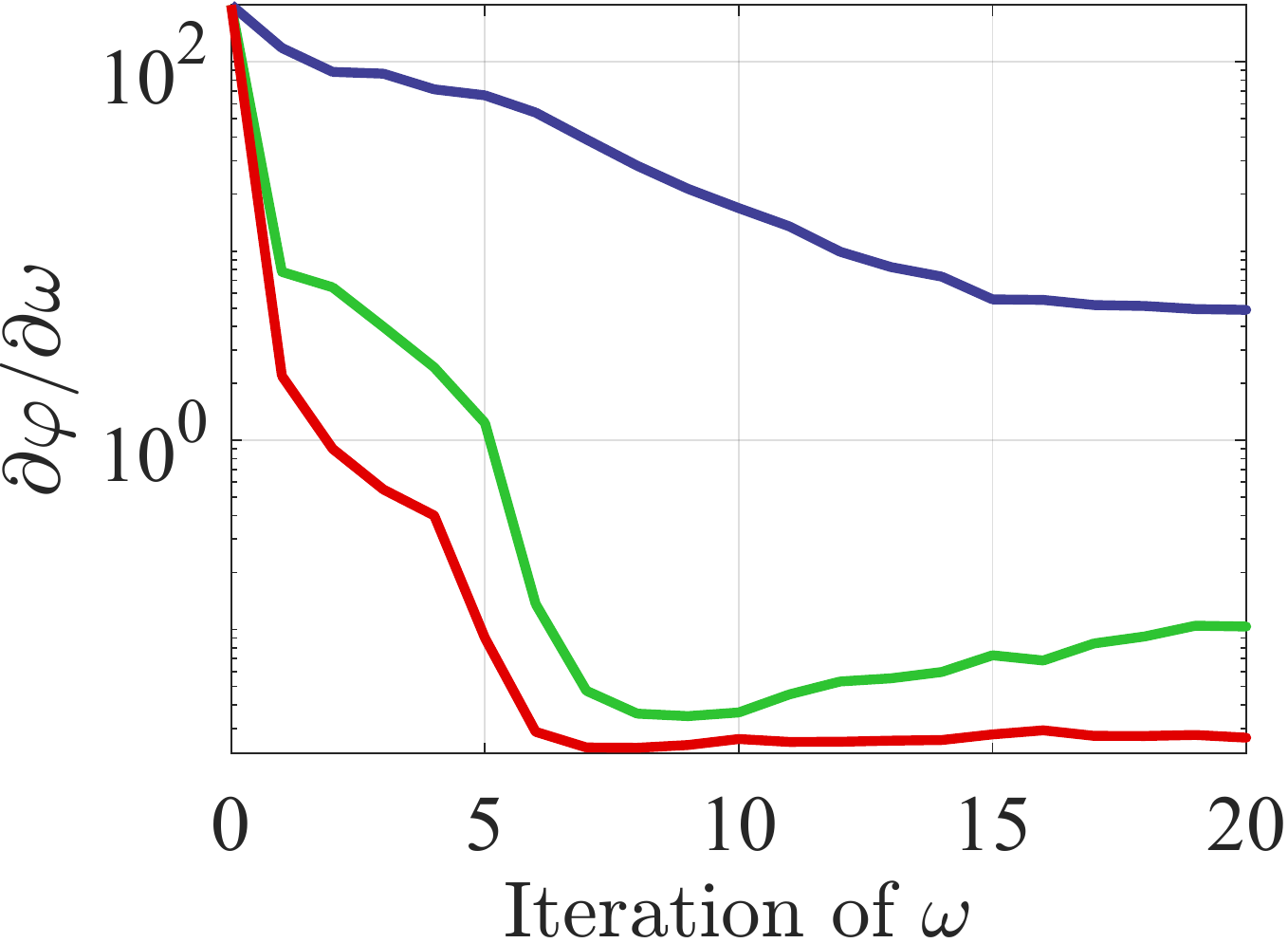}
		\caption{The convergence curves of $\varphi$ and ${\partial\varphi}/{\partial\ome}$ with respect to $\ome$ for constrained sparse coding. 
			UNH does not add learnable knowledge to optimization and ENA ignores the optimization structure during training. %
			It can be seen that our method achieves the optimal convergence of loss function with a stationary gradient curve.
		}
		\label{fig:convergence behavior}
	\end{figure}

To illustrate in detail how HODL improves the performance of ODL, we next analyze the convergence of learning variables $\ome$ and optimization variables $\u$, respectively.
In Figure~\ref{fig:convergence behavior}, we first analyze the convergence behavior of learning variables $\ome$ in the %
objective function of learning process $\varphi_K(\ome)= \ell(\u^K(\ome),\ome)$ defined in Eq.~\eqref{eq:phiK_def} with a fixed~$K$.
ENA and UNH perform poorly in the convergence of learning objective function, while HODL is able to effectively obtain better convergence.

	\begin{figure}
		\centering
		\begin{subfigure}[t]{0.23\textwidth}{
				\includegraphics[height=3cm,width=4cm]{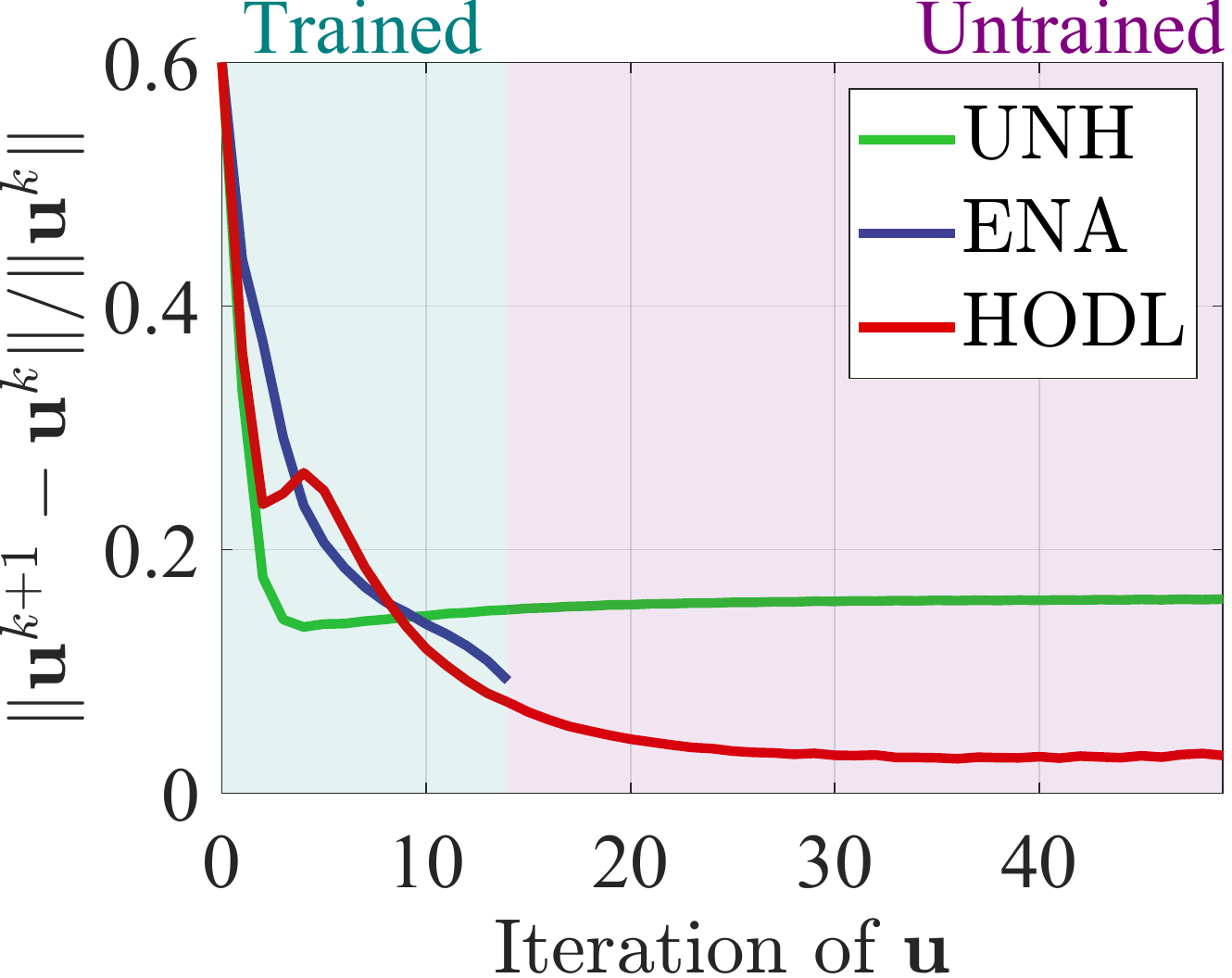}
				\subcaption{$K=15$ in training}
				\label{sub:15_ext}}
		\end{subfigure}
		\begin{subfigure}[t]{0.23\textwidth}{
				\includegraphics[height=3cm,width=4cm]{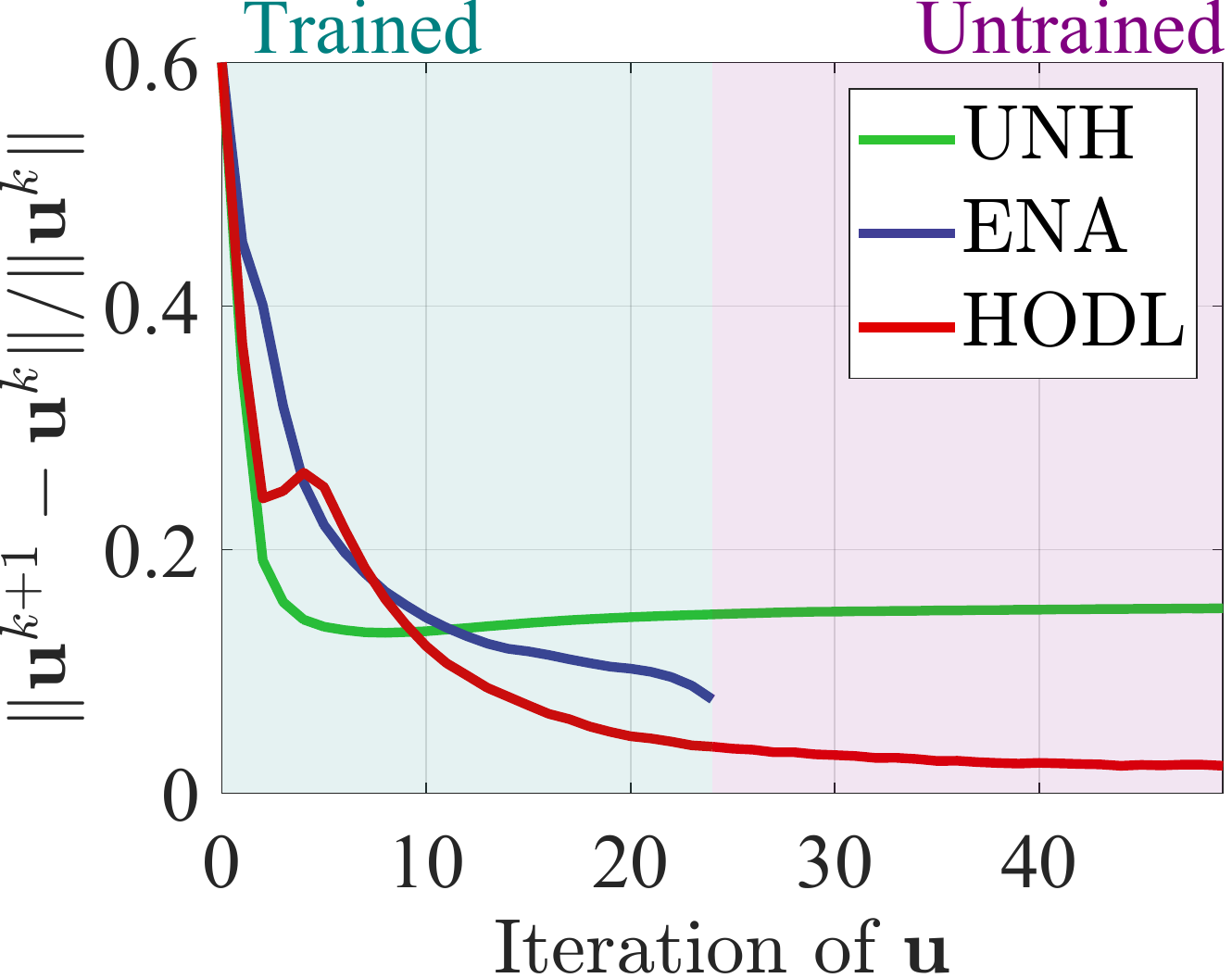}
				\subcaption{$K=25$ in training}
				\label{sub:25_ext}}
		\end{subfigure}
		
		\caption{\textcolor{black}{
			The convergence curves of $\Vert\u^{k+1}-\u^{k}\Vert/\Vert\u^{k}\Vert$ with respect to $\u$ 
			after  (\subref{sub:15_ext}) $K=15$ and  (\subref{sub:25_ext}) $K=25$ as iterations of~$\u$ in training, %
			where $k$ is the number of iterations of $\u$ for optimization in testing.
			The green background indicates when the training iteration of $\u$ is less than testing  iteration, while the pink background represents the testing iteration is beyond training iteration.
			It can be seen that our method can successfully learn the non-expansive mapping and converge better after different iterations in training. 			
			}
		}
		\label{fig:inner loop_ext}
	\end{figure}

Next, we verify the convergence of optimization variables~$\u$. %
\textcolor{black}{
On one hand, from the green background part of Figure~\ref{fig:inner loop_ext},
it can be seen that HODL outperforms other methods in convergence stability and convergence speed. 
UNH converges fast at first, 
but it cannot further improve the convergence performance. 
ENA has slow convergence speed 
because its neglect of optimization structure during training.
Conversely, HODL can effectively reduce the required number of iterations in training to control the expected error, 
which increases the computational efficiency.
On the other hand,
in practical applications, limited by the high computational burden on training time, one tends to train in a smaller number of optimization iterations and subsequently expects to obtain higher performance in testing. %
This requires ODL methods to be able to learn a stable non-expansive mapping. 
Therefore, we additionally observe the convergence curves of the optimization variables $\u$ in this case to further verify the stability and non-expansive property of the trained optimization iterative module. %
In Figure~\ref{fig:inner loop_ext}, 
we also show the convergence curve when the number of optimization iterations of $\u$ in testing is more than those in training (the pink background part).  %
Note that since for ENA the number of iterations of $\u$ is fixed in training, it cannot be compared in this case. %
Still, we find that HODL is superior to UNH,
and the mapping learned by our HODL can indeed continue to converge in the testing iterations beyond training steps, 
implying that we have effectively learned a non-expansive mapping with convergence. 
}

\begin{figure}
	\centering
	\includegraphics[height=3cm]{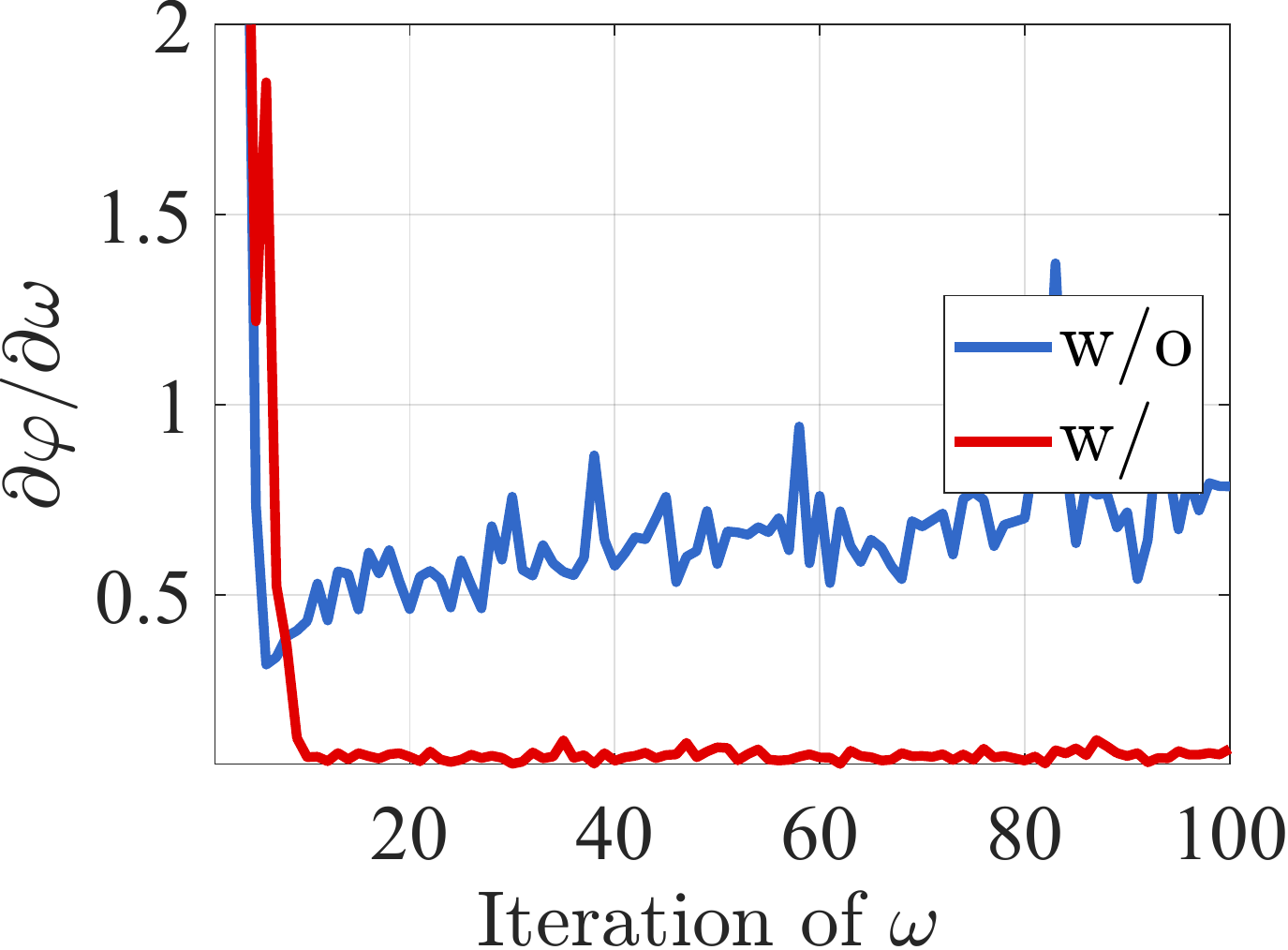}
	\includegraphics[height=3cm,width=4cm]{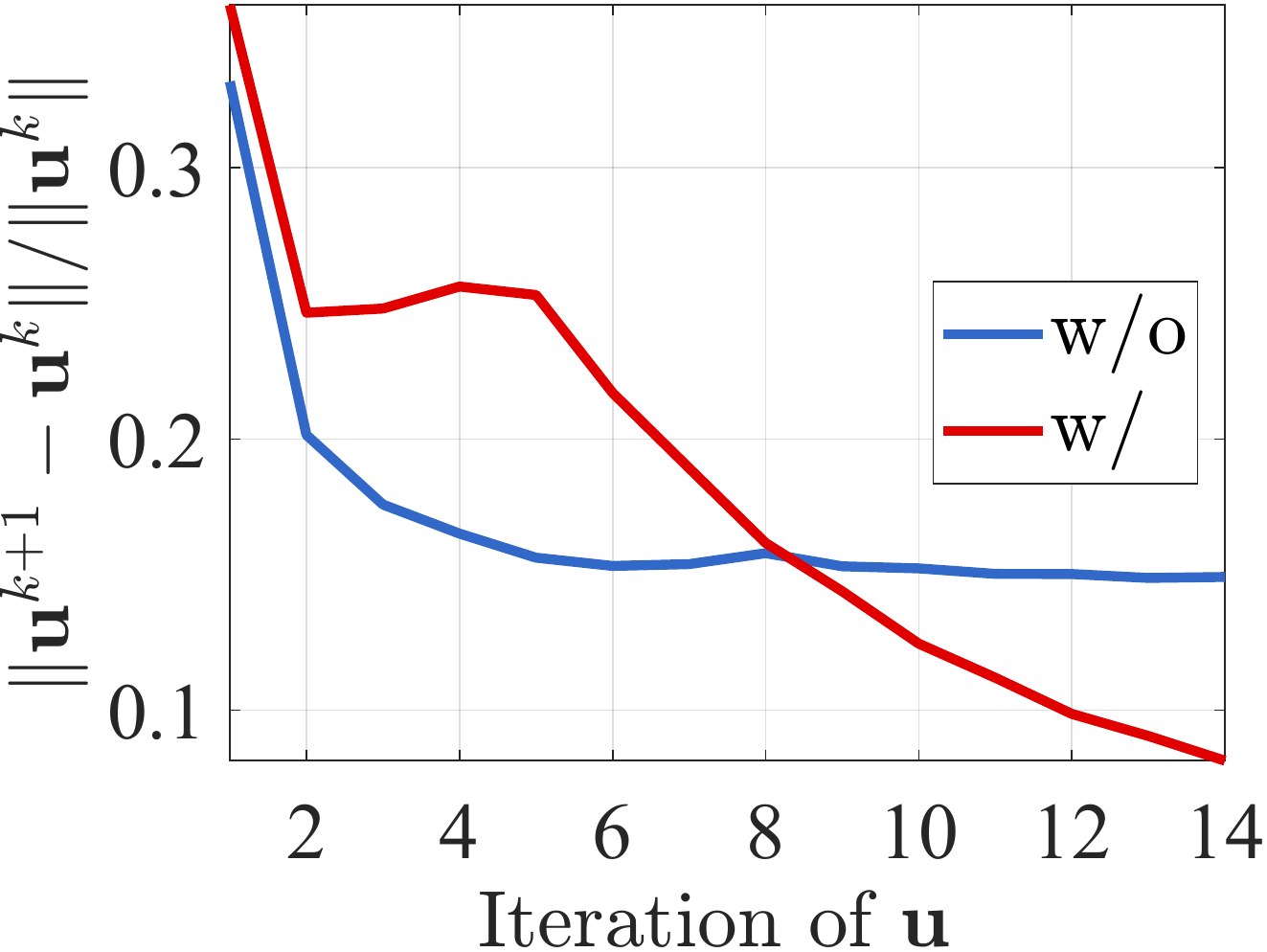} %
	\caption{Convergence curves of ${\partial\varphi}/{\partial\ome}$ with respect to $\ome$ and $\Vert\u^{k+1}-\u^{k}\Vert/\Vert\u^{k}\Vert$ with respect to $\u$, with or without the non-expansive property of operator $\D$ in HODL. 
	The necessity of non-expansive property of $\D$ for HODL can be observed.}
	\label{SN}
\end{figure}

In addition, we investigate some factors that may affect the performance of HODL, including the necessity of non-expansive property, and the influence of parameter $\mu$ on convergence.
In Figure~\ref{SN},  we verify the effect of non-expansive property of $\D$ on the convergence.
\textcolor{black}{It can be seen that the non-expansive property reduces the gradient of the learning objective $\varphi$ by an order of magnitude, increases the convergence stability, and also provides a better convergence of the optimization iteration.
}
These verify the importance of the non-expansive property on the convergence. %
\begin{figure}
	\centering
		\begin{subfigure}[t]{0.23\textwidth}{
			\includegraphics[height=3cm,width=4cm]{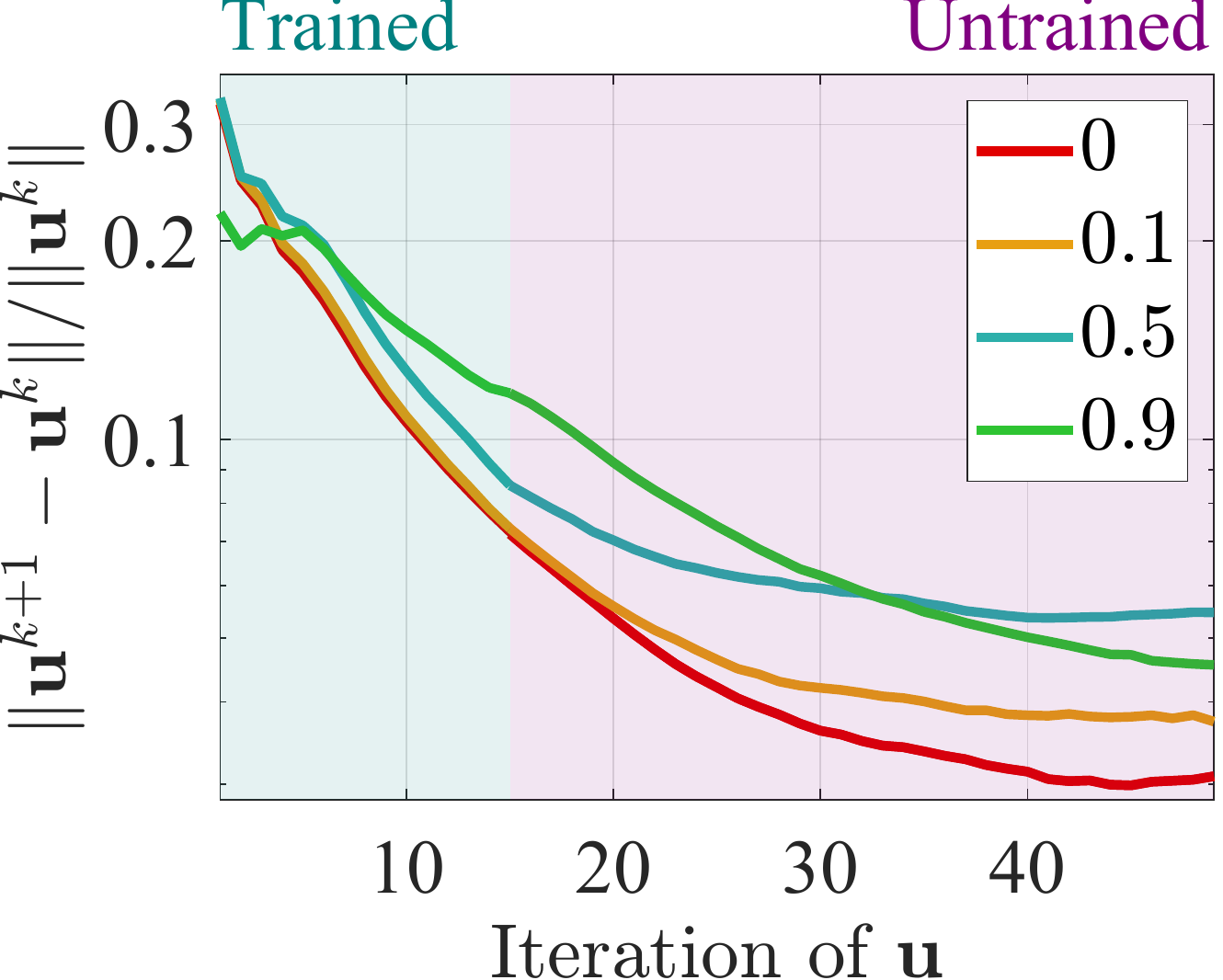}
			\subcaption{$K=15$ in training}
			\label{sub:15_mu}
		}
		\end{subfigure}
		\begin{subfigure}[t]{0.23\textwidth}{
			\includegraphics[height=3cm,width=4cm]{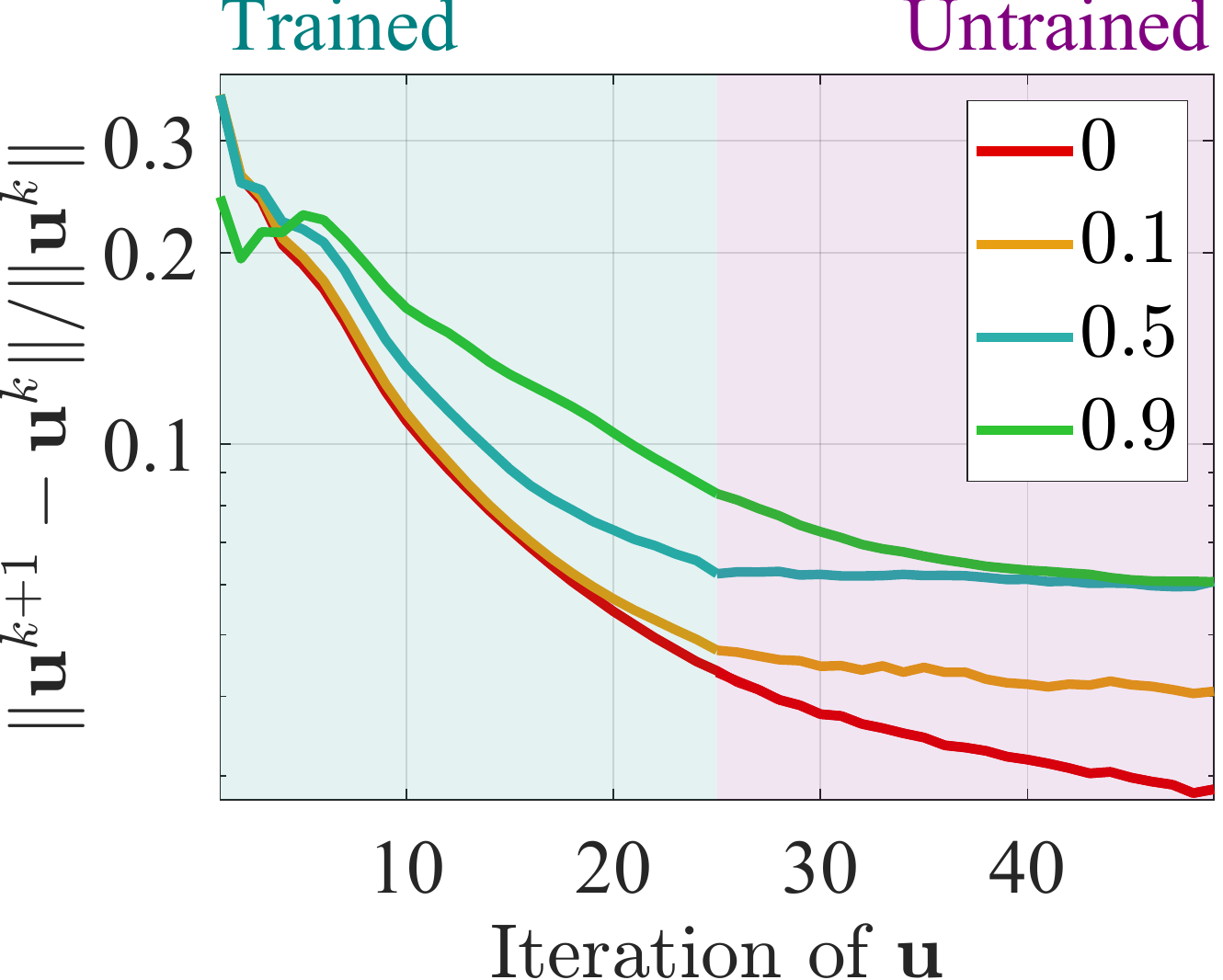}
			\subcaption{$K=25$ in training}
			\label{sub:25_mu}
		}
		\end{subfigure}
	\caption{Convergence curves of $\Vert\u^{k+1}-\u^{k}\Vert/\Vert\u^{k}\Vert$ with different $\mu$ in Eq.~\eqref{eq:simple_bilevel_alg} (aHODL) in Algorithm~\ref{alg:bmo}. Note that HODL with $\mu=0$ is equivalent to sHODL. In the case of complex networks, sHODL is more likely to achieve satisfying performance early in the iteration.
	}
	\label{fig:mu}
\end{figure}
\begin{figure}[t]
	\centering
	\includegraphics[height=3cm]{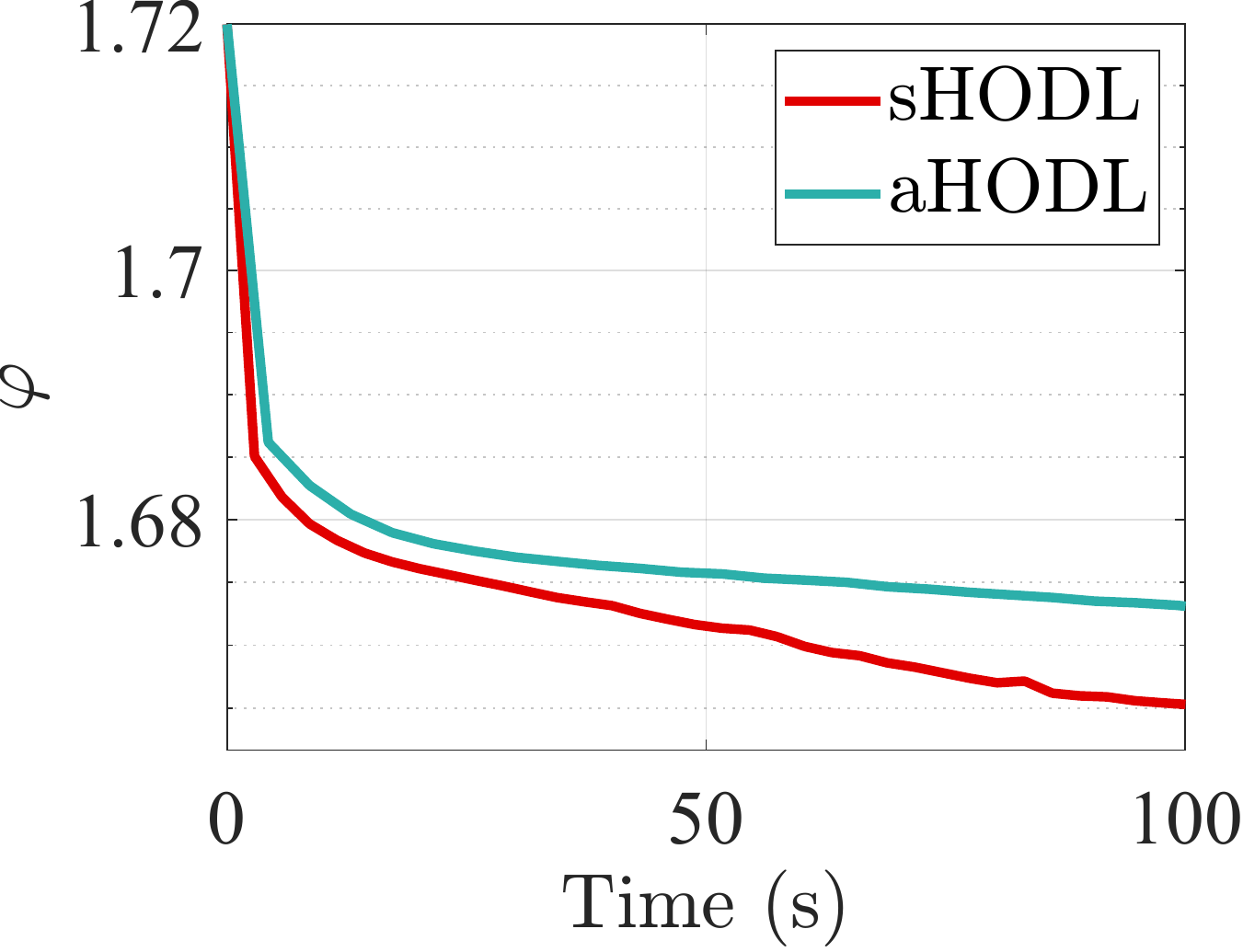}
	\includegraphics[height=3cm]{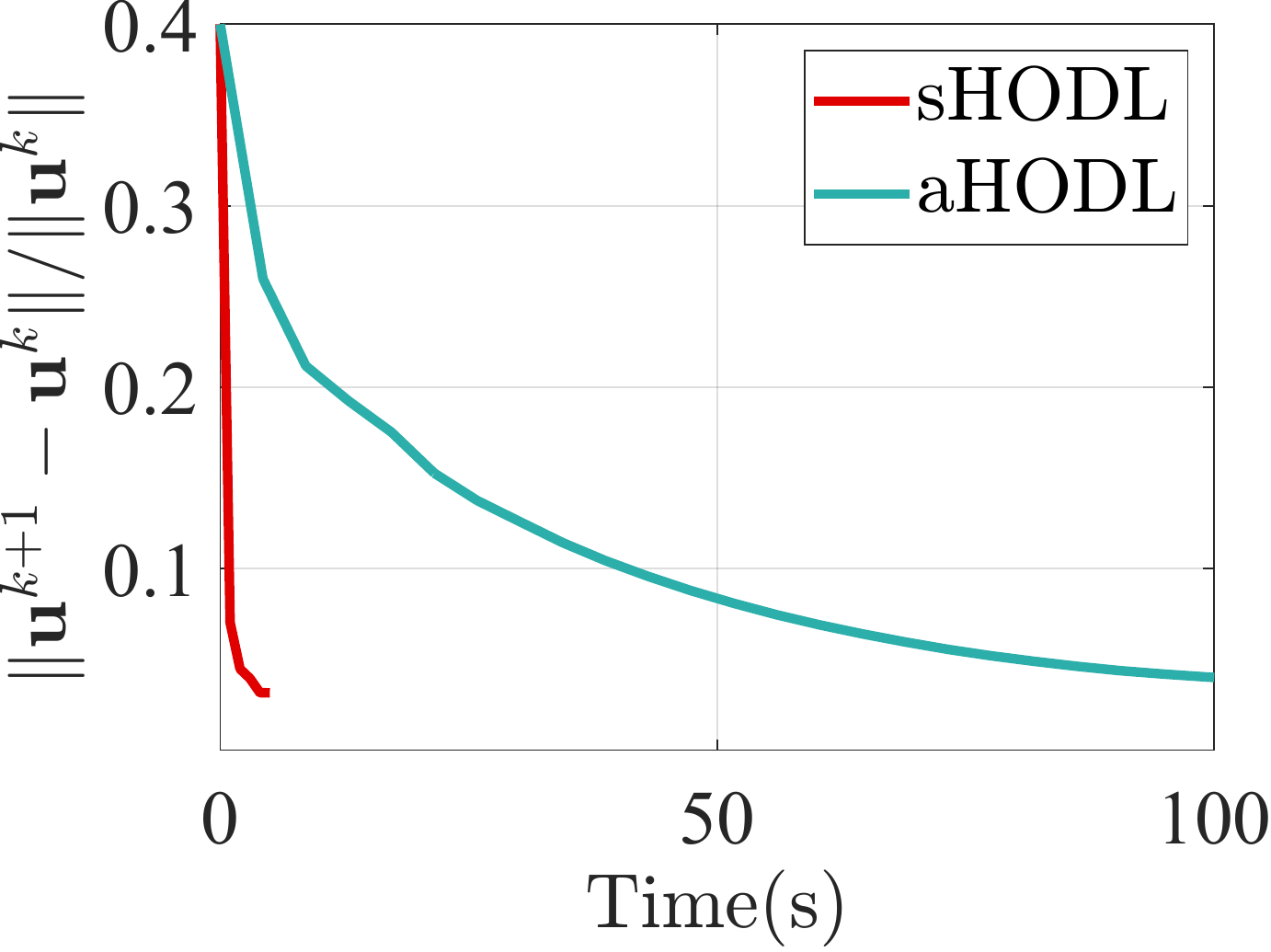}
	\caption{\textcolor{black}{Computational efficiency of sHODL and aHODL. }}
	\label{fig:timeas}
\end{figure}
In Figure~\ref{fig:mu}, we show the impact of different values of parameter~$\mu$ in Eq.~\ref{eq:simple_bilevel_alg} (aHODL) in Algorithm~\ref{alg:bmo} on the convergence of optimization process. %
\textcolor{black}{
	We can find that with the iteration of $\u$, the best performance is obtained for smaller $\mu$ and even for the smallest 0 (i.e., sHODL in Eq.~\ref{eq:sHODL}).
	Comparing~(\subref{sub:15_mu}) and~\textsc{(\subref{sub:25_mu})} in Figure~\ref{fig:mu}, it can be seen that when the network is relatively more fully trained, the advantage of choosing an appropriate $\mu$ is more obvious.
	Figure~\ref{fig:timeas} further compares the computational efficiency of sHODL and aHODL,
	which shows that sHODL is less computationally intensive than aHODL.
}
Therefore, considering the computational complexity burden in practical applications, 
we focus on sHODL from now on, including in the subsequent experiments of vision tasks and extended applications.

	\begin{table} 
		\centering
		\small
		\caption{PSNR results of constrained sparse coding using ALM and ADMM. ADMM performs better than ALM for UNH and ENA, but our approach achieves the best performance only using ALM. 
			Best and second best results are marked in red and blue respectively.
		}
		\label{tab:ALM or ADMM table}
		\begin{tabular}{c|c|c|c} %
			\hline
			Methods&  UNH & ENA&HODL \\
			\hline
			\multirow{1}{*}{ADMM} &  11.27$\pm$2.71  & \textcolor{blue}{\textbf{15.58$\pm$0.89}}&N/A \\
			
			\hline
			\multirow{1}{*}{ALM}  & 7.32$\pm$4.65  & 13.78$\pm$5.23&\textcolor{red}{\textbf{18.64$\pm$0.74}}  \\
			\hline
		\end{tabular}%
	\end{table}

Finally, it should be noted that for constrained problems, existing methods typically use ALM or ADMM, while HODL uses ALM as $\D_\num$. 
For the fairness of comparisons, we examine the performance of UNH and ENA using ALM and ADMM, and HODL using ALM. 
As can be seen in Table~\ref{tab:ALM or ADMM table}, the performance using ALM is weaker than ADMM in both existing UNA and ENA methods, while our HODL only using ALM is able to outperform other methods, further demonstrating the effectiveness of HODL.
Actually, aforementioned experiments of UNH and ENA for comparison are conducted using ADMM as the base method. %

\begin{table}
	\centering
	\caption{Averaged PSNR and SSIM results for the single image rain removal task on two widely used synthesized datasets, Rain100L and Rain100H~\cite{derain}. 
		Best and second best results are marked in red and blue respectively.}
	
	\begin{tabular}{c|cc|cc}
		\hline   { \multirow{1}{*}{Datasets}  } &   \multicolumn{2}{c|}{ Rain 100L } & \multicolumn{2}{c}{  { Rain 100H }} \\
		{Metrics} &   { PSNR} &   {SSIM } &   { PSNR} &   {SSIM } \\
		\hline   
		{ DSC \textcolor{black}{(ENA)} } & 27.34 & 0.849 & 13.77 & 0.319 \\
		{ GMM \textcolor{black}{(ENA)}} & 29.05 & 0.871 & 15.23 & 0.449 \\
		{ JCAS } & 28.54 & 0.852 & 14.62 & 0.451 \\
		{ Clear } & 30.24 & 0.934 & 15.33 & 0.742 \\
		{ DDN } & 32.38 & 0.925 & 22.85 & 0.725 \\
		{ RESCAN } & 38.52 & 0.981 & 29.62 & 0.872 \\
		{ PReNet \textcolor{black}{(ENA)}} & 37.45 & 0.979 & 30.11 & \textcolor{red}{\textbf{0.905}} \\
		{ SPANet } & 35.33 & 0.969 & 25.11 & 0.833 \\
		\text{ JORDER\_E } & 38.59 & \textcolor{blue}{\textbf{0.983}} & 30.50 & 0.896 \\
		{ SIRR } & 32.37 & 0.925 & 22.47 & 0.716 \\
		{ MPRNet } & 36.40 & 0.965 & 30.41 & 0.890 \\
		{ RCDNet \textcolor{black}{(ENA)}} & \textcolor{blue}{\textbf{40.00}} & \textcolor{red}{\textbf{0.986}} & \textcolor{red}{\textbf{31.28}} & \textcolor{blue}{\textbf{0.903}} \\
		{ HODL } & \textcolor{red}{\textbf{40.07}} & \textcolor{red}{\textbf{0.986}} &\textcolor{blue}{\textbf{30.96}} & \textcolor{red}{\textbf{0.905}} \\
		\hline
		
	\end{tabular}
	\label{tab:derain_table}
\end{table}

\begin{figure*}[!t]
	\begin{center}
		\begin{tabular}{c@{\extracolsep{0.3em}}c@{\extracolsep{0.3em}}c@{\extracolsep{0.3em}}c@{\extracolsep{0.3em}}c@{\extracolsep{0.3em}}c@{\extracolsep{0.3em}}c@{\extracolsep{0.3em}}}
			
			\includegraphics[width=0.13\linewidth,trim=0 0 0 0,clip]{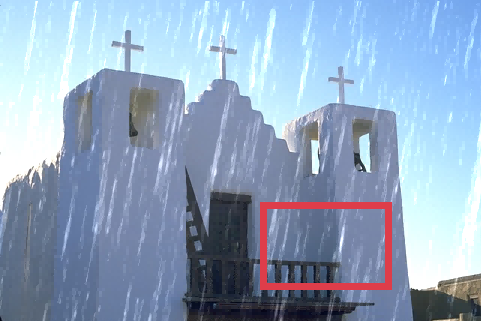}&
			\includegraphics[width=0.13\linewidth,trim=0 0 0 0,clip]{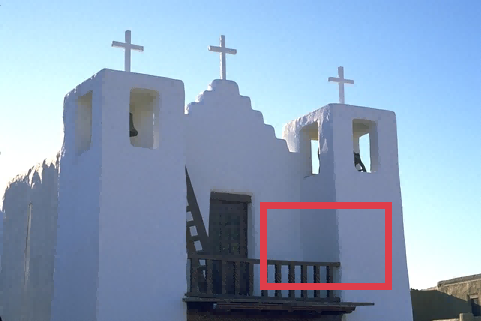}&\includegraphics[width=0.13\linewidth,trim=0 0 0 0,clip]{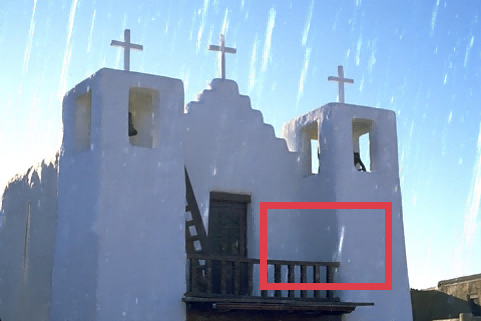}&\includegraphics[width=0.13\linewidth,trim=0 0 0 0,clip]{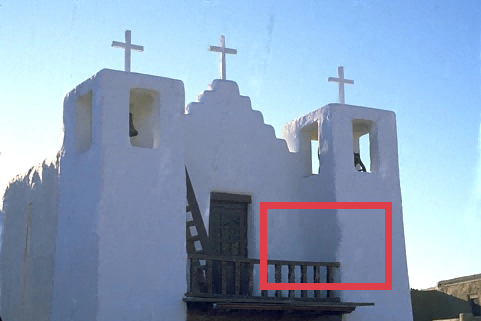}&\includegraphics[width=0.13\linewidth,trim=0 0 0 0,clip]{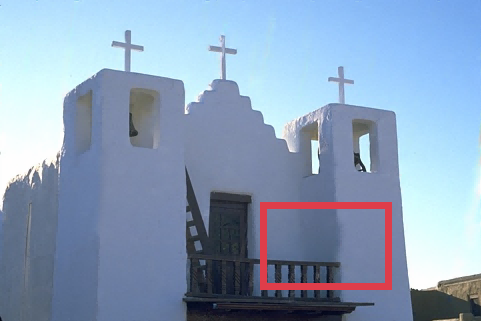}&\includegraphics[width=0.13\linewidth,trim=0 0 0 0,clip]{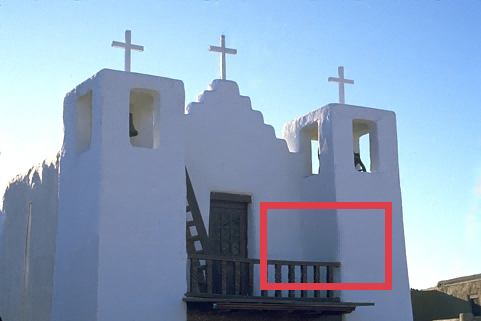}&\includegraphics[width=0.13\linewidth,trim=0 0 0 0,clip]{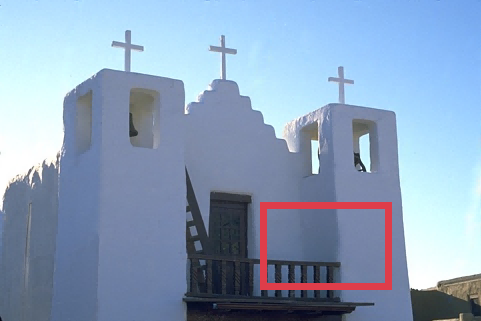}\\ \specialrule{0em}{-0.5pt}{-1pt}
			\includegraphics[width=0.13\linewidth,trim=0 0 0 0,clip]{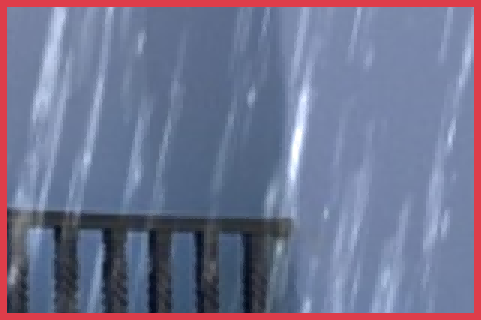}&\includegraphics[width=0.13\linewidth,trim=0 0 0 0,clip]{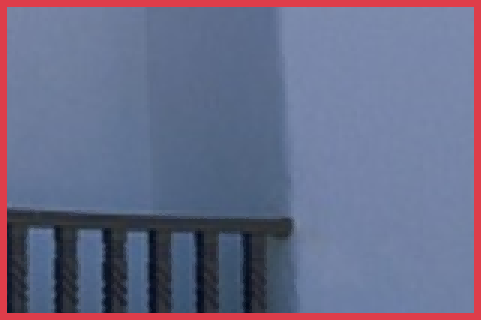}&\includegraphics[width=0.13\linewidth,trim=0 0 0 0,clip]{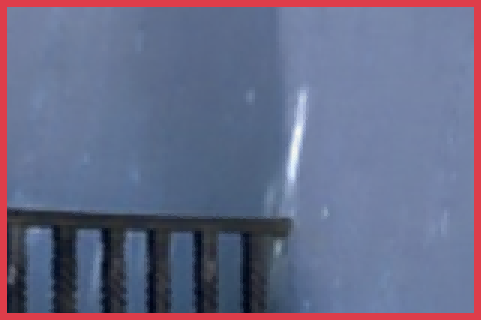}&\includegraphics[width=0.13\linewidth,trim=0 0 0 0,clip]{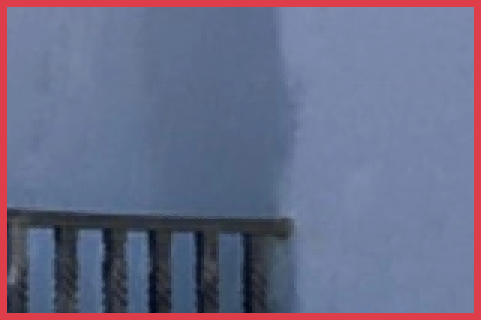}&\includegraphics[width=0.13\linewidth,trim=0 0 0 0,clip]{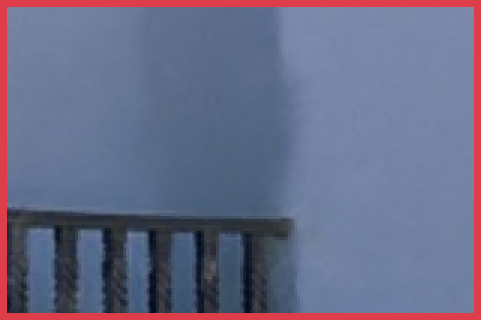}&\includegraphics[width=0.13\linewidth,trim=0 0 0 0,clip]{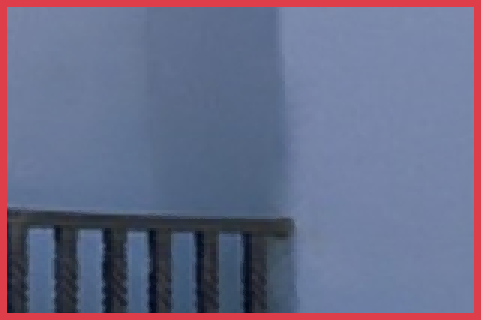}&\includegraphics[width=0.13\linewidth,trim=0 0 0 0,clip]{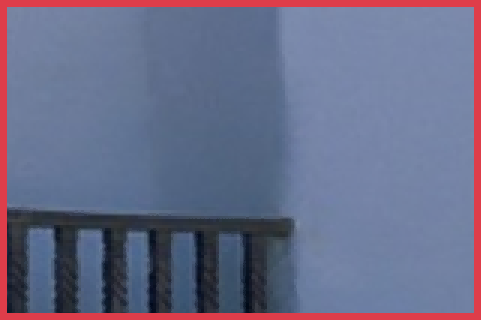}\\  	
			\specialrule{0em}{-0.5pt}{-1pt} 
			\footnotesize - & -  &\footnotesize20.94/0.84& \footnotesize 31.87/0.97 & \footnotesize  33.14/0.98 & \footnotesize \textcolor{blue}{\textbf{36.00/0.98}}& \footnotesize  \textcolor{red}{\textbf{36.27/0.99}} \\   
			
			\includegraphics[width=0.13\linewidth,trim=0 0 0 0,clip]{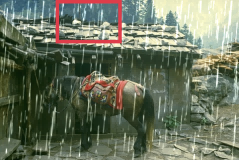}&
			\includegraphics[width=0.13\linewidth,trim=0 0 0 0,clip]{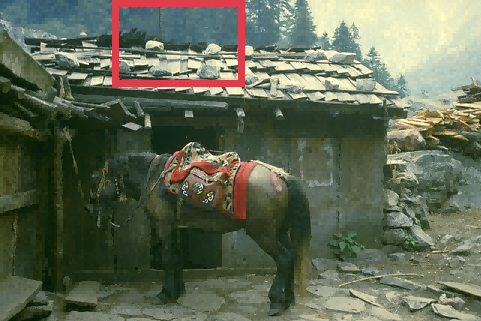}&\includegraphics[width=0.13\linewidth,trim=0 0 0 0,clip]{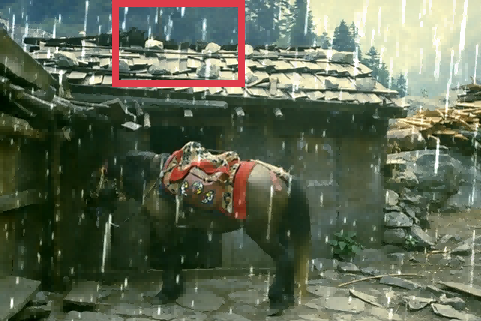}&\includegraphics[width=0.13\linewidth,trim=0 0 0 0,clip]{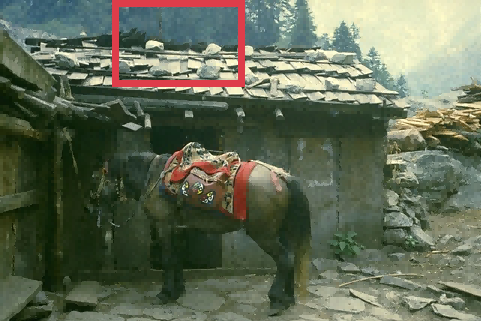}&\includegraphics[width=0.13\linewidth,trim=0 0 0 0,clip]{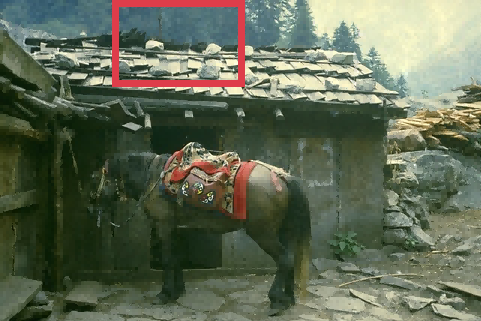}&\includegraphics[width=0.13\linewidth,trim=0 0 0 0,clip]{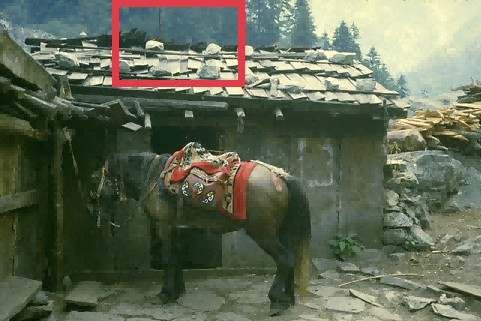}&\includegraphics[width=0.13\linewidth,trim=0 0 0 0,clip]{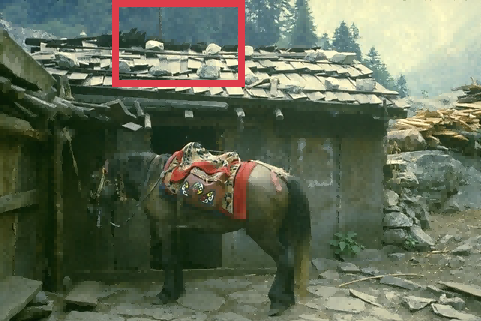}\\ \specialrule{0em}{-0.5pt}{-1pt}
			\includegraphics[width=0.13\linewidth,trim=0 0 0 0,clip]{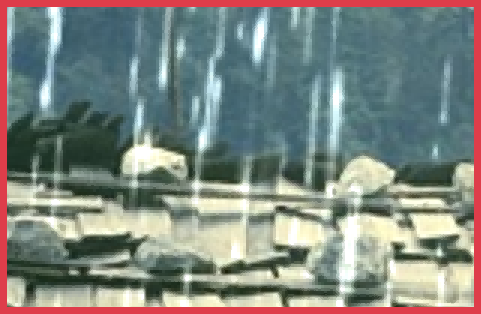}&\includegraphics[width=0.13\linewidth,trim=0 0 0 0,clip]{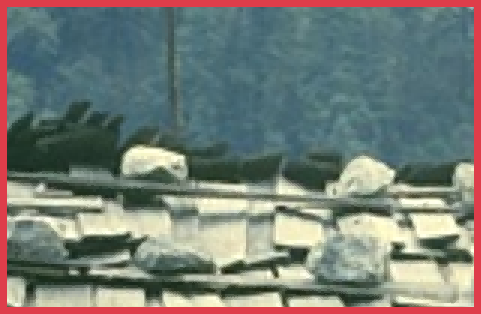}&\includegraphics[width=0.13\linewidth,trim=0 0 0 0,clip]{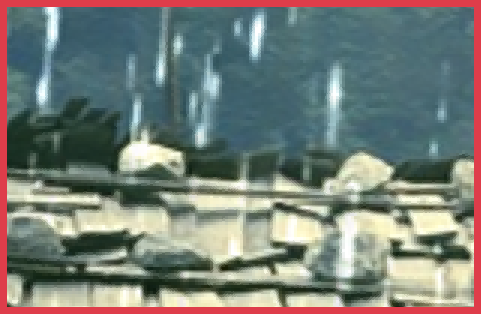}&\includegraphics[width=0.13\linewidth,trim=0 0 0 0,clip]{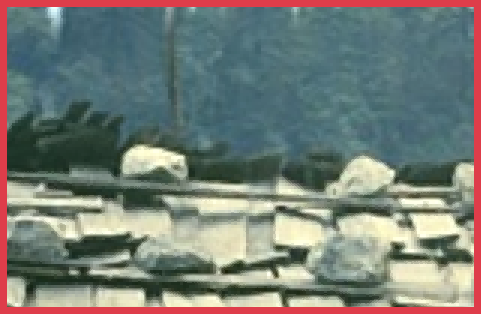}&\includegraphics[width=0.13\linewidth,trim=0 0 0 0,clip]{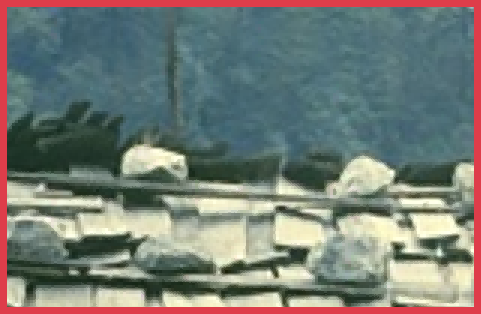}&\includegraphics[width=0.13\linewidth,trim=0 0 0 0,clip]{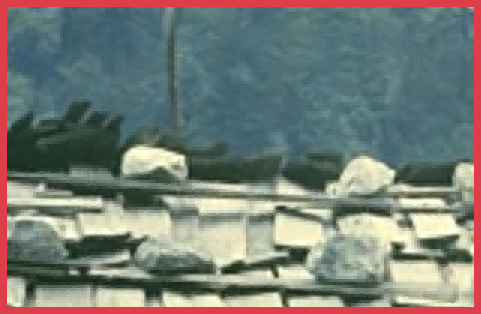}&\includegraphics[width=0.13\linewidth,trim=0 0 0 0,clip]{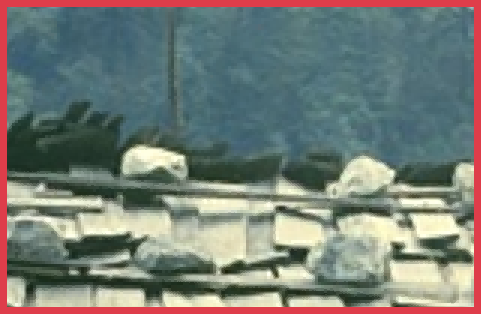}\\  	
			\specialrule{0em}{-0.5pt}{-1pt} 
			\footnotesize - & -  &\footnotesize26.23/0.90& \footnotesize 35.41/0.97 & \footnotesize 36.69/0.98 & \footnotesize \textcolor{blue}{\textbf{38.85/0.99}}& \footnotesize  \textcolor{red}{\textbf{39.16/0.99}} \\   
			\footnotesize Input &\footnotesize Ground Truth & \footnotesize  DDN & \footnotesize JORDER & \footnotesize  PReNet \textcolor{black}{(ENA)} & \footnotesize RCDNet \textcolor{black}{(ENA)}& \footnotesize  HODL\\

		\end{tabular}
	\end{center}
	\caption{Visual results of the rain streak removal task on two samples from Rain100L, compared with DDN, JORDER, PReNet and RCDNet. The hierarchical structure of HODL reduces the distortion and blur introduced by removing rain lines. 
		Two metrics (PSNR / SSIM) are listed below each image to quantify the quality of generated images. 
		Best and second best results are marked in red and blue respectively.
	}
	\label{fig Deraining results}
\end{figure*}

\begin{figure*}[!t]
	\begin{center}
		\begin{tabular}{c@{\extracolsep{0.3em}}c@{\extracolsep{0.3em}}c@{\extracolsep{0.3em}}c@{\extracolsep{0.3em}}c@{\extracolsep{0.3em}}c@{\extracolsep{0.3em}}c@{\extracolsep{0.3em}}}
			
			\includegraphics[width=0.13\linewidth,trim=0 0 0 0,clip]{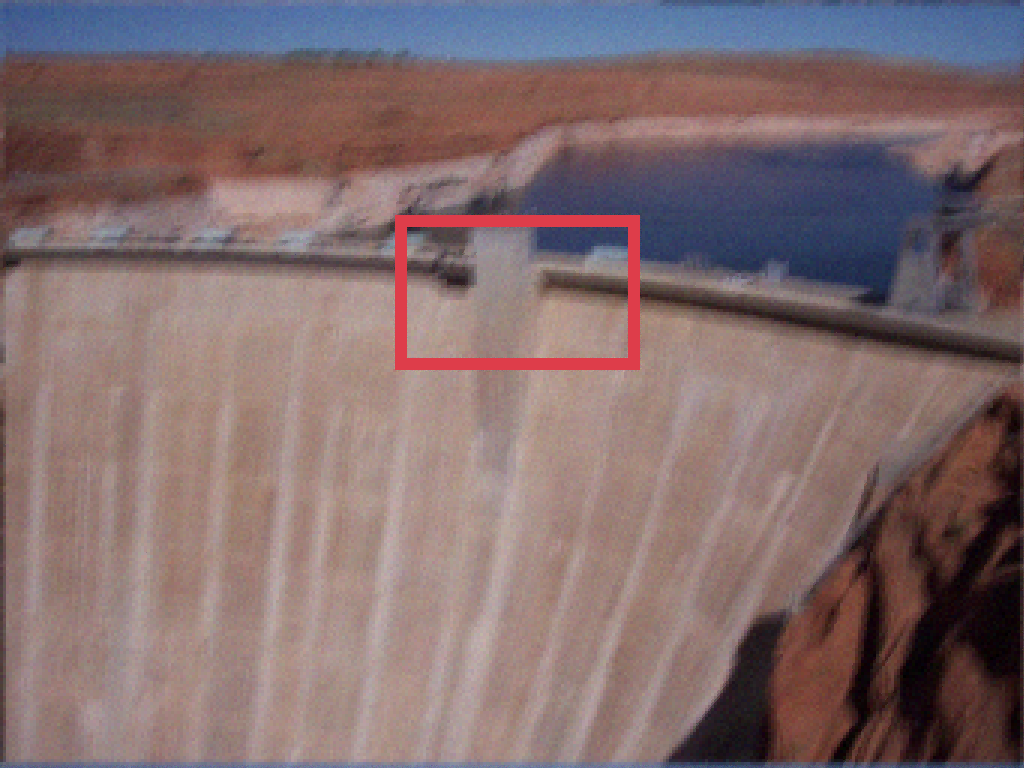}&
			\includegraphics[width=0.13\linewidth,trim=0 0 0 0,clip]{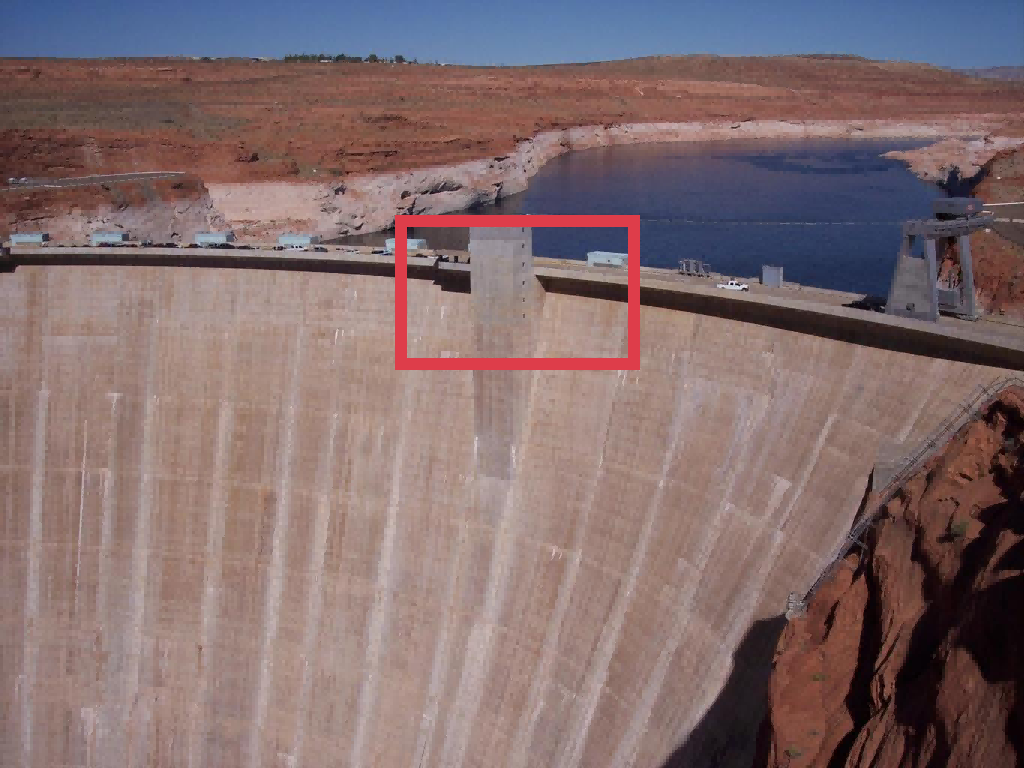}&\includegraphics[width=0.13\linewidth,trim=0 0 0 0,clip]{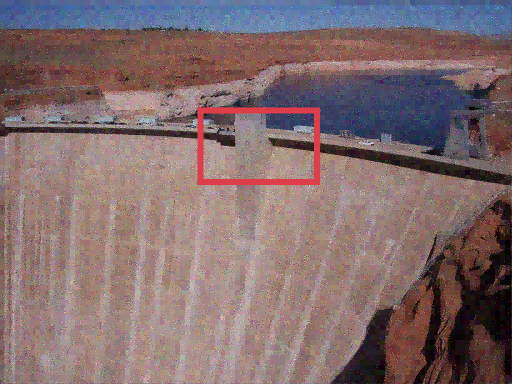}&\includegraphics[width=0.13\linewidth,trim=0 0 0 0,clip]{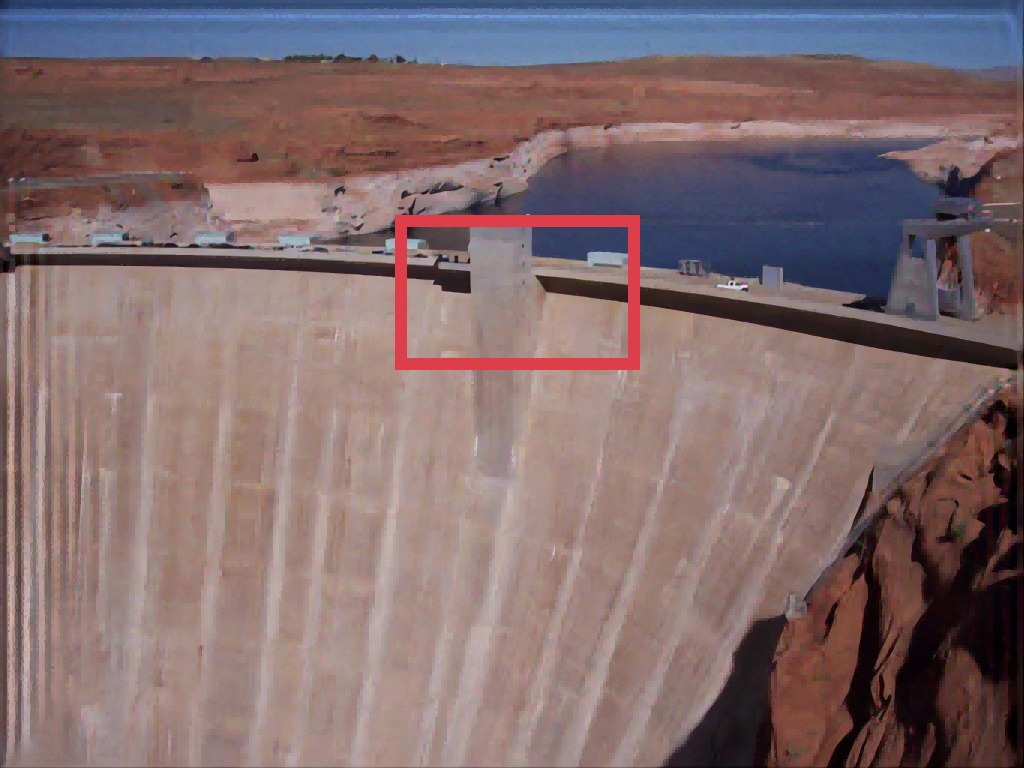}&\includegraphics[width=0.13\linewidth,trim=0 0 0 0,clip]{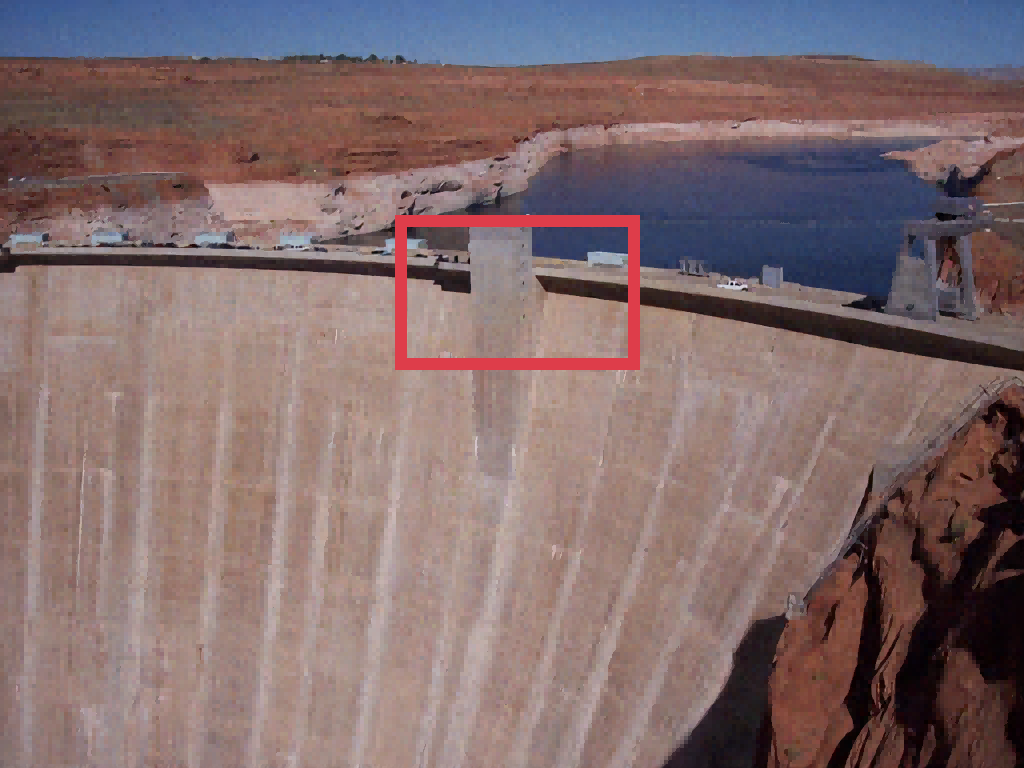}&\includegraphics[width=0.13\linewidth,trim=0 0 0 0,clip]{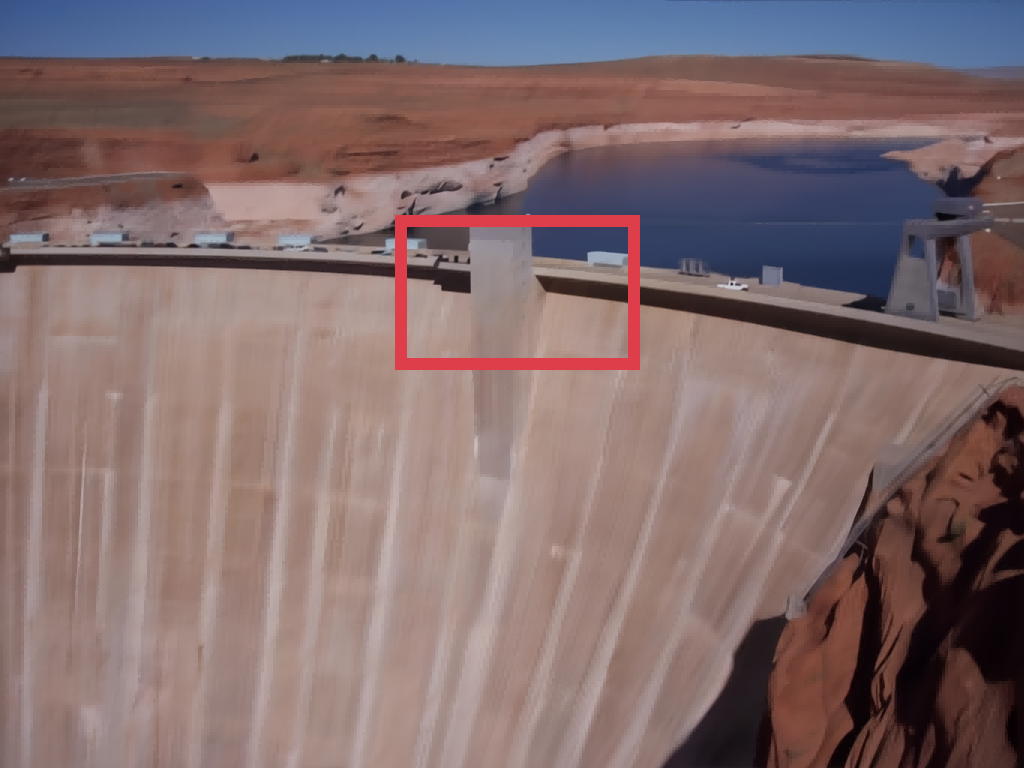}&\includegraphics[width=0.13\linewidth,trim=0 0 0 0,clip]{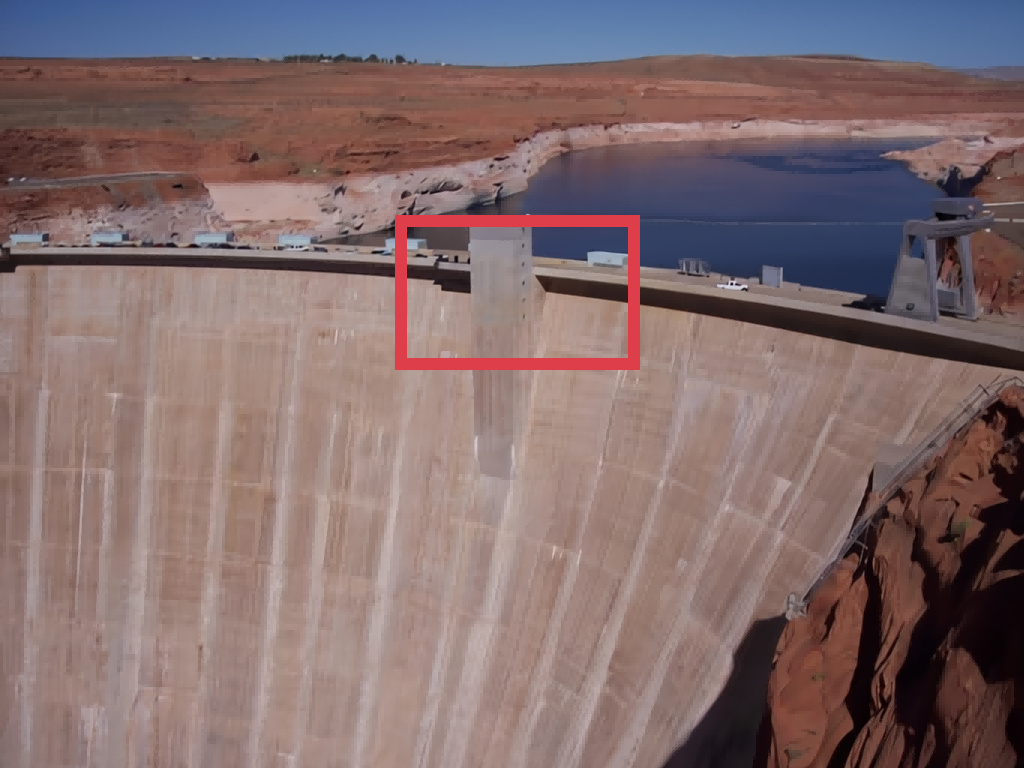}\\ \specialrule{0em}{-0.5pt}{-1pt}
			\includegraphics[width=0.13\linewidth,trim=0 0 0 0,clip]{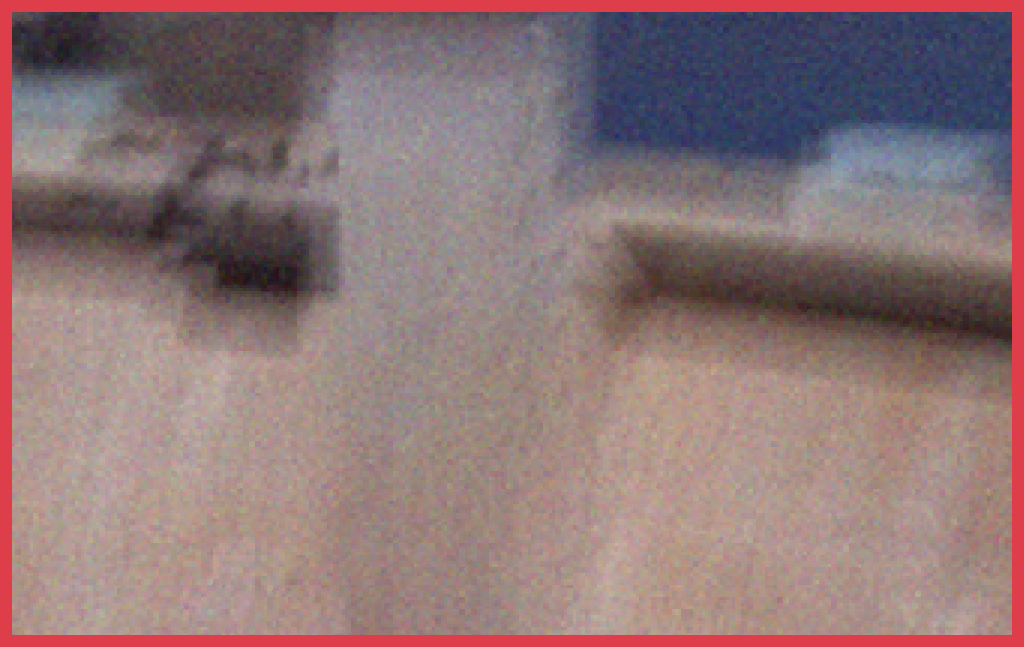}&\includegraphics[width=0.13\linewidth,trim=0 0 0 0,clip]{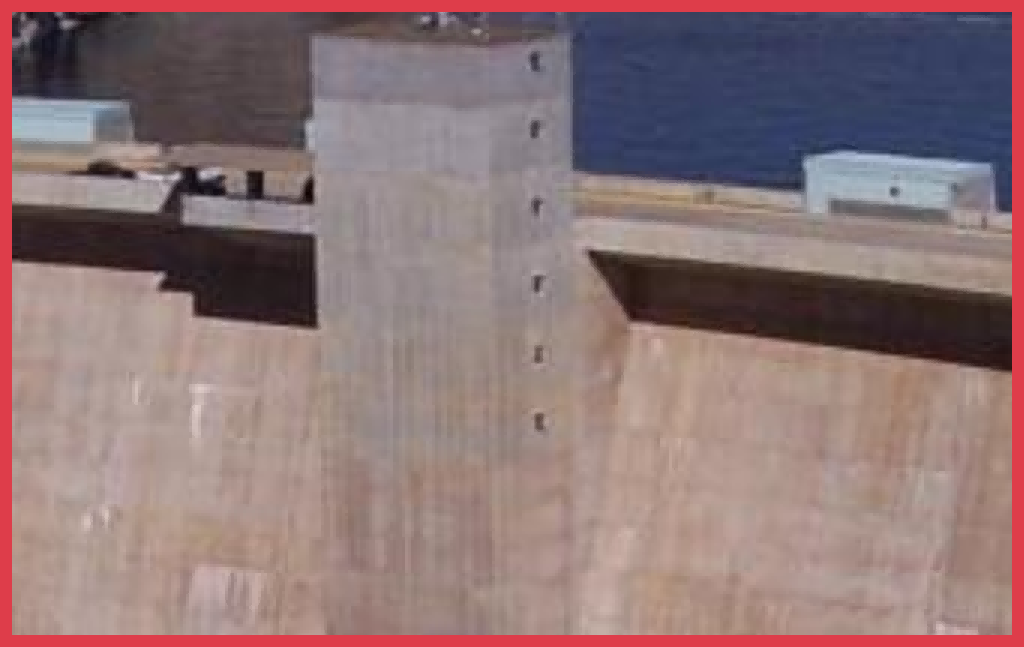}&\includegraphics[width=0.13\linewidth,trim=0 0 0 0,clip]{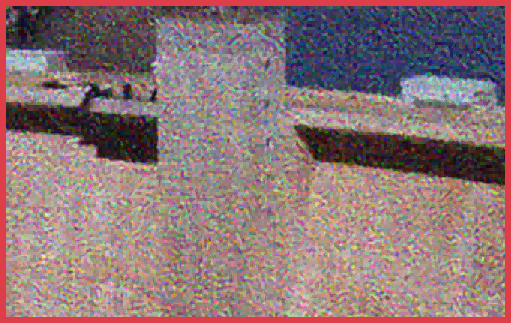}&\includegraphics[width=0.13\linewidth,trim=0 0 0 0,clip]{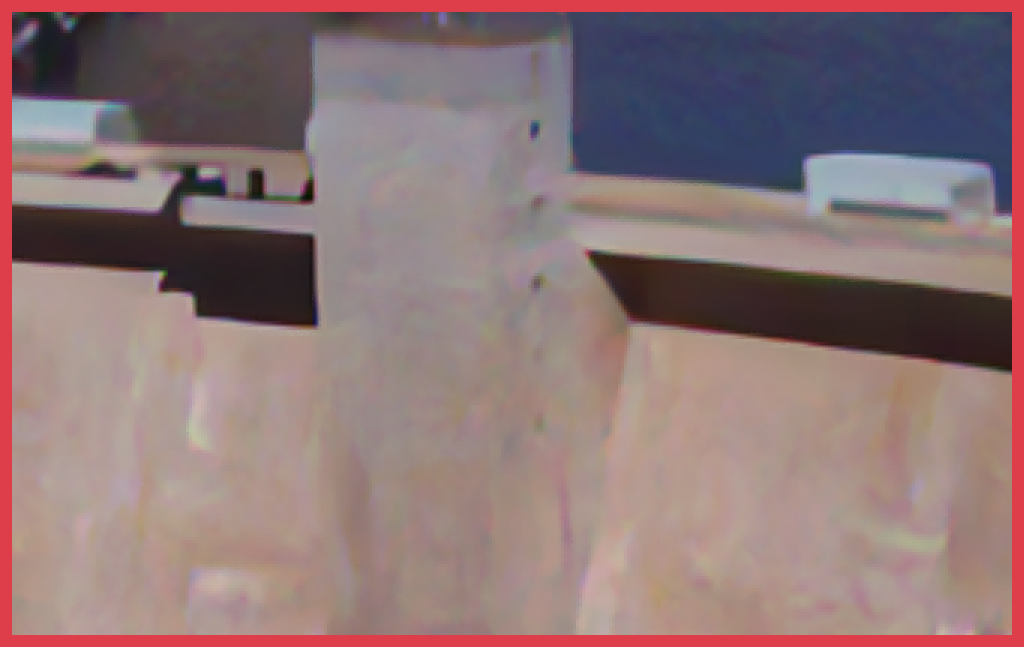}&\includegraphics[width=0.13\linewidth,trim=0 0 0 0,clip]{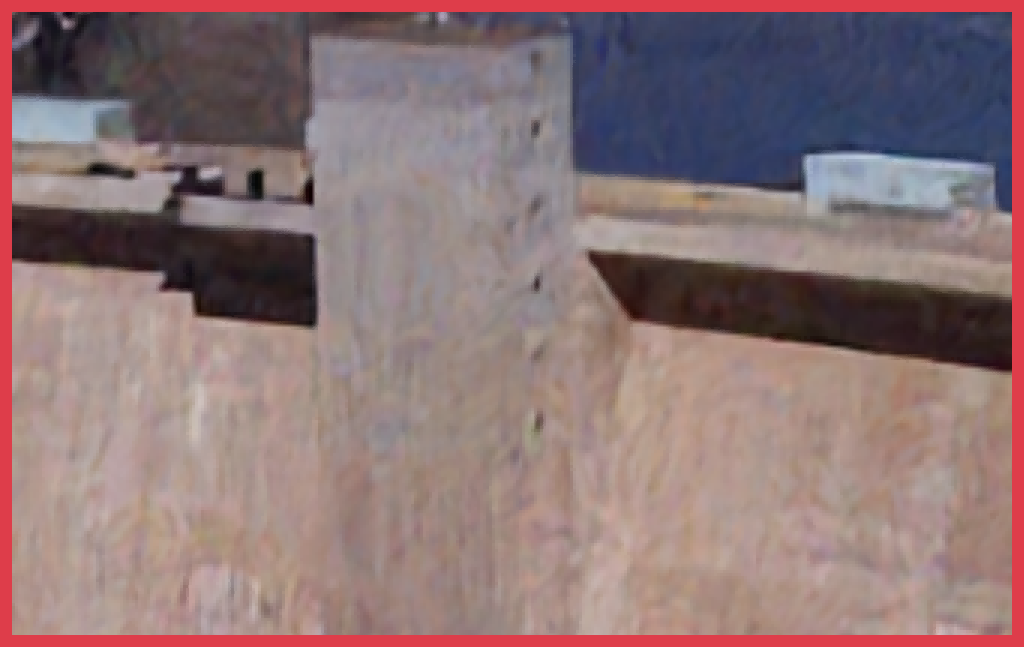}&\includegraphics[width=0.13\linewidth,trim=0 0 0 0,clip]{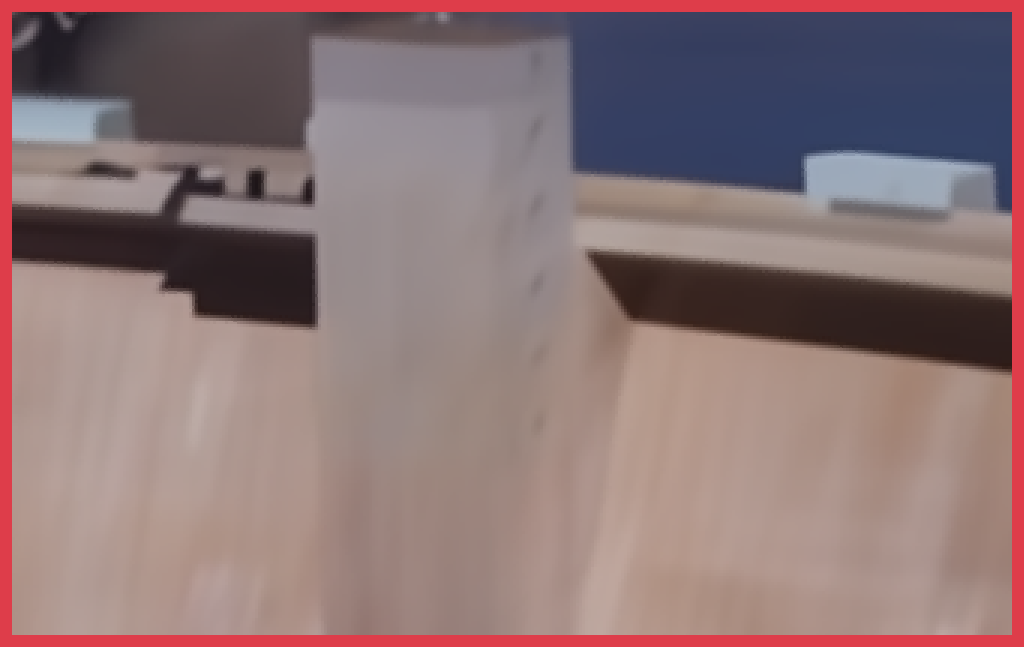}&\includegraphics[width=0.13\linewidth,trim=0 0 0 0,clip]{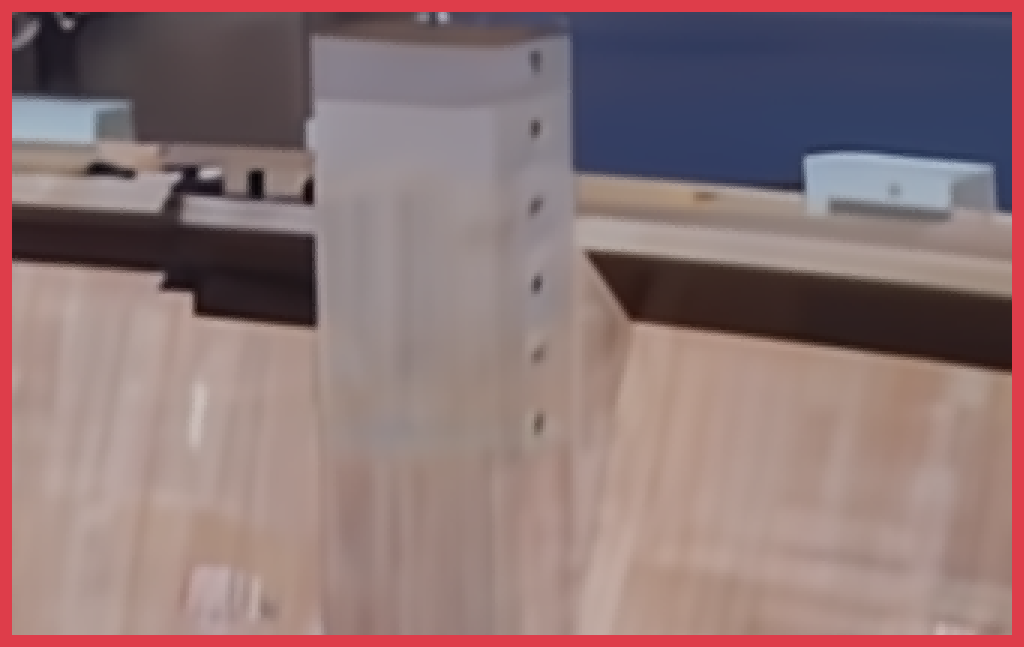}\\  	
			\specialrule{0em}{-0.5pt}{-1pt} 
			\footnotesize - & -  &\footnotesize17.46/0.35& \footnotesize 28.84/0.91 & \footnotesize  31.72/0.93 & \footnotesize \textcolor{blue}{\textbf{31.78/0.94}}& \footnotesize  \textcolor{red}{\textbf{33.07/0.95}} \\   
			
			\includegraphics[width=0.13\linewidth,trim=0 0 0 0,clip]{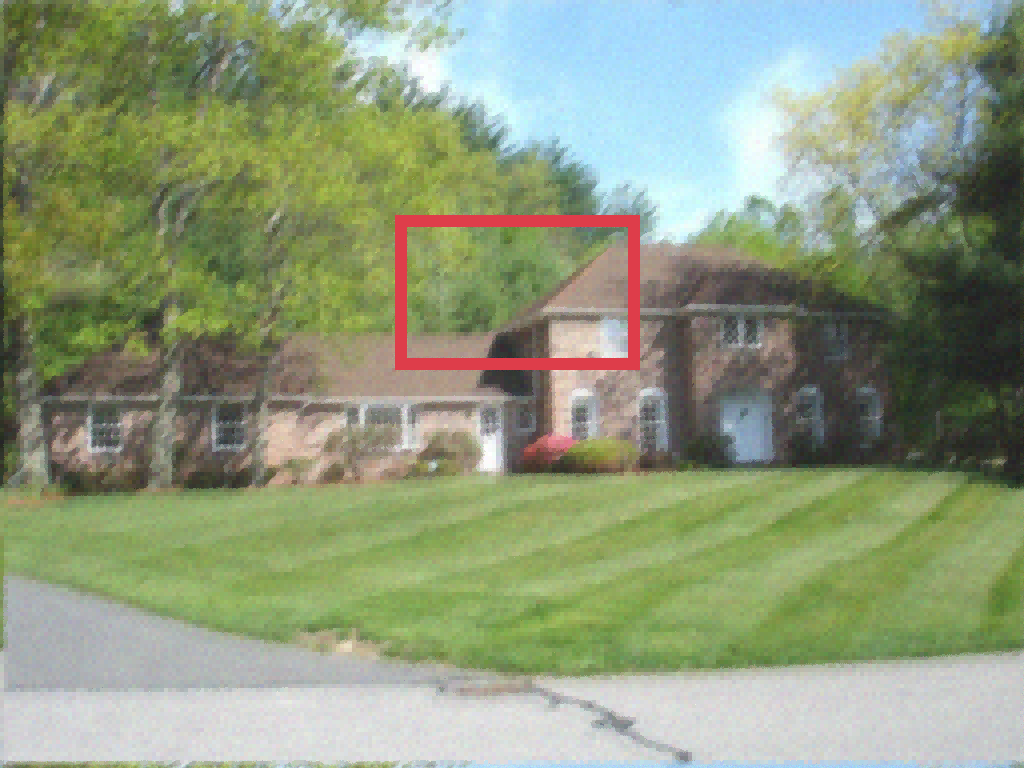}&
			\includegraphics[width=0.13\linewidth,trim=0 0 0 0,clip]{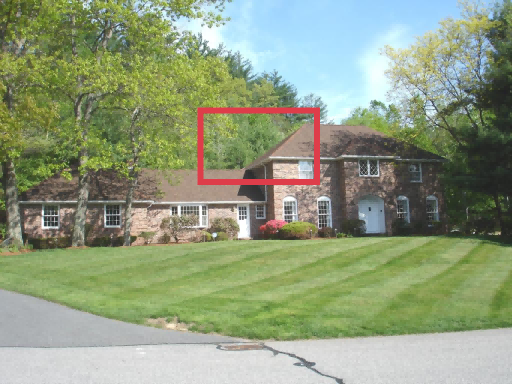}&\includegraphics[width=0.13\linewidth,trim=0 0 0 0,clip]{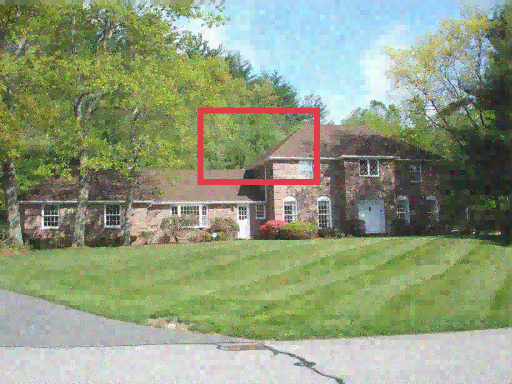}&\includegraphics[width=0.13\linewidth,trim=0 0 0 0,clip]{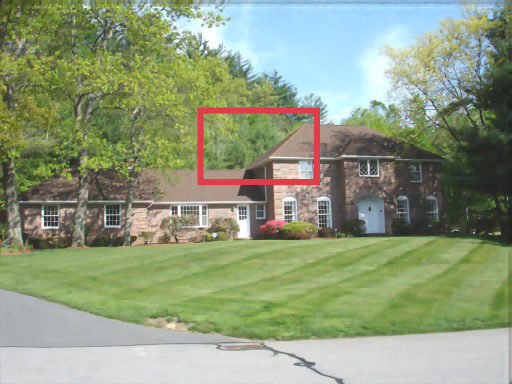}&\includegraphics[width=0.13\linewidth,trim=0 0 0 0,clip]{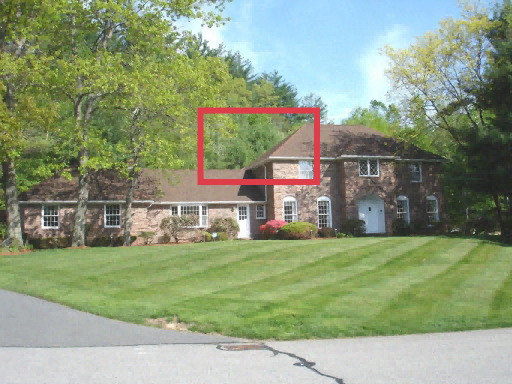}&\includegraphics[width=0.13\linewidth,trim=0 0 0 0,clip]{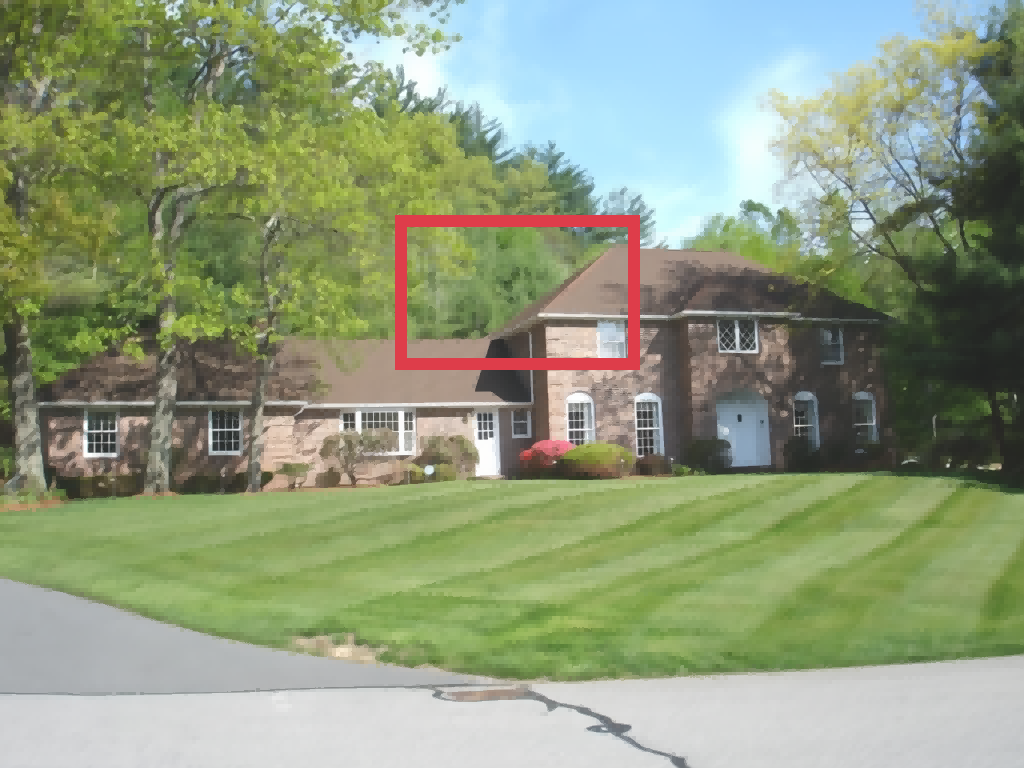}&\includegraphics[width=0.13\linewidth,trim=0 0 0 0,clip]{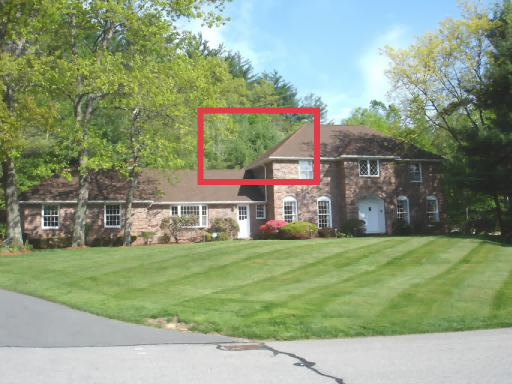}\\ \specialrule{0em}{-0.5pt}{-1pt}
			\includegraphics[width=0.13\linewidth,trim=0 0 0 0,clip]{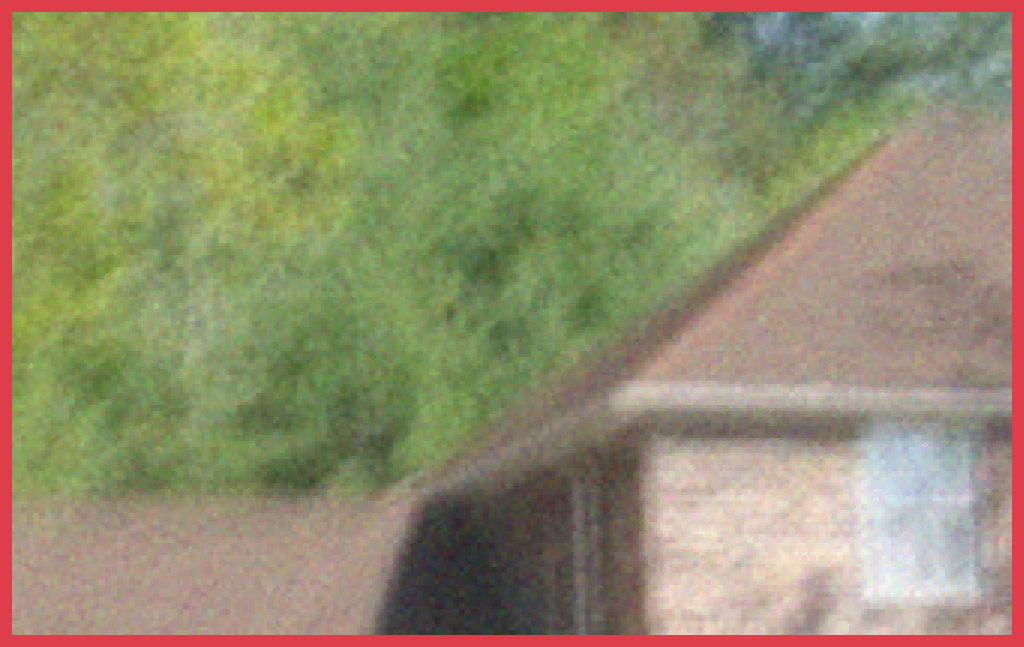}&\includegraphics[width=0.13\linewidth,trim=0 0 0 0,clip]{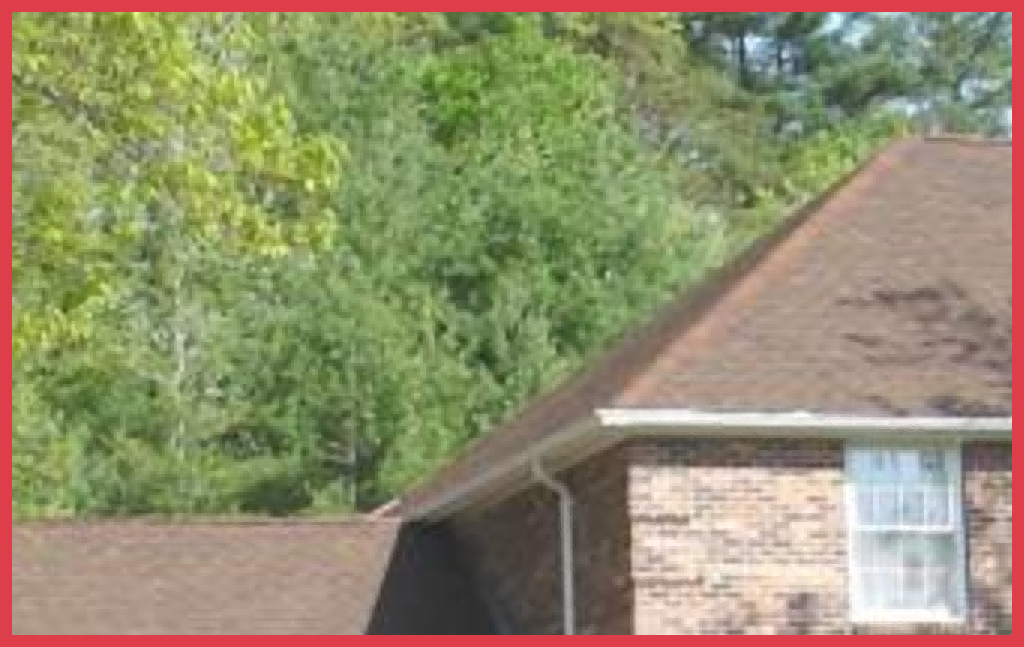}&\includegraphics[width=0.13\linewidth,trim=0 0 0 0,clip]{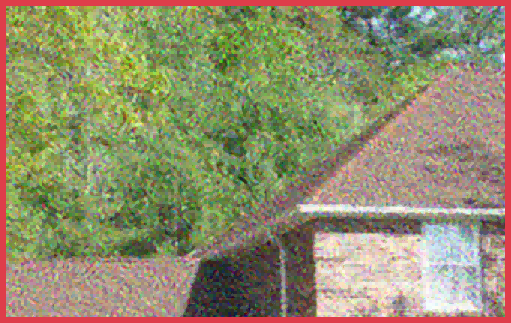}&\includegraphics[width=0.13\linewidth,trim=0 0 0 0,clip]{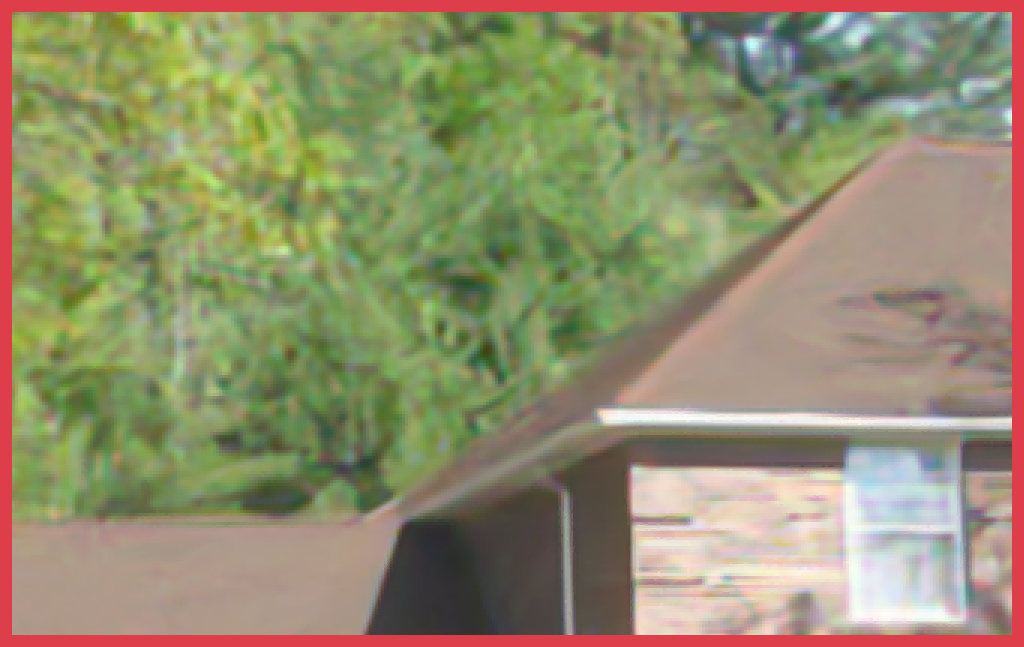}&\includegraphics[width=0.13\linewidth,trim=0 0 0 0,clip]{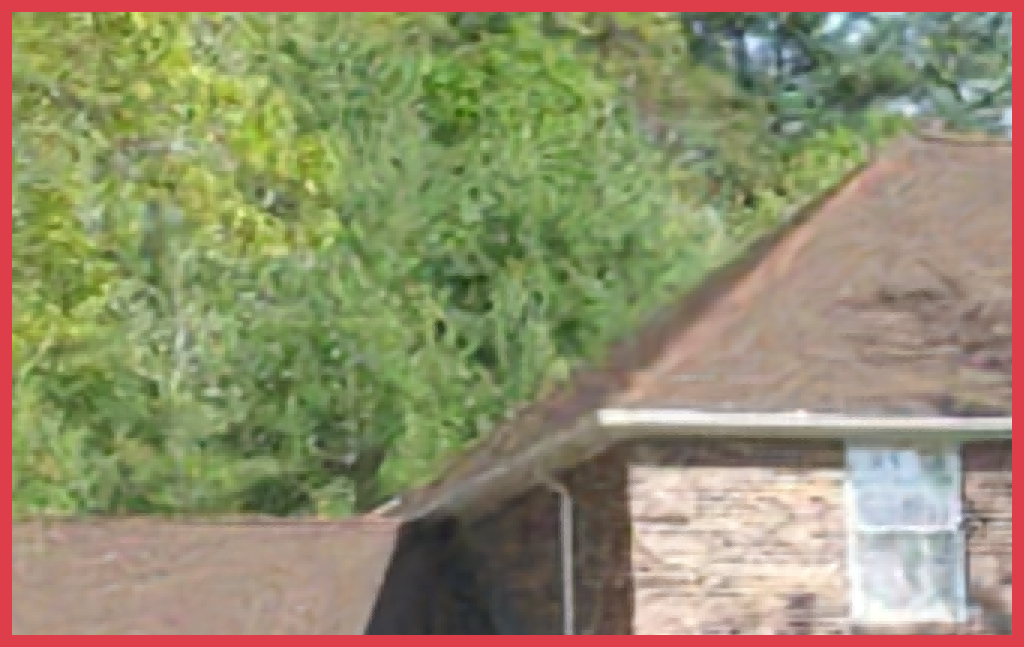}&\includegraphics[width=0.13\linewidth,trim=0 0 0 0,clip]{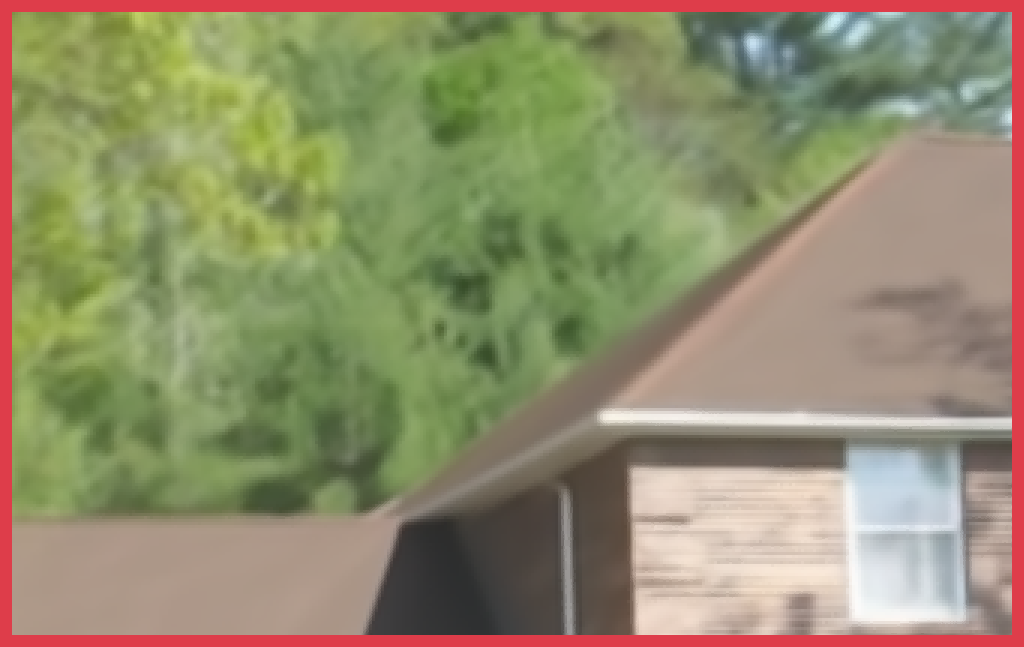}&\includegraphics[width=0.13\linewidth,trim=0 0 0 0,clip]{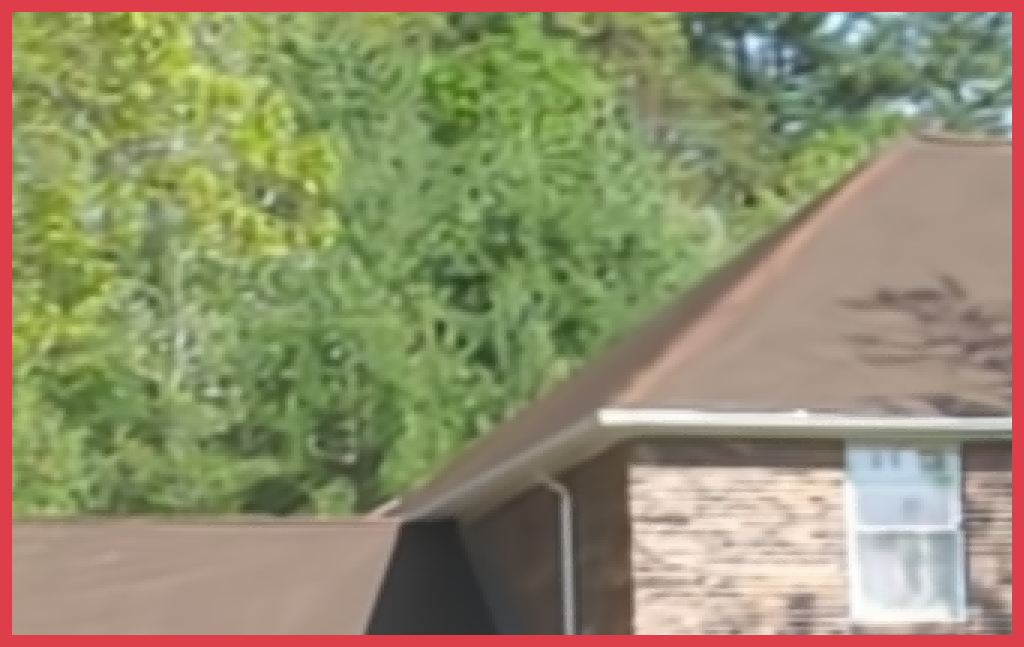}\\  	
			\specialrule{0em}{-0.5pt}{-1pt} 			
			\footnotesize - & -  &\footnotesize18.33/0.50& \footnotesize 26.64/0.87 &\footnotesize \textcolor{blue}{\textbf{  28.10/0.89}} & \footnotesize {{27.42/0.87}}& \footnotesize  \textcolor{red}{\textbf{28.67/0.91}} \\   
			\footnotesize Input &\footnotesize Ground Truth & \footnotesize  FDN & \footnotesize IRCNN \textcolor{black}{(ENA)} & \footnotesize  IRCNN+ \textcolor{black}{(ENA)} & \footnotesize DPIR \textcolor{black}{(ENA)}& \footnotesize  HODL\\

		\end{tabular}
	\end{center}
\vspace{-0.134cm}
	\caption{Visual results of the image deconvolution task on two samples, compared with FDN, IRCNN, IRCNN+, and DPIR. 
		The hierarchical modeling of HODL improves the clarity of details and maintains the high level of color restoration. 
		Two metrics (PSNR / SSIM) are listed below each image to quantify the quality of generated images. 
		Best and second best results are marked in red and blue respectively.
	}
	\label{fig:deconv}
\end{figure*}

\subsection{Vision Tasks}

This subsection provides experimental results in vision tasks including rain streak removal, image deconvolution, and low-light enhancement.

\textbf{Rain Streak Removal}. %
In the rain streak removal task, we use datasets Rain100L and Rain100H~\cite{derain}. 
As a constrained problem, $\D_\num$ is set to be $\D_{\mathtt{ALM}}$.
For the network architecture $\D_\net$, we adopt a 2-layer convolutional network with $\u_r$ and $\b$ as the network input to estimate $\u_b$, and a 3-layer convolutional network with $\u_b,\u_r$, and $\b$ as the input to estimate $\u_r$.
In the network to estimate $\u_r$, some prior information of $\u_r$ is employed as input just like in~\cite{wang2020model}. 
In practice, we decide proper $\Omega$ such that for all $\ome \in \Omega$ it holds that $ \H_{\ome} \succ 0$,
and $\H_{\ome}$ can be inverted fast by Fourier transform.
\textcolor{black}{Here we use MSE as the loss function and use Adam optimizer with step size $0.001$, and set batchsize~$=64$.}

We report the quantitative comparison of HODL in Table~\ref{tab:derain_table}
with a series of state-of-the-art methods.
It can be seen that on both benchmark datasets HODL achieves higher PSNR and SSIM. 
Note that HODL has a competitive performance compared with RCDNet, and it possesses superior theoretical property as well.
In Figure~\ref{fig Deraining results},
we visually present the performance of rain streak removal task on two images from Rain100L~\cite{derain}, compared with DDN~\cite{fu2017removing}, JORDER~\cite{derain}, PReNet~\cite{ren2019progressive} and RCDNet~\cite{wang2020model}.
From both rows, one can observe that our HODL preserves the original counter line of wall and roof in the background and performs the best on PSNR and SSIM, while other methods produce some unsatisfactory distortion, blur some textures, or even leave noticeable rain streaks.

\textbf{Image Deconvolution}.
In the image deconvolution task, similar to~\cite{zhang2020plug}, 
we use a large dataset containing 400 images from Berkeley Segmentation Dataset, 
4744 images from Waterloo Exploration Database, 900 images from DIV2K Dataset, 
and 2750 images from Flick2K Dataset. 
As for the network architectures $\D_\net$, we use DRUNet containing four scales, 
each of which has an identity skip connection between 2 $\times$ 2 strided convolution downscaling and 2 $\times$ 2 transposed convolution upscaling operators. 
From the first scale to the fourth scale, the numbers of channels in each layer are respectively 64, 128, 256, and 512. 
We employ four successive residual blocks in the downscaling and upscaling of each scale. 
For the numerical update operator $\D_\num$, by introducing the auxiliary variable $\z=\W\u$, 
we transform the objective function to be $\Vert\Q\W^{-1}\z-\b\Vert^2_2$ with a regularization term $\Vert\z\Vert_1$. 
\textcolor{black}{Here we use MSE as the loss function for $\ome$, use downsample mode as strideconv, upsample mode as convtranspose, and Adam optimizer with step size $0.001$, and set batchsize $=1$.}

\begin{table}
	\caption{PSNR (dB) results compared with state-of-the-art methods 
		for the image deconvolution task with noise levels $\sigma=1\%$ and $3\%$. 
		Best and second best results are marked in red and blue respectively.
	} 
	\renewcommand\arraystretch{1.1}
	\setlength\tabcolsep{2pt}
	\begin{center}
		\begin{small}
		\resizebox{0.5\textwidth}{!}{
			\begin{tabular}{c|ccc|ccc}
				\hline
				Noise level &\multicolumn{3}{c|}{$\sigma=1\%$}&\multicolumn{3}{|c}{$\sigma=3\%$}\\
				Image&Butterfly&Leaves&Starfish&Butterfly&Leaves&Starfish\\
				\hline
				EPLL&20.55&19.22&24.84&18.64&17.54&22.47\\
				FDN&27.40&26.51&27.48&24.27&23.53&24.71\\
				IRCNN \textcolor{black}{(ENA)}&32.74&33.22&33.53&28.53&28.45&28.42\\
				IRCNN+ \textcolor{black}{(ENA)}&32.48&33.59&32.18&28.40&28.14&28.20\\
				DPIR \textcolor{black}{(ENA)}&\textcolor{red}{\textbf{34.18}}&\textcolor{blue}{\textbf{35.12}}&\textcolor{blue}{\textbf{33.91}}&\textcolor{blue}{\textbf{29.45}}&\textcolor{blue}{\textbf{30.27}}&\textcolor{blue}{\textbf{29.46}}\\
				HODL&\textcolor{blue}{\textbf{33.67}}&\textcolor{red}{\textbf{35.39}}&\textcolor{red}{\textbf{33.98}}&\textcolor{red}{\textbf{29.46}}&\textcolor{red}{\textbf{30.69}}&\textcolor{red}{\textbf{29.64}}\\
				
				\hline
			\end{tabular}
		}
		\end{small}
	\end{center}
	\label{tab:deblurring table}
\end{table}

\begin{figure}
	\centering
	\includegraphics[height=3cm,width=4cm]{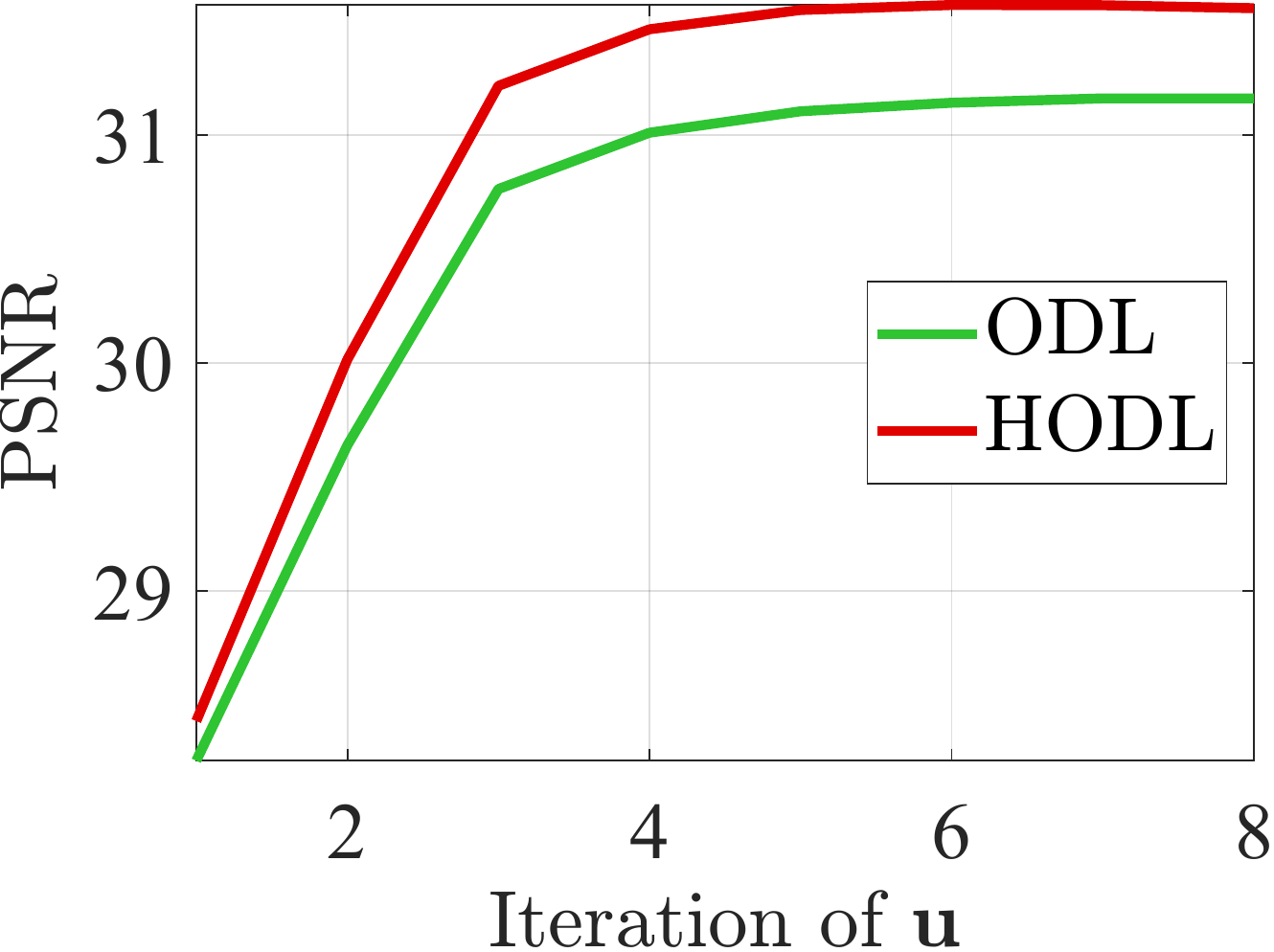}
	\includegraphics[height=3cm,width=4cm]{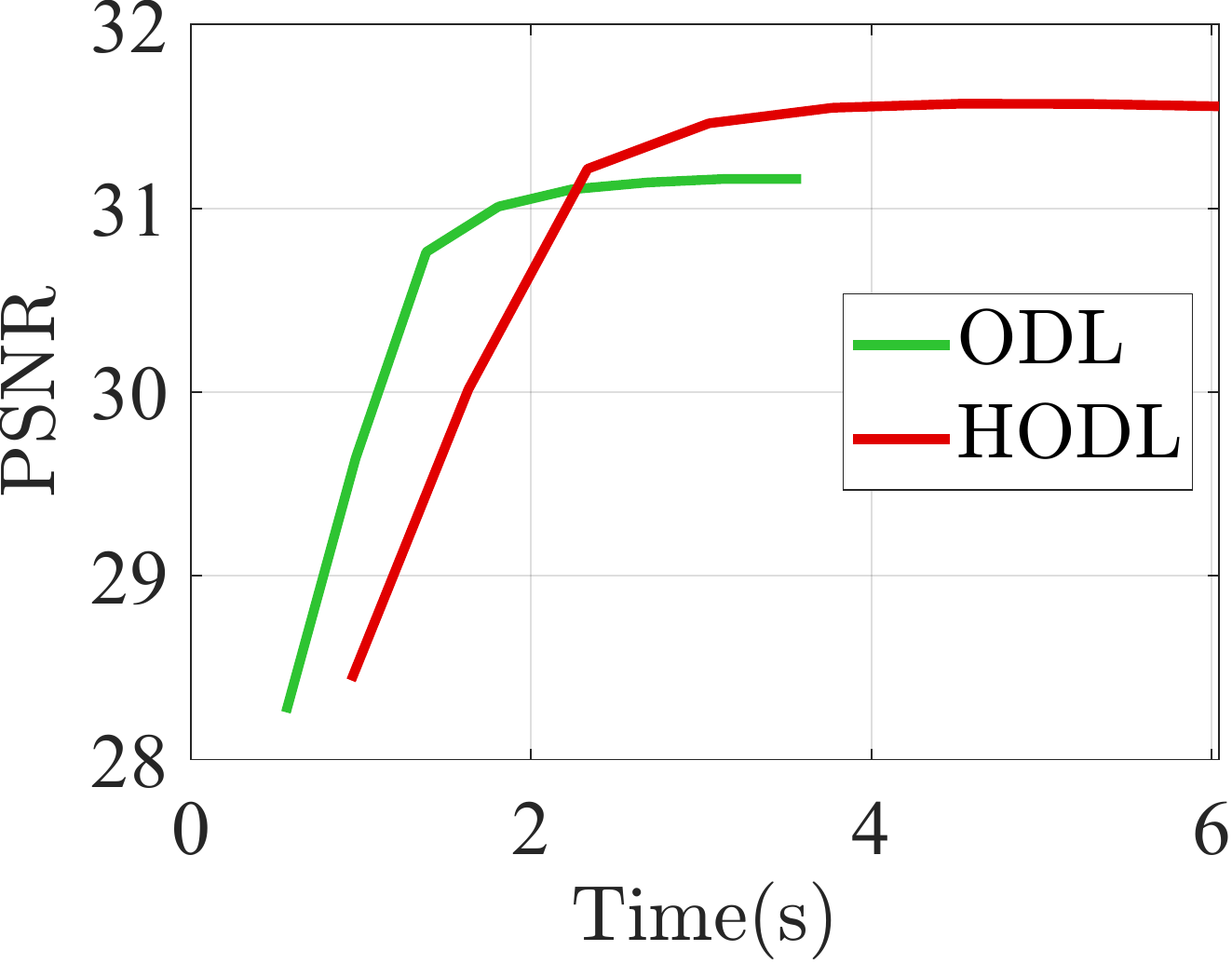}	
	\caption{\color{black}{PSNR results of ODL (DPIR which can be classified as ENA) and HODL with iteration of~$\u$ and inference time for the image deconvolution task.} }
	\label{fig:deblurtime}
\end{figure}

\begin{table*}[h!]
	\begin{center}  %
		\caption{Quantitative results (PSNR, SSIM and LPIPS) on the MIT and LOL datasets for low-light enhancement. 
			Best and second best results are marked in red and blue respectively.}
		\label{tab:real_lle}%
		\setlength{\tabcolsep}{1.9mm}{
			\begin{tabular}{cc|c|c|c|c|c|c|c|c|c}
				\hline  
				Datasets & Metrics & MBLLEN & GLADNet (UNH) & RetinexNet \textcolor{black}{(ENA)} & KinD & ZeroDCE & FIDE & EnGAN & DRBN (ENA)  & HODL\\
				\hline 
				
				\multirow{3}[0]{*}{MIT} & PSNR  & 15.59 & 16.73 & 12.69 &\textcolor{blue}{ \textbf{17.17}} & 16.46 & \textcolor{blue}{\textbf{17.17}} & 15.95 & 15.01 & \textcolor{red}{\textbf{20.54}}   \\
				& SSIM  & 0.71  &  \textcolor{blue}{\textbf{0.76}}  & 0.64   & 0.70   &  \textcolor{blue}{\textbf{0.76}}  & 0.70   & 0.70   &  \textcolor{red}{\textbf{0.77}}  &  \textcolor{red}{\textbf{0.77}} \\
				& LPIPS & 0.31  & 0.69  & 0.34  & 0.32  & 0.23  &\textcolor{blue} {\textbf{0.19}}  & 0.29  & 0.21  &    \textcolor{red}{\textbf{0.09}}\\
				\hline  
				\multirow{3}[0]{*}{LOL} & PSNR  & 13.93 & 16.19 & 13.10 & 14.62 & 15.51 & \textcolor{blue}{\textbf{16.72}} & 15.32 & 15.83 &  \textcolor{red}{\textbf{20.86}} \\
				& SSIM  & 0.49  & 0.61  & 0.43  & 0.64  & 0.55  & 0.67  &\textcolor{blue} {\textbf{0.70}}  & 0.64  &  \textcolor{red}{\textbf{0.82}} \\
				& LPIPS & 0.70   & \textcolor{blue}{\textbf{0.21}}  & 0.86   & 0.68  & 0.68  & 0.72  & 0.51  & 0.36  &   \textcolor{red}{\textbf{0.08}}\\
				\hline
			\end{tabular}%
		}
	\end{center}
\end{table*}%

\begin{figure*}[!t]
	\begin{center}
		\begin{tabular}{c@{\extracolsep{1em}}c@{\extracolsep{1em}}c@{\extracolsep{1em}}c@{\extracolsep{1em}}c@{\extracolsep{1em}}c@{\extracolsep{1em}}}
			\footnotesize Target & \footnotesize VGAN & \footnotesize WGAN & \footnotesize ProxGAN & \footnotesize LCGAN & \footnotesize HODL \\
			{\includegraphics[height=2.2 cm,trim=0 0 0 0,clip]{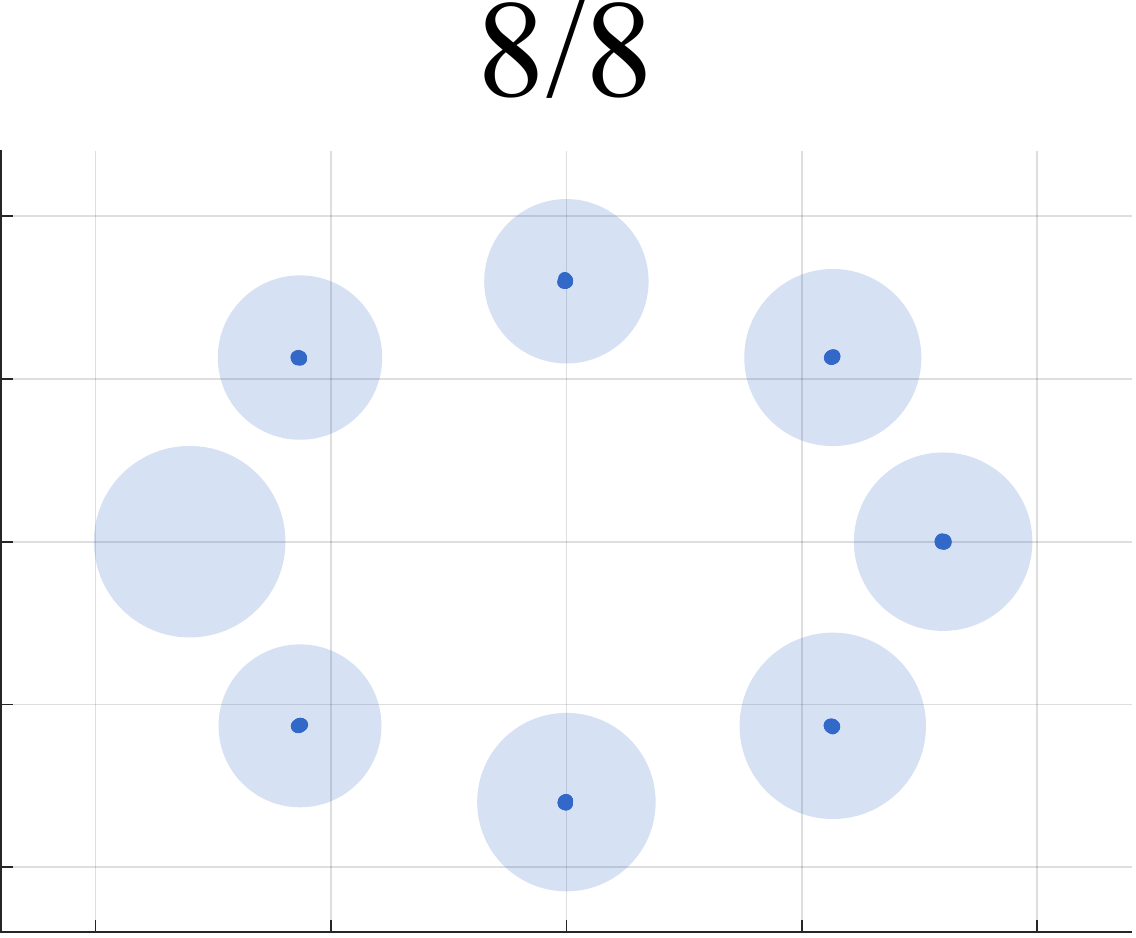}}
			&\includegraphics[height=2.2cm]{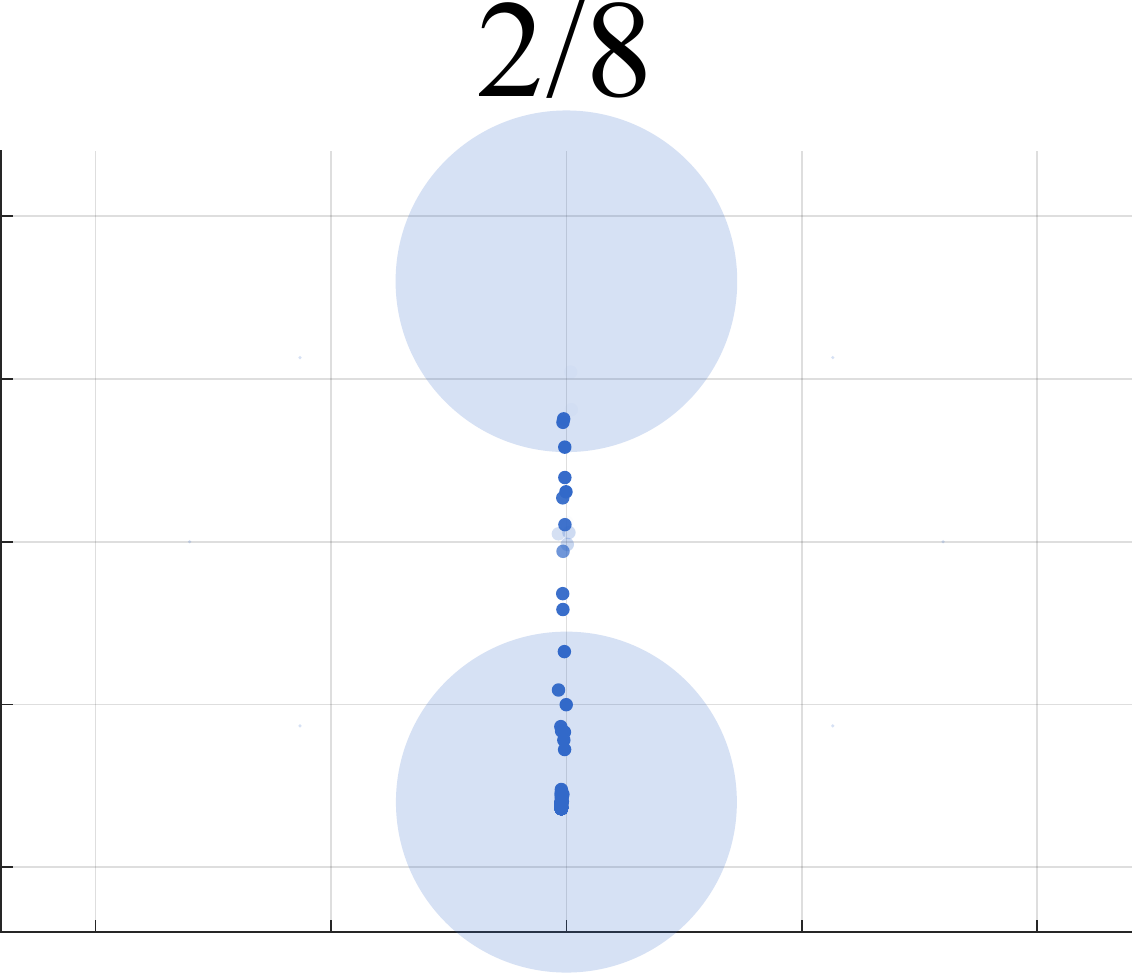}
			& \includegraphics[height=2.2cm]{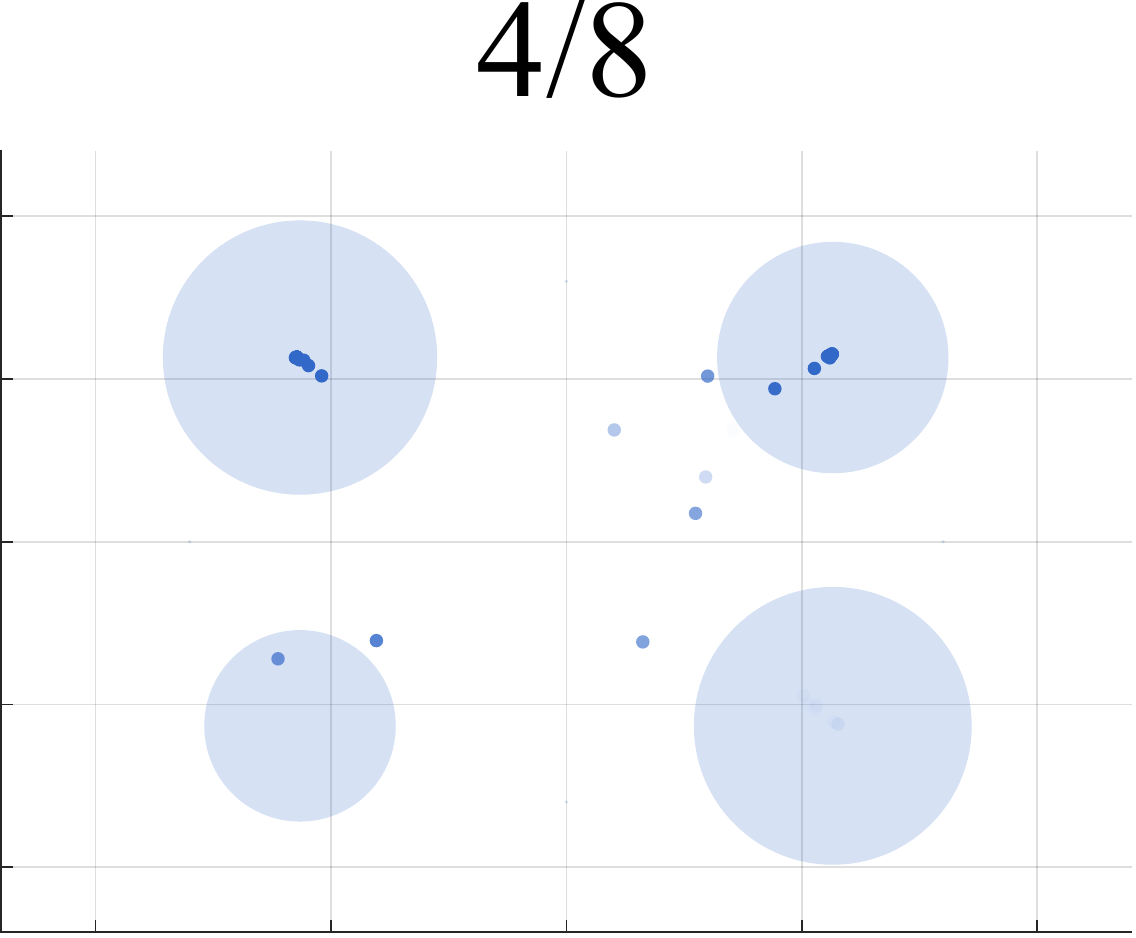} 
			& \includegraphics[height=2.2cm]{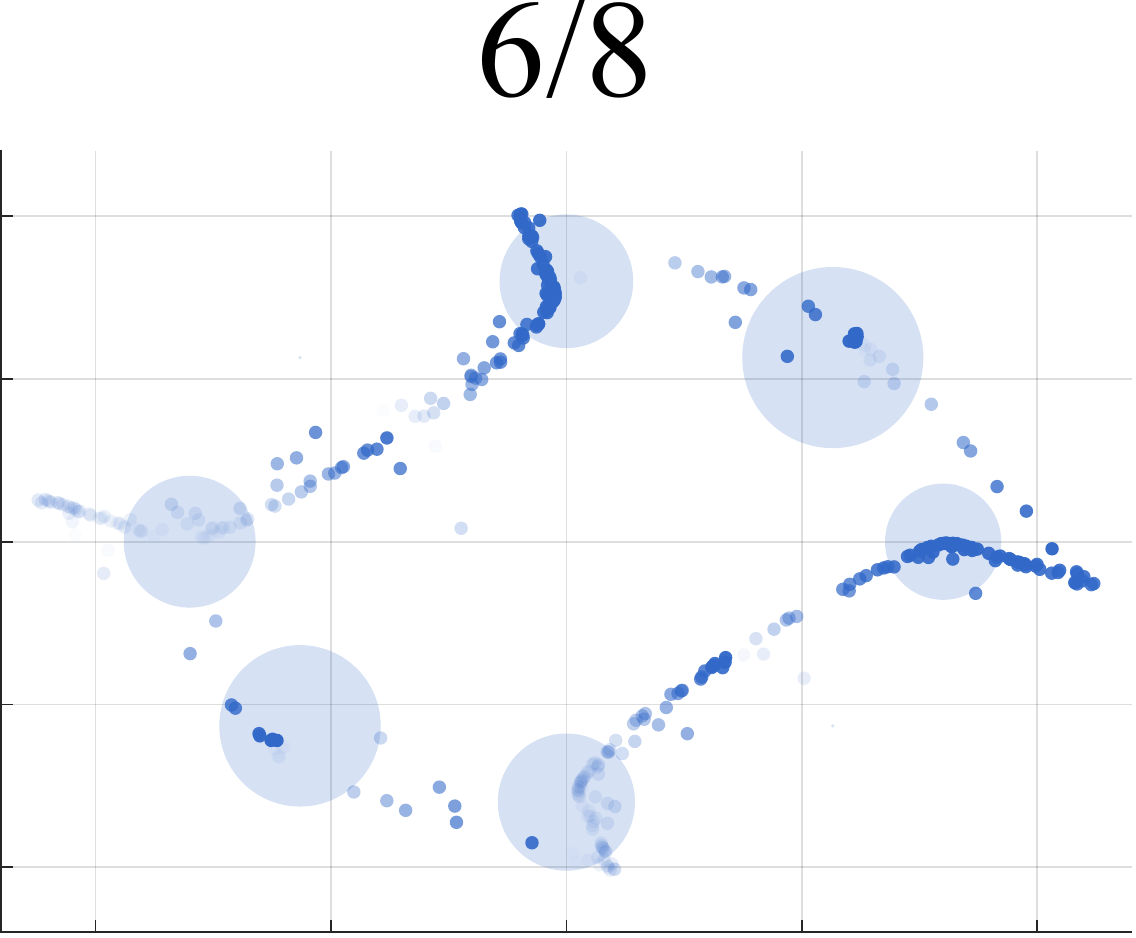} 
			& \includegraphics[height=2.2cm]{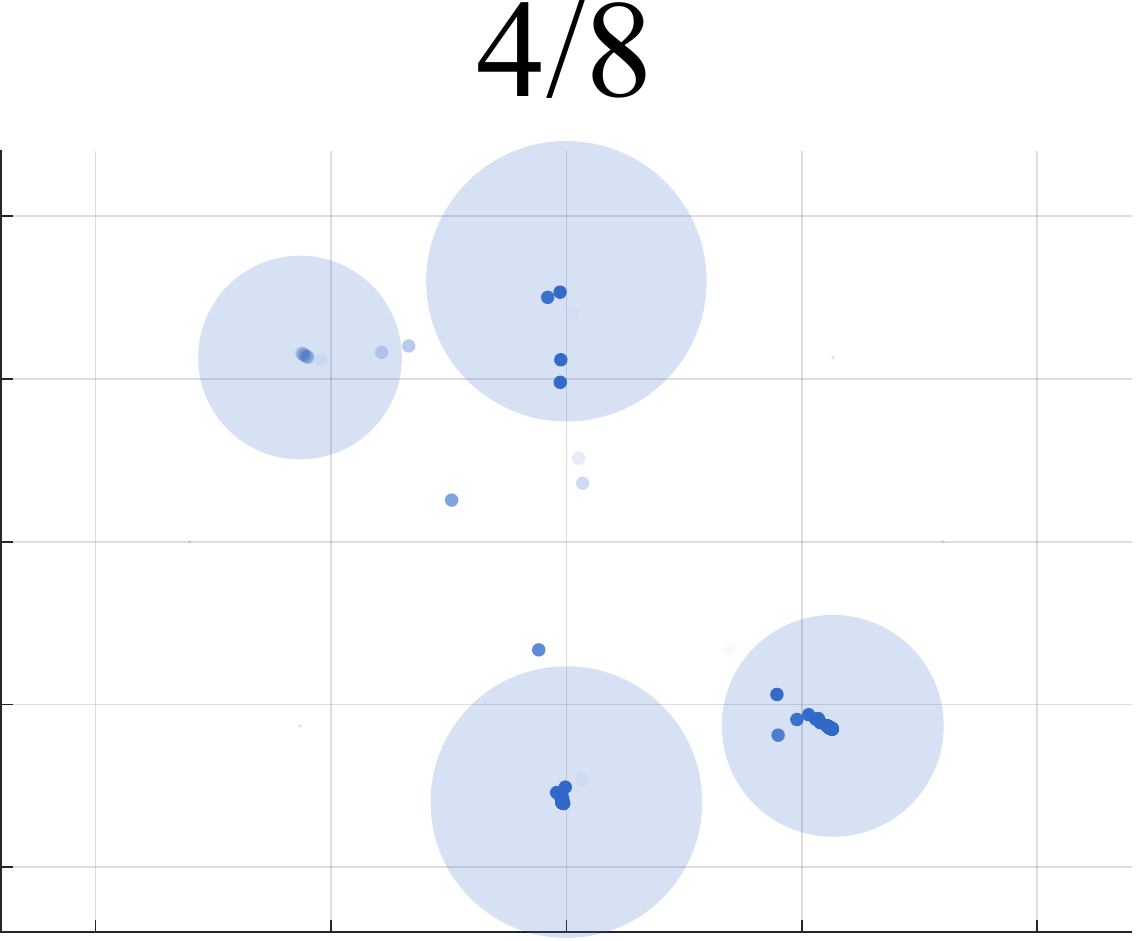}
			&\includegraphics[height=2.2cm]{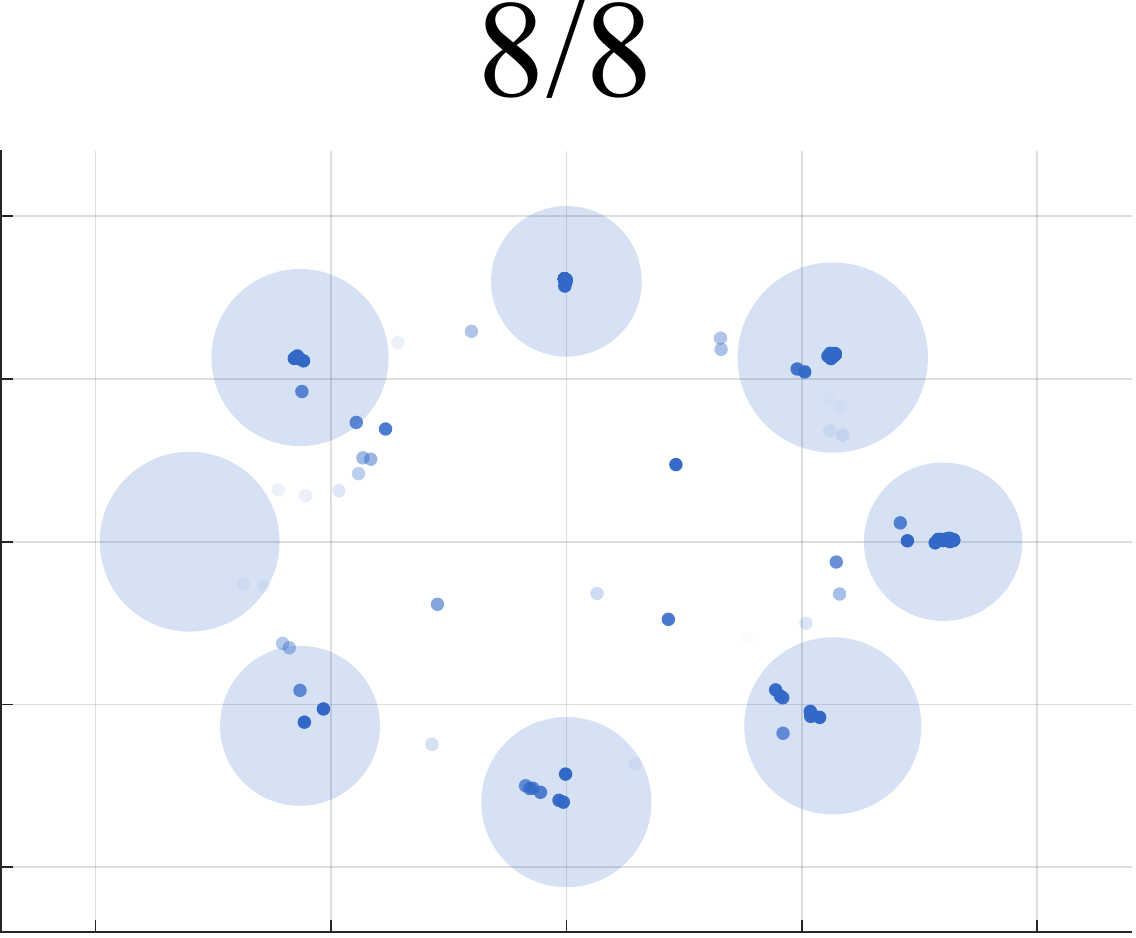}\\
		\end{tabular}
	\end{center}
	\caption{Comparison among four mainstream GAN methods (i.e., vanilla GAN (VGAN), WGAN, ProxGAN, and LCGAN) and HODL on the synthetic 2D ring mixed of Gaussian distribution data. 
		The gap of generated samples (generated/targeted number of classes) is listed on the top. 
		The shading of dots represents the density of final distribution, with darker dots representing greater density.}
	\label{fig:syn_mode}
\end{figure*}

For the practical application in image deconvolution, 
we verify the performance of HODL on three classical testing images in Table~\ref{tab:deblurring table},
and compare our method with representative methods. 
For traditional methods, we compare with numerically designed method EPLL~\cite{zoran2011learning} and
learning-based method FDN~\cite{kruse2017learning}. 
For ODL methods,  we compare with IRCNN, IRCNN+, and DPIR ~\cite{zhang2017learning,zhang2020plug}. %
By applying a meta-optimization perspective on handcrafted network $\D_{\net}$ and numerical schemes $\D_{\mathtt{PG}}$ as a regularized problem, HODL performs best in the last five columns and achieves top two in the first, in three testing images of different noise levels.
\textcolor{black}{
Note that here we choose DRUNet in DPIR~\cite{zhang2020plug} as $\D_{\net}$ for HODL,
and the overall preferable results of HODL than directly using DPIR demonstrate the effect of compositing of $\D_{\num}$ and $\D_{\net}$
and the ability of HODL to improve the performance based on previous methods.
}
In addition, we show the visual results in Figure~\ref{fig:deconv}. It can be seen that our method is superior to other methods in color restoration, detail retention and quantitative metrics.
\textcolor{black}{Figure~\ref{fig:deblurtime} further compares the computational efficiency of ODL (DPIR) and HODL.
It can be seen that HODL can reduce the number of iterations without performance degradation.
}

\textbf{Low-light Enhancement.}
To further verify the effectiveness of our method on low-level vision tasks, 
we conduct experiments in the low-light enhancement task. 
Specifically, we perform experiments on two prominent MIT and LOL datasets, 
and adopt PSNR, SSIM and LPIPS as our evaluated metrics.
\textcolor{black}{Here we use MSE as the loss function for $\ome$,
set batchsize $=2$, and use SGD optimizer with step size $0.015$. 
As for~$\D_{\net}$, by adopting the continuous relaxation technique used in differentiable Neural Architecture Search (NAS) literature, we search the network structure in the search space which includes $1\times1$ and $3\times3$ Convolution, $1\times1$ and $3\times3$ Residual Convolution, $3\times3$ Dilation Convolution with dilation rate of $2$, $3\times3$ Residual Dilation Convolution with dilation rate of $2$, and Skip Connection~\cite{liu2021retinex}. Then we add spectral normalization to ensure the non-expansive property. 
}
For a complete evaluation, we compare HODL with MBLLEN~\cite{lv2019attention},  GLADNet~\cite{wang2018gladnet}, 
RetinexNet~\cite{Chen2018Retinex},  KinD~\cite{zhang2019kindling}, ZeroDCE~\cite{guo2020zero}, FIDE~\cite{xu2020learning}, EnGAN~\cite{jiang2019enlightengan}, 
and DRBN~\cite{yang2020fidelity}.  
In the first three rows of Table~\ref{tab:real_lle}, we evaluate HODL quantitatively on the MIT Adobe 5K dataset as a simple real-world scenario.
In the last three rows of Table~\ref{tab:real_lle}, we also perform a quantitative assessment on the LOL dataset that increases the difficulty of enhancement due to the inclusion of sensible noise as a demonstration on extremely challenging real world scenarios.
It can be seen that HODL obtains the best results on both~datasets.

\begin{table}[htbp]
	\caption{Comparison with existing methods for solving the data hyper-cleaning task on MNIST and FashionMNIST as an example of hyper-parameter optimization. The F1 score denotes the harmonic mean of precision and recall.
		Best and second best results are marked in red and blue respectively.}
	\begin{center}
		\begin{small}
			\begin{tabular}{c|cc|cc}
				\hline
				\multirow{2}{*}{Method}&\multicolumn{2}{c|}{MNIST}&\multicolumn{2}{c}{FashionMNIST}\\
				&Acc.&F1 score&Acc.&F1 score\\
				\hline
				RHG&87.90&89.36&81.91&87.12\\
				TRHG&88.57&\textcolor{blue}{\textbf{89.77}}&81.85&86.76\\
				CG&\textcolor{blue}{\textbf{89.19}}&85.96&\textcolor{red}{\textbf{83.15}}&85.13\\
				NS&87.54&89.58&81.37&\textcolor{blue}{\textbf{87.28}}\\
				HODL&\textcolor{red}{\textbf{89.75}}&\textcolor{red}{\textbf{90.38}}&\textcolor{blue}{\textbf{82.04}}&\textcolor{red}{\textbf{88.24}}\\
				\hline
			\end{tabular}
		\end{small}
	\end{center}
	\label{tab:hypercleaning}
\end{table}

\begin{figure}[htbp]
	\begin{center}
		\includegraphics[height=3.5cm,width=3.5cm]{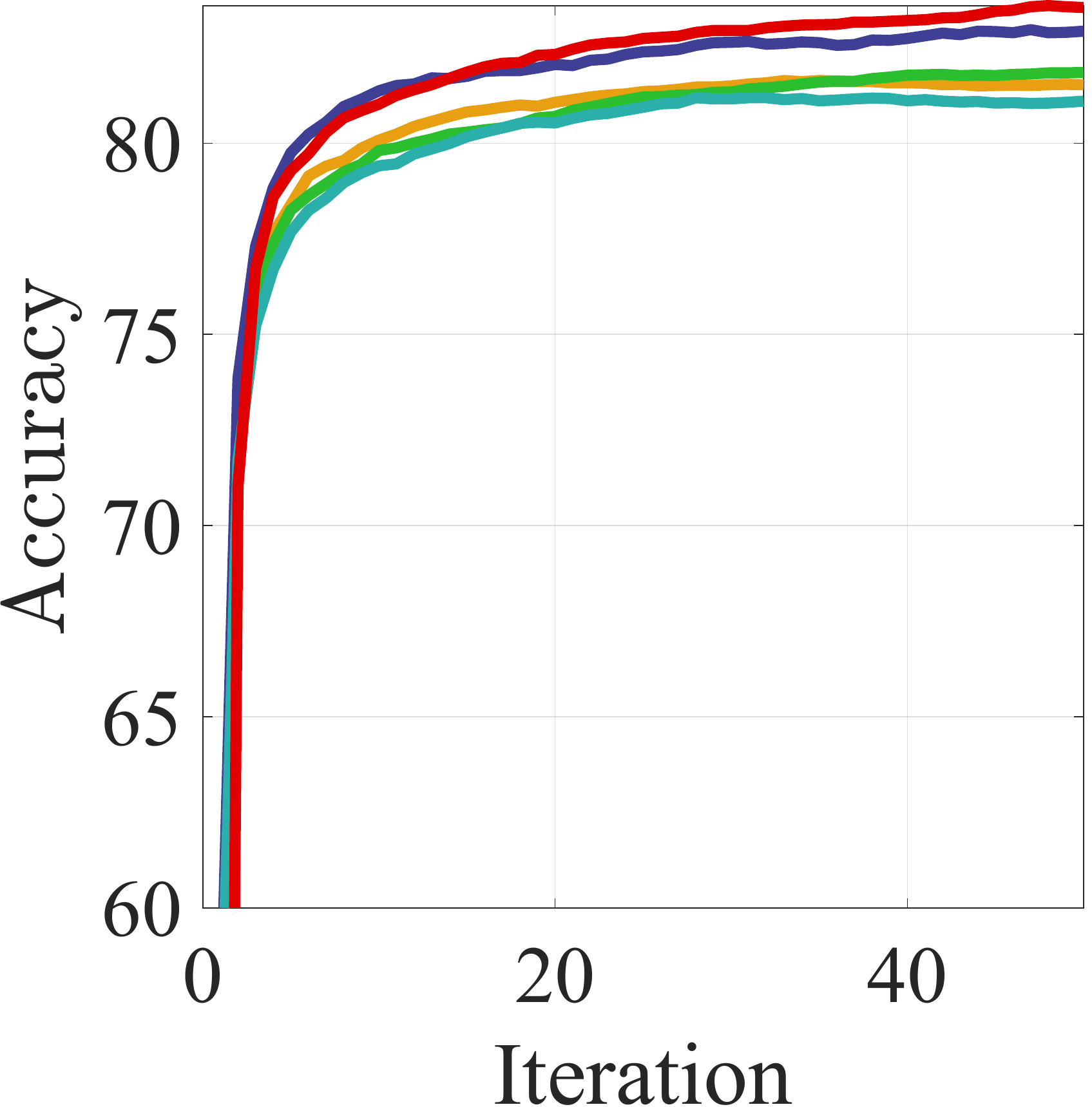}
		\includegraphics[height=3.5cm,width=3.5cm]{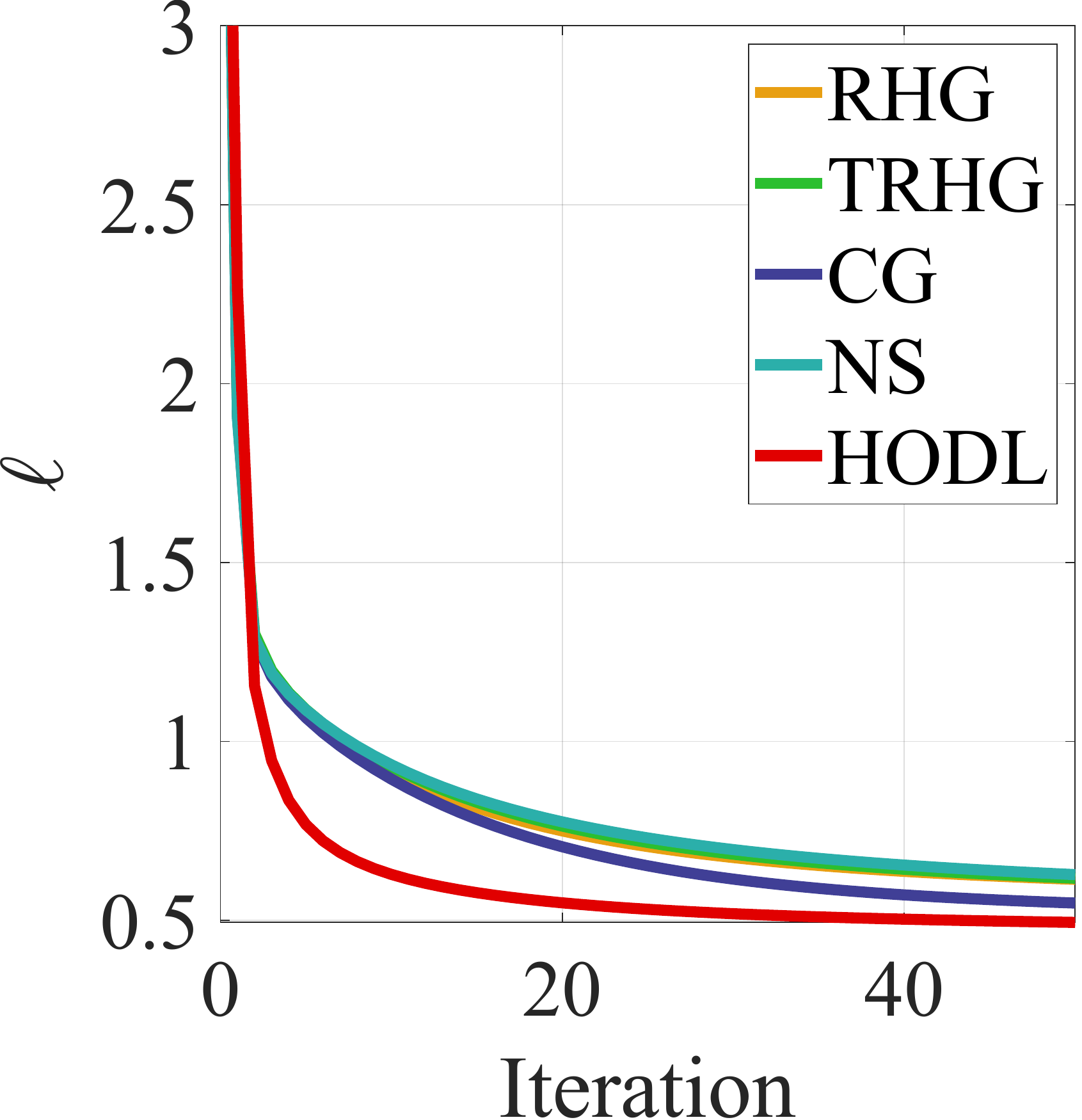}
	\end{center}
	\caption{Comparison of the accuracy and validation loss
		$\ell$ for data hyper-cleaning as an example of hyper-parameter optimization with other unrolling algorithms.}
	\label{fig:acc loss}
\end{figure}

\begin{table}
	\caption{The averaged accuracy for few-shot classification on Omniglot and MiniImageNet datasets ($N$-way $M$-shot with $M=1$ and $N=5,20$).
		Best and second best results are marked in red and blue respectively.}
	\begin{center}
		\begin{small}
			\begin{tabular}{c|cc|c}
				\hline
				\multirow{2}{*}{Method}&\multicolumn{2}{c|}{Omniglot}&{MiniImageNet}\\
				&5-way&20-way&5-way\\
				\hline
				RHG&98.60&95.50&\textcolor{blue}{\textbf{48.89}}\\
				TRHG&98.74&95.82&47.67\\
				CG&\textcolor{blue}{\textbf{98.96}}&94.46&48.42\\
				NS&98.40&\textcolor{blue}{\textbf{96.06}}&48.61\\
				HODL&\textcolor{red}{\textbf{99.04}}&\textcolor{red}{\textbf{96.50}}&\textcolor{red}{\textbf{49.08}}\\
				
				\hline
				
			\end{tabular}
		\end{small}
	\end{center}
	\label{few-shot learning}
	
\end{table}

\subsection{Extended Applications}

The followings are experimental results on other learning tasks beyond ODL introduced in Section~\ref{sec:extension} as the extended applications of HODL.

\textbf{Adversarial Learning.}
In the adversarial learning task, we visualize the two-dimensional mixed Gaussian distribution data to verify the effectiveness of our method.
\textcolor{black}{We set batchsize $=32$ and use SGD optimizer with step size $0.001$.}
Performance of HODL is investigated compared to current mainstream and well-known GAN architectures which mitigate mode collapse and maintain stable training, including vanilla GAN (VGAN)~\cite{goodfellow2014generative}, WGAN~\cite{arjovsky2017wasserstein}, ProxGAN~\cite{farnia2020gans}, LCGAN~\cite{engel2017latent}.
Figure~\ref{fig:syn_mode} visually shows a comparison of results by various methods regarding the number of samples generated.
One can find that mainstream GAN methods only capture a part of distributions, getting into severe mode collapse dilemma and failing to achieve satisfactory performance, 
while our HODL generates all modes and is significantly better than other methods.

\textbf{Hyper-parameter Optimization}.
In this experiment, we consider a widely used hyper-parameter optimization example, i.e., data hyper-cleaning, to evaluate the HODL. Assuming that some labels in our dataset are contaminated, the purpose of data hyper-cleaning is to reduce the impact of incorrect samples by adding hyper-parameters. 
We follow the settings in \cite{liu2020generic} and conduct experiments on MNIST and FashionMNIST datasets. 
\textcolor{black}{We set batchsize $=32$ and use Adam optimizer with step size $0.01$.}
To demonstrate the advantage of our method, we show the accuracy and F1 scores in Table~\ref{tab:hypercleaning}, compared with different methods containing Reverse Hyper-Gradient (RHG)~\cite{franceschi2017forward}, Truncated RHG (TRHG)~\cite{shaban2019truncated}, Conjugate Gradient (CG)~\cite{rajeswaran2019meta}, and Neumann Series (NS)~\cite{lorraine2020optimizing}. 
Figure~\ref{fig:acc loss} also shows the  accuracy and validation loss using different methods. 
It can be seen that our method achieves higher accuracy, higher F1 score, and~lower loss.

\textbf{Few-shot Learning}.
Next, we test the application in few-shot learning under high dimensions on Omniglot and MiniImageNet datasets to verify the computational efficiency of our method.
In this experiment, we follow the settings in~\cite{liu2020generic}. 
\textcolor{black}{We set batchsize~$=4$ and use Adam optimizer with step size $0.01$.}
It can be seen in Table~\ref{few-shot learning} that our HODL gives the best performance in different tasks.

\section{Conclusions} %

This paper first 
proposes the HODL framework to nest the optimization and learning processes in ODL problems,
and then presents solution strategies for HODL to jointly solve the optimization variables and learning variables.
We prove the joint convergence of  optimization variables and learning variables from the perspective of both the approximation quality, and the stationary analysis.
Experiments demonstrate our efficiency on sparse coding, real-world applications in image processing (e.g., rain streak removal, image deconvolution, and low-light enhancement), and other learning tasks (e.g., adversarial learning, hyper-parameter optimization and few-shot learning).
\textcolor{black}{
	As a flexible and general framework,
	HODL is also applicable for various networks designed for large-scale problems in real-world applications.
	Exploring HODL on more large-scale datasets and large neural networks is a future direction.
}

\ifCLASSOPTIONcompsoc
\section*{Acknowledgments}
\else
\section*{Acknowledgment}
\fi

%This work is partially supported by the National Natural Science Foundation of China (Nos. U22B2052, 61922019, 12222106), 
%the National Key R\&D Program of China (2020YFB1313503, 2022YFA1004101), 
%Shenzhen Science and Technology Program (No. RCYX20200714114700072), the Guangdong Basic and Applied Basic Research Foundation (No. 2022B1515020082),
%and Pacific Institute for the Mathematical Sciences (PIMS).

This work was supported in part by 
the National Key R\&D Program of China under Grants 2020YFB1313503 and 2022YFA1004101, 
in part by the National Natural Science Foundation of China under Grants U22B2052 and 12222106, 
in part by Shenzhen Science and Technology Program under Grant RCYX20200714114700072, 
in part by Guangdong Basic and Applied Basic Research Foundation under Grant 2022B1515020082,
in part by Shandong Province Natural Science Foundation under Grant ZR2023MA020,
and in part by Pacific Institute for the Mathematical Sciences (PIMS).

\ifCLASSOPTIONcaptionsoff
\newpage
\fi

\bibliographystyle{IEEEtran}
\bibliography{output}

\begin{IEEEbiography}[{\includegraphics[width=1in,height=1.25in,clip,keepaspectratio]{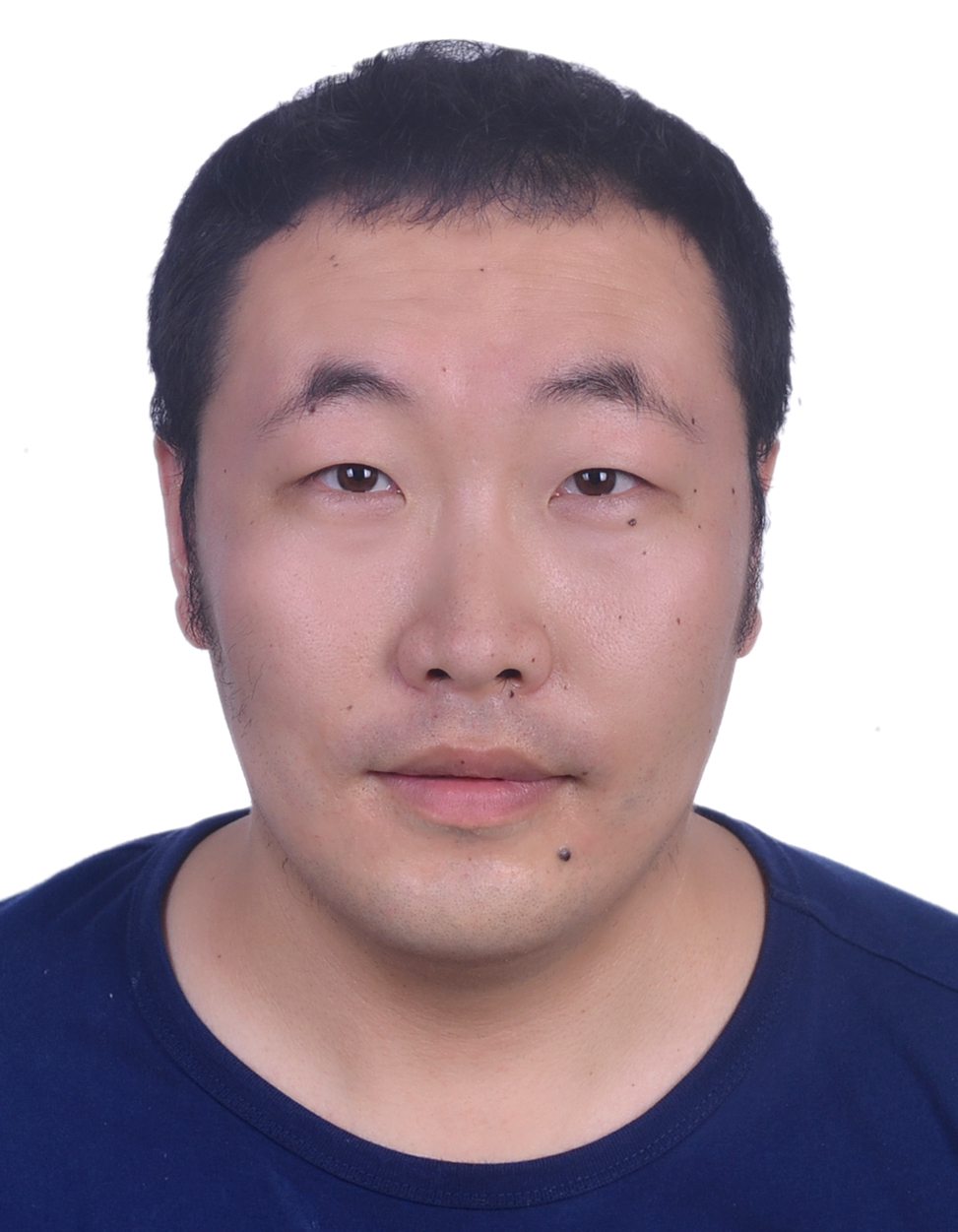}}]{Risheng Liu}
	(Member, IEEE)
	received the BSc
	and PhD degrees in mathematics from Dalian
	University of Technology, in 2007 and 2012,
	respectively. He was a visiting scholar with the
	Robotics Institute, Carnegie Mellon University,
	from 2010 to 2012. He served as a Hong Kong
	Scholar research fellow with the Hong Kong Polytechnic University from 2016 to 2017. He is currently a professor with the DUT-RU International
	School of Information Science \& Engineering,
	Dalian University of Technology. His research interests include machine learning, optimization, computer vision, and multimedia.
	
\end{IEEEbiography}

\begin{IEEEbiography}[{\includegraphics[width=1in,clip,keepaspectratio]{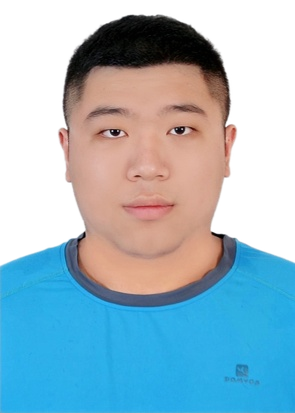}}]{Xuan Liu}
	received the BSc degree in  mathematics from Dalian University of Technology, in 2020.  He is currently working toward the MPhil degree with the Department of Software Engineering, Dalian University of Technology. His research interests include computer vision, machine learning, and control and optimization.
\end{IEEEbiography}

\begin{IEEEbiography}[{\includegraphics[width=1in,height=1.25in,clip,keepaspectratio]{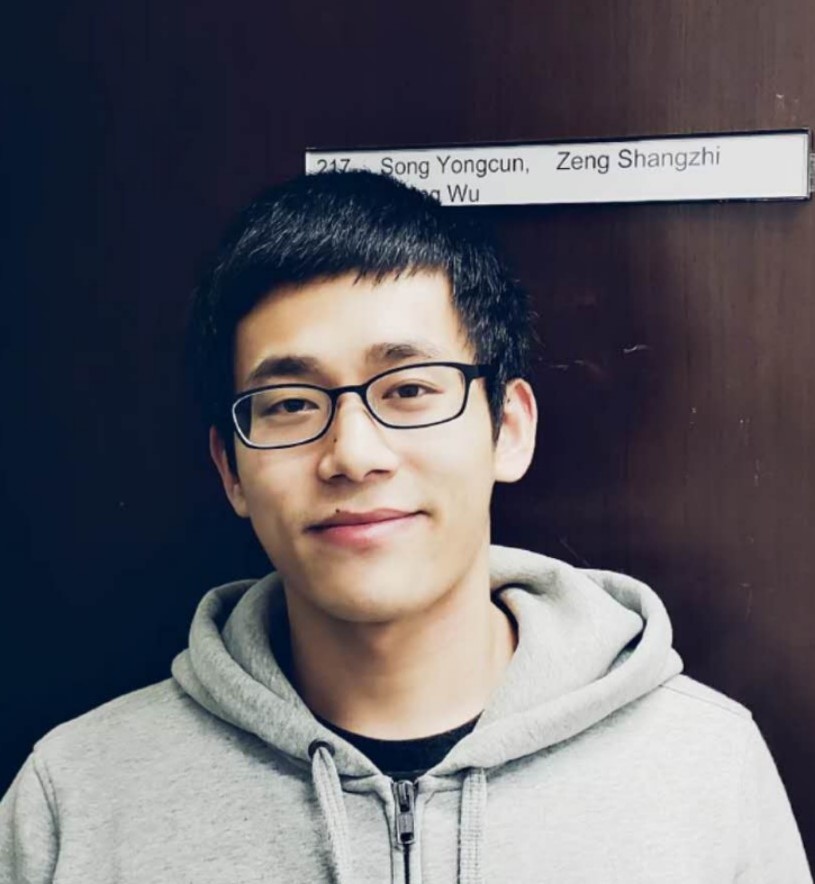}}]{Shangzhi Zeng}
	received the BSc degree in
	mathematics and applied mathematics from
	Wuhan University, in 2015, the MPhil degree
	from Hong Kong Baptist University, in 2017, and
	the PhD degree from the University of Hong
	Kong, in 2021. He is currently a PIMS postdoctoral
	fellow with the Department of Mathematics and
	Statistics at University of Victoria. His current
	research interests include variational analysis
	and bilevel optimization.
\end{IEEEbiography}

\begin{IEEEbiography}[{\includegraphics[width=1in,height=1.32in,clip,keepaspectratio]{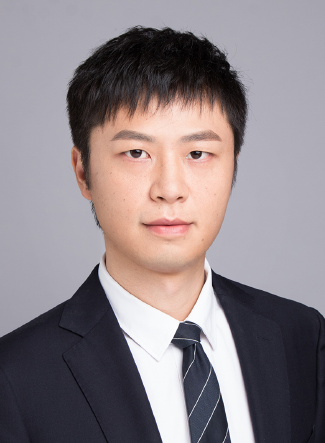}}]{Jin Zhang}
	received the BA degree in journalism
	and the MPhil degree in mathematics and operational
	research and cybernetics from Dalian University of Technology, China, in 2007 and 2010, respectively,
	and the PhD degree in applied mathematics from
	University of Victoria, Canada, in 2015. After
	working with Hong Kong Baptist University for
	three years, he joined Southern University of
	Science and Technology as a tenure-track assistant
	professor with the Department of Mathematics
	and promoted to an associate professor in 2022. His
	broad research area is comprised of optimization,
	variational analysis and their applications in economics, engineering, and
	data science. %
\end{IEEEbiography}

\begin{IEEEbiography}[{\includegraphics[width=1in,clip,keepaspectratio]{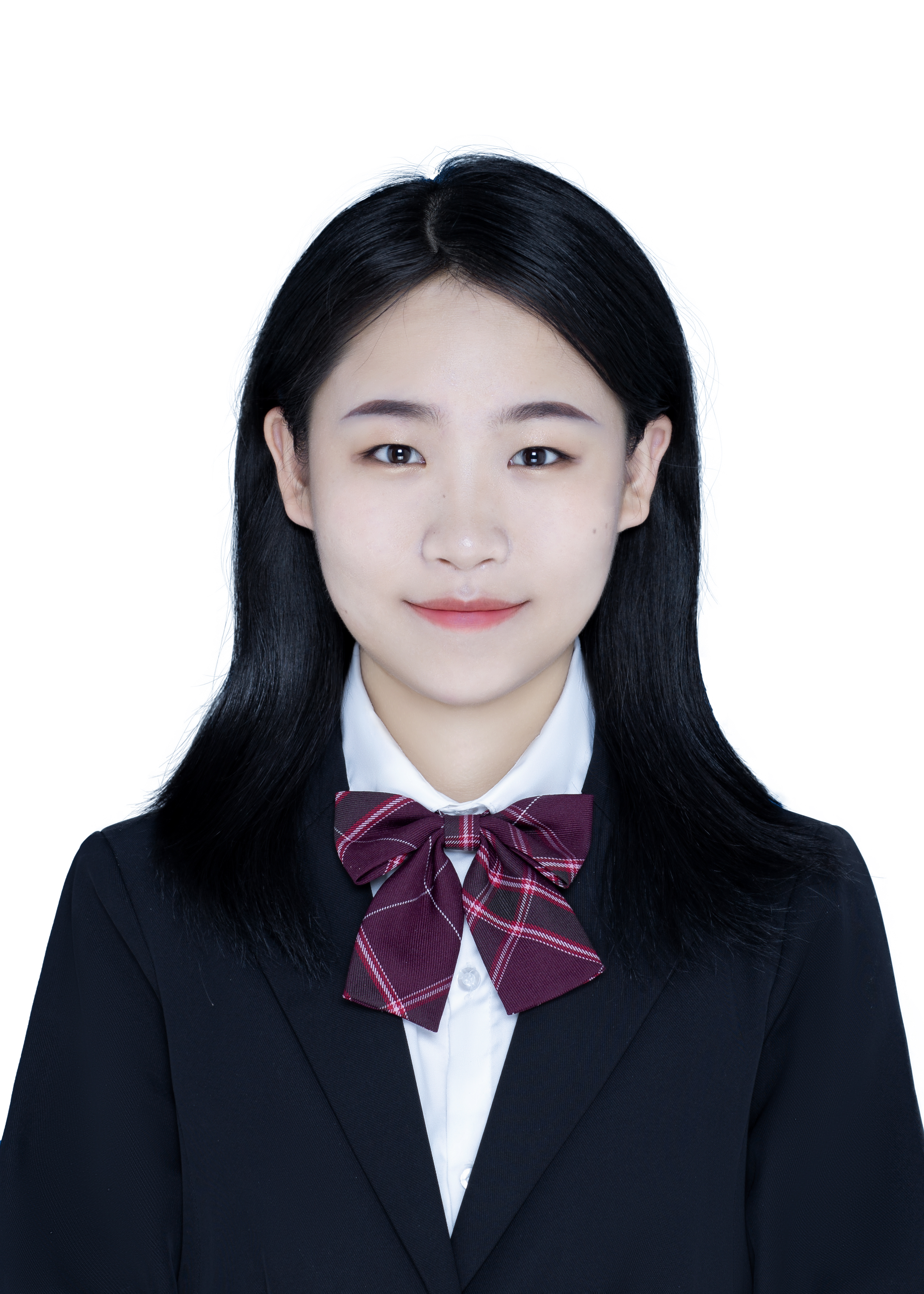}}]{Yixuan Zhang}
	received the BSc degree in mathematics and applied mathematics from Beijing Normal University, in 2020,
	and the MPhil degree from Southern University of Science and Technology, in 2022.
	She is currently working toward the PhD degree with the Department of Applied Mathematics, the Hong Kong
	Polytechnic University.
	Her current research interests include optimization and machine learning.
	
\end{IEEEbiography}

\end{document}